\documentclass{article}

\usepackage{microtype}
\usepackage{graphicx}
\usepackage{subfigure}
\usepackage{booktabs} 

\usepackage{amsmath,amsfonts,bm}

\def\eqref#1{equation~\ref{#1}}

\def\1{\bm{1}}

\DeclareMathAlphabet{\mathsfit}{\encodingdefault}{\sfdefault}{m}{sl}
\SetMathAlphabet{\mathsfit}{bold}{\encodingdefault}{\sfdefault}{bx}{n}

\newcommand{\R}{\mathbb{R}}

\usepackage{hyperref}
\usepackage{url}
\usepackage{graphicx}
\usepackage{enumitem}
\usepackage{bbm}
\usepackage{xspace}
\usepackage{dsfont}
\usepackage{multirow}
\usepackage{wrapfig}
\usepackage[font=small,labelfont=bf,tableposition=top]{caption}

\DeclareCaptionLabelFormat{andtable}{#1~#2  \&  \tablename~\thetable}

\newcommand{\curves}{\textsc{curves}\xspace}
\newcommand{\greg}{\textsc{greg}\xspace}

\newcommand{\ap}{Suppl. Section~}

\usepackage{hyperref}

\usepackage[accepted]{icml2023}

\usepackage{amsmath}
\usepackage{amssymb}
\usepackage{mathtools}
\usepackage{amsthm}

\usepackage[capitalize,noabbrev]{cleveref}

\theoremstyle{plain}

\theoremstyle{definition}

\theoremstyle{remark}

\usepackage[textsize=tiny]{todonotes}

\icmltitlerunning{Emergence of Graphical Sensory-Motor Communication}

\begin{document}

\twocolumn[
\icmltitle{Contrastive Multimodal Learning for \\ Emergence of Graphical Sensory-Motor Communication}



\icmlsetsymbol{equal}{*}

\begin{icmlauthorlist}
\icmlauthor{Tristan Karch}{equal,inria,ub}
\icmlauthor{Yoann Lemesle}{equal,dauphine}
\icmlauthor{Romain Laroche}{micro}
\icmlauthor{Cl\'ement Moulin-Frier}{inria,ub}
\icmlauthor{Pierre-Yves Oudeyer}{inria,ub}


\end{icmlauthorlist}

\icmlaffiliation{inria}{Inria, Flowers Team}
\icmlaffiliation{dauphine}{Universit\'e Paris-Dauphine-PSL}
\icmlaffiliation{ub}{Universit\'e de Bordeaux}
\icmlaffiliation{micro}{Microsoft Research, Montreal}

\icmlcorrespondingauthor{Tristsan Karch}{tristan.karch@inria.fr}

\icmlkeywords{Machine Learning, ICML}

\vskip 0.3in
]



\printAffiliationsAndNotice{\icmlEqualContribution} 

\begin{abstract}

In this paper, we investigate whether artificial agents can develop a shared language in an ecological setting where communication relies on a \emph{sensory-motor channel}.
To this end, we introduce the Graphical Referential Game (\greg) where a speaker must produce a graphical utterance to name a visual referent object
while a listener has to select the corresponding object among distractor referents, given the delivered message. The utterances are drawing images produced using dynamical motor primitives combined with a sketching library. To tackle \greg we present \curves: a multimodal contrastive deep learning mechanism that represents the energy (alignment) between named referents and utterances generated through gradient ascent on the learned energy landscape. We demonstrate that \curves not only succeeds at solving the \greg but also enables agents to self-organize a language that generalizes to feature compositions never seen during training. In addition to evaluating the communication performance of our approach, we also explore the structure of the emerging language. Specifically, we show that the resulting language forms a coherent lexicon shared between agents and that basic compositional rules on the graphical productions could not explain the compositional generalization. 
\end{abstract}

\section{Introduction}
\label{sec:intro}

Understanding the emergence and evolution of human languages is a significant challenge that has involved many fields, from linguistics to developmental cognitive sciences~\cite{CHRISTIANSEN2003300}. Computational experimental semiotics~\cite{galantucci2011experimental} has seen some success in modeling the formation of communication systems in populations of artificial agents~\cite{cangelosi2002simulating, kirby2014iterated}. More specifically, \textit{Language Game} models~\cite{steels2012grounded}, have been used to show how a population of agents can self-organize a culturally shared lexicon without centralized coordination. Given the recent successes of artificial neural networks in solving complex tasks such as image classification~\cite{krizhevsky2012imagenet, he2015delving, he2016deep, dosovitskiy2021image} and natural language understanding~\cite{devlin2019bert, radford2019language, brown2020language}, many works have leveraged them to study the emergence of communication in groups of agents~\cite{lazaridou2020emergent}, mainly using multi-agent deep reinforcement learning and language games~\cite{nguyen2020deep, mordatch2018emergence, lazaridou2018emergence, portelance2021emergence, chaabouni2021communicating}. These advances have made it possible to scale up language game models to environments where linguistic conventions are jointly learned with visual representations of raw image perception, as well as to environments where emergent communication is used as a tool to achieve joint cooperative tasks~\cite{barde2022learning}. 

\begin{figure*}[t!]
\centering 
\includegraphics[width=0.95\textwidth]{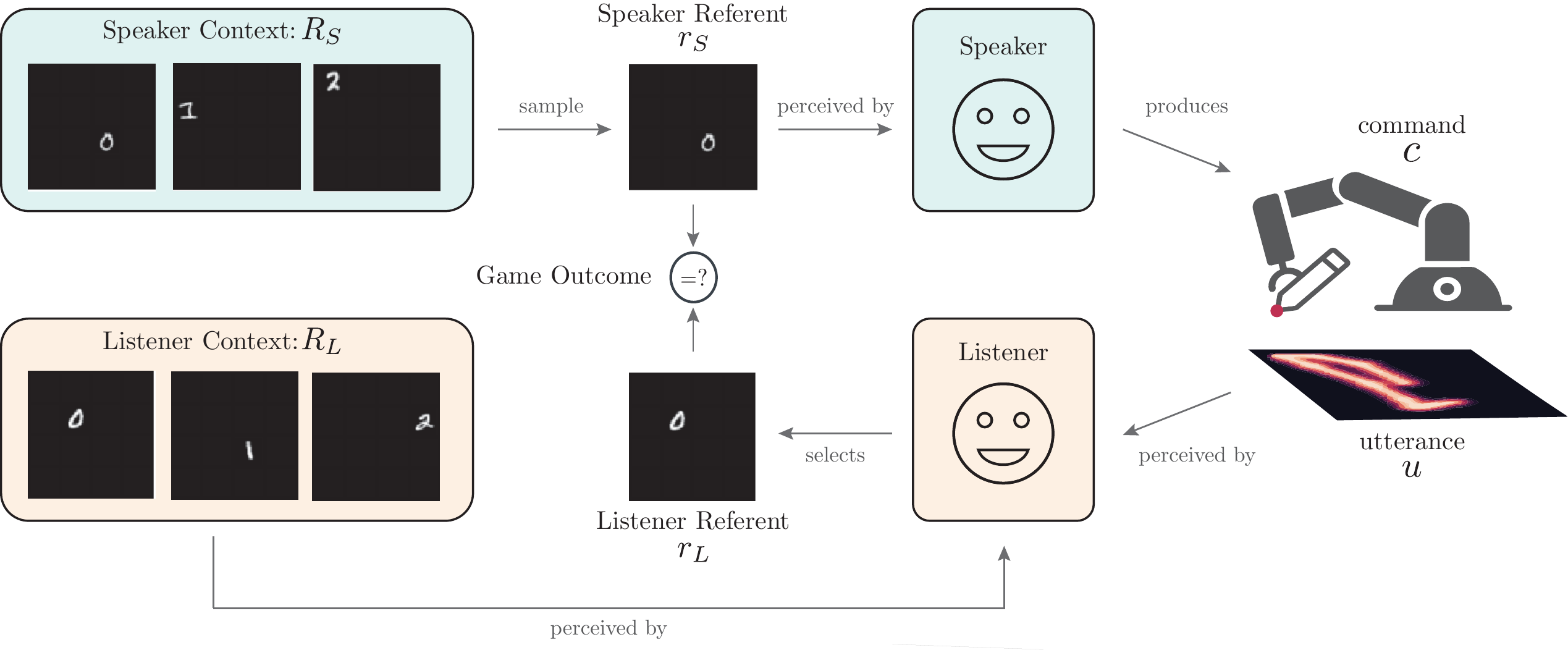}
\caption{\textbf{The Graphical Referential Game:} During the game, the speaker's goal is to produce a motor command $c$ that will yield an utterance $u$ in order to denote a referent $r_S$ sampled from a context $\tilde{R}_S$. Following this step, the listener needs to interpret the utterance in order to guess the referent it denotes among a context $\tilde{R}_L$. The game is a success if the listener and the speaker agree on the referent ($r_L\equiv r_S$).}
\label{fig:1}
\end{figure*}

So far, most of these methods have considered only idealized symbolic communication channels based on discrete tokens~\cite{lazaridou2017multiagent,mordatch2018emergence,chaabouni2021communicating} or fixed-size sequences of word tokens~\cite{havrylov2017emergence,portelance2021emergence}. This predefined means of communication is motivated by language's discrete and compositional nature. But how can this specific structure emerge during vocalization or drawing, for instance? Although fundamental in the investigation of the origin of language~\cite{Dessalles2000,cheney2005constraints,oller2019language}, this question seems to be neglected by recent approaches to Language Games~\cite{moulinfrier2020multi}. We, therefore, propose to study how communication could emerge between agents producing and perceiving continuous signals with a constrained \textit{sensory-motor system}. 

Such continuous constrained systems have been used in the cognitive science literature as models of sign production to study the self-organization of speech in artificial systems~\cite{deBoer2000selforganization,oudeyer2006selforganization,MOULINFRIER20155}. 
In this paper, we focus on a drawing sensory-motor system producing graphical signs. The sensory-motor system is made of Dynamical Motor Primitives (DMPs)~\cite{schaal2006dynamic} combined with a sketching system~\cite{Mihai2021DifferentiableDA} enabling the conversion of motor commands into images.  Drawing systems have the advantage of producing 2D trajectories interpretable by humans while preserving the non-linear properties of speech models, which were shown to ease the discretization of the produced signals~\cite{STEVENS19893,MOULINFRIER20155}. We introduce the \textit{Graphical Referential Game}: a variation of the original referential game, where a \textit{Speaker} agent (top of Figure~\ref{fig:1}) has to produce a graphical \textit{utterance} given a single target \textit{referent} while a \textit{Listener} agent (bottom of Figure~\ref{fig:1}) has to select an element among a context made of several referents, given the produced utterance (agents alternate their roles).  In this setting, we first investigate whether a population of agents can converge on an efficient communication protocol to solve the graphical language game. Then, we evaluate the coherence and compositional properties of the emergent language, since it is one of the main characteristics of human languages.

Early language game implementations~\cite{steels1995selforganizing,steels2001language} achieve communication convergence by using contrastive methods to update association tables between object referents and utterances. While recent works use deep learning methods to target high-dimensional signals they do not explore contrastive approaches. Instead, they model interactions as a multi-agent reinforcement learning problem where utterances are actions, and agents are optimized with policy gradients, using the outcomes of the games as the reward signal~\cite{lazaridou2017multiagent}. In the meantime, recent models leveraging contrastive multimodal mechanisms such as CLIP~\cite{radford2021learning} have achieved impressive results in modeling associations between images and texts. Combined with efficient generative methods~\cite{ramesh2021zero-shot}, they can compose textual elements that are reflected in image form as the composition of their associated visual concepts. Inspired by these techniques, we propose \curves: Contrastive Utterance-Referent associatiVE Scoring, an algorithmic solution to the graphical referential game. \curves relies on two mechanisms: 1) The contrastive learning of an energy landscape representing the alignment between utterances and referents and 2) the generation of utterances that maximize the energy for a given target referent. We evaluate \curves in two instantiations of the graphical referential game: one with symbolic referents encoded by one-hot vectors and another with visual referents derived from the multiple MNIST digits~\cite{LeCun1998GradientbasedLA}. We show that \curves converges to a shared graphical language that enables a population of agents not only to name complex visual referents but also to name new referent compositions that were never encountered during training.

\paragraph{Scope.} The idea of using a sensory-motor system to study the emergence of forms of combinatoriality in language dates back to methods investigating the origins of digital vocalization systems~\cite{deBoer2000selforganization,oudeyer2005selforganization,zuidema2009evolution}. Such studies were conducted in the context of imitation games at the level of phonemes to observe the formation of speech utterances (syllables, words) that were systematically composed from lower-level meaningless elements (phonemes). This corresponded to the first level of compositionality within the notion of duality of patterning~\cite{hockett1960origin}. Yet, these works did not consider referential games and did not study agents' ability to compose meaningful words to denote referents, i.e. they did not address the second level of the duality of patterning. 

One of the goals of emergent communication research is to develop machines that can interact with humans. As a result, a variety of referential game approaches ensure that the emergent language is as close to natural language. This can be achieved by adding a supervised image captioning objective to encourage agents to use natural language in order to solve their communicative tasks~\cite{havrylov2017emergence,lazaridou2017multiagent}. Other methods use constraints such as memory restrictions~\cite{kottur2017natural} to act as an information bottleneck to increase interpretability and compositionality. While we purposefully chose a graphical sensory-motor system to ease the visualization of the emerging language, we do not inject prior knowledge or pressures to facilitate the emergence of an iconic language. Our produced utterances are completely arbitrary. This fundamentally differentiates our work from ~\citet{mihai2021learning} that trains agents to communicate via sketches replicating the visual referents they name. Note also that their drawing setup does not include dynamical motor primitives and utterances are directly optimized in image space. They, moreover, allow gradients to back-propagate from listener to speaker while we use a decentralized approach. Finally, they do not consider contrastive learning. To our knowledge, \curves is the first contrastive deep-learning algorithm successfully applied to a referential game.

There is a large body of work exploring the factors that promote
compositionally in emerging languages~\cite{kottur2017natural,li2019ease,rodriguez-luna-etal-2020-internal,Ren2020Compositional,chaabouni2020compositionality,gupta-etal-2020-compositionality}. In this context, a crucial question is how to actually measure it in the first place~\cite{mu2021emergent}. To this end, \cite{choi2018compositional} proposes to measure communicative performances on unseen compositions of known objects as a way to evaluate compositionality. However, it has been shown that a good performance in this test may be achieved without leveraging any actual compositionality in language~\cite{andreas2019measuring, chaabouni2020compositionality}. Thus, others instead compute topographic similarities \cite{brighton2006understanding}, measuring the correlation between distances in the utterance space (distance between signs) and distances in the referents space (such as the cosine similarity between the embeddings of objects)~\cite{lazaridou2018emergence}. In this paper we propose to do both and study 1) the generalization to unseen combinations of abstract features and 2) topographic measures based on the Hausdorff distances between utterances denoting composition and utterances denoting isolated features. 

\textbf{Contributions.} This paper introduces:
\begin{itemize}[noitemsep,topsep=0pt]
    \item The Graphical Referential Game (\greg): a variation of the referential language game to study the formation of signs from a graphical sensory-motor system.
    \item \curves: an algorithmic solution to \greg, consisting of a contrastive multimodal encoder coupled with a generative model enabling the emergence of a graphical language.
    \item A study of \curves's generalization performances on compositions of features never seen during training in a simplified control setting and a more perceptually challenging one.
    \item A complementary analysis of the structure of the emerging graphical language measuring lexicon coherence and compositionality scores derived from the Haussdorf distance.
\end{itemize}

\section{Problem Definition}
\label{sec:prob_def}

\begin{figure*}[t]
\centering 
\begin{tabular}{cc}
\includegraphics[width=0.56\textwidth]{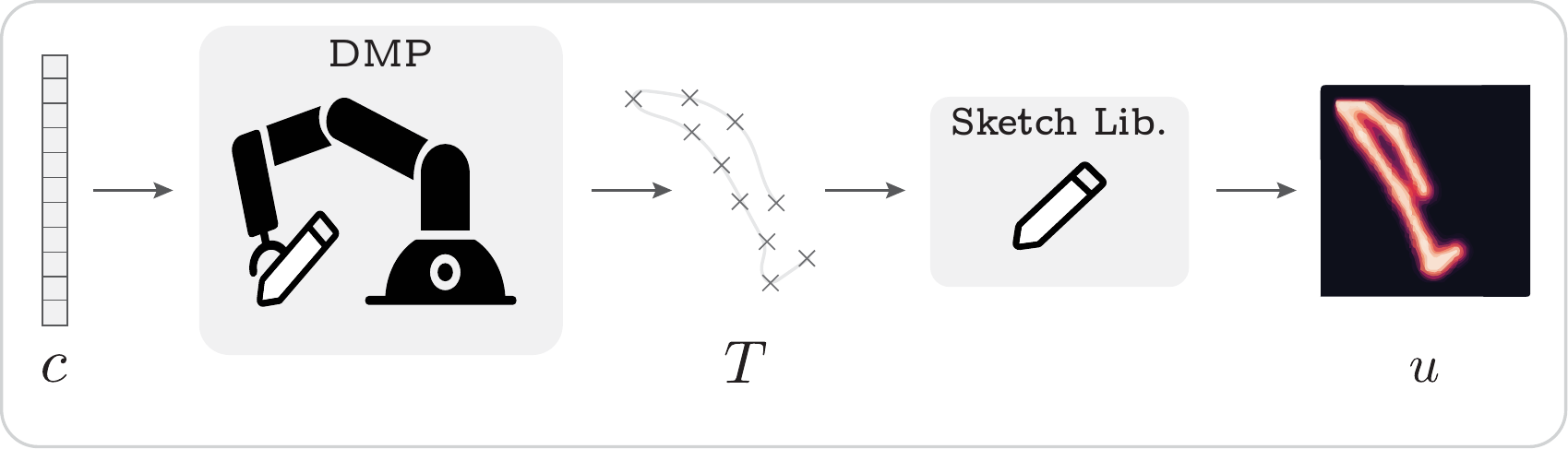} &  \includegraphics[width=0.42\textwidth]{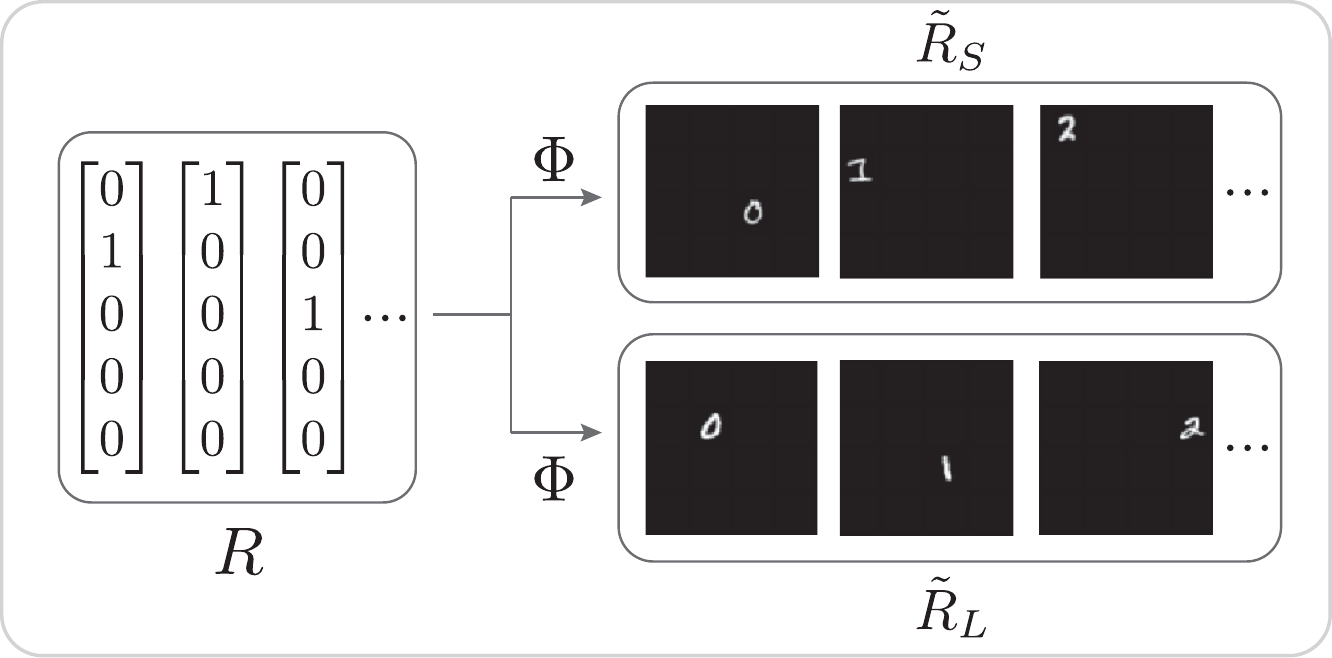}\\
(a) & (b)
\end{tabular}
\vspace{-0.2cm}
\caption{(a) \textbf{Sketching sensory-motor system}: The sensory-motor system imitates a robotic arm drawing a sketch on a 2D plan. DMPs first convert a continuous command $c$ into a sequence of coordinates $T$. This trajectory is then rendered as a $52\times52$ graphical utterance thanks to a differentiable sketching library. (b) \textbf{Referent transformation:} An example of a one-hot context $R$ being transformed into two contexts $\tilde{R}_S$ and $\tilde{R}_L$ by the stochastic transformation $\Phi$. The two contexts are different perspectives of the same objects.}
\label{fig:prob_def}
\end{figure*}

\paragraph{Graphical referential game.} 

We consider a group of two agents playing a fixed number of referential games, each time alternating their roles (speaker or listener). During a game, we first present a context $R$ of $n$ objects, called referents to a speaker $S$ and a listener $L$. At the beginning of each game, the target $r^\star \in R$ is assigned to the speaker. Given this target referent $r^\star$, $S$ produces an utterance ($u$) to designate it. Based on the produced utterance $u$, $L$ selects a referent ($\hat{r}$) in $R$. The game outcome $o$ is a success if the selected referent ($\hat{r}$) matches the target $r^\star$. 

\paragraph{Referents.}
Referents are compositions of orthogonal vector features (one-hot vectors). Given a set of $m$ orthogonal features $F_m$, we define the set of all possible referents as ${\mathcal{R}_m = \{ \textstyle\sum_{f \in S} f | S \subseteq F_m \}}$. The subset of referents made of exactly $k$ features are thus: ${\mathcal{R}^k_m=\{ \textstyle\sum_{f \in S} f | S \subseteq F_m, |S| = k\}}$. In our experiments, we fix $m=5$.

From these orthogonal referents, we propose to generate objects made of digit images sampled from the MNIST dataset~\cite{LeCun1998GradientbasedLA}. More precisely, we define the stochastic mapping $\Phi: \mathcal{R}_m \rightarrow \tilde{\mathcal{R}}_m$ that maps each feature $f \in F_m$ to a digit class in the MNIST dataset. For each feature in a referent, we sample a random instance from the corresponding class and randomly place it on a $4\times4$ grid such that no number overlap. Note that the listener and speaker can perceive different realizations of $\Phi$, in this case, we say that they see different \textit{perspectives} of the referents. More precisely, the speaker perceives the context $R$ as $\tilde{R}_S$ and its target $r^\star$ as $r^\star_S$. Similarly, the listener perceives the context $R$ as $\tilde{R}_L$ and selects a referent $\hat{r}$ among it.
\newpage
We use this formalism to instantiate three settings of the Graphical Referential Game (\greg):
\begin{itemize}[noitemsep,topsep=0pt]
    \item \textit{one-hot}: where referents are one-hot vectors $r \in \mathcal{R}_m$.
    \item \textit{visual-shared}: where referents are MNIST digits $r \in \tilde{\mathcal{R}}_m$ and agents share the same perspective: $\tilde{R}_S = \tilde{R}_L$.
    \item \textit{visual-unshared} where referents are MNIST digits $r \in \tilde{\mathcal{R}}_m$ and agents have different perspectives of referents in their contexts $\tilde{R}_S \neq \tilde{R}_L$.
\end{itemize}

\paragraph{Sensory-motor drawing system.} 
Utterances are produced by a sensory-motor system ${M: \R^m \rightarrow \mathcal{U} \subset \R^{D\times D}}$ mimicking an arm drawing sketches displayed in Figure~\ref{fig:prob_def}(a). The arm motion is derived from Dynamical Motor Primitives (DMPs)~\cite{schaal2006dynamic}. The DMP is parametrized by a command vector $c\in\R^{20}$. It converts $c$ into a 2-dimensional drawing trajectory $T$ made of 10 coordinates $T=\{v_i\}_{i=0,...,9}$. This trajectory is then fed to a Differentiable Sketching model~\cite{Mihai2021DifferentiableDA} generating an $D\times D$ image (in our implementation, $D=52$). See \ap\ref{sec:suppl_sensorymotor} for additional implementation details of the Sensory-motor drawing system.

\paragraph{Objectives.} In this study, we aim to answer the three following questions:
\begin{enumerate}[noitemsep,topsep=0pt]
\item What are agents' communicative performances in the \greg? Are agents able to solve the game? Are they able to generalize to compositional referents?
\item Are the emergent signs coherent? Do agents produce the same utterances to denote the same referents?
\item Are the emergent signs compositional? Are there compositional rules in the production of signs naming compositional referents? \footnote{Note that the ability to perform compositional generalization (question 1) and the presence of compositional structure in utterances (question 3) are two separate investigations.}
\end{enumerate}

\textit{Are agents able to solve the \greg? } To answer the first question,  we will monitor the communicative
performance of agents on both training and testing referents.  The training referents consist of a single feature: $\mathcal{R}_{\text{train}} = \mathcal{R}_5^1$ while the testing referents consists of two features: $\mathcal{R}_{\text{test}}=\mathcal{R}_5^2$. For visual examples of compositional referents, see~ \ap\ref{sup:ref_comp}.

\textit{Are the emergent signs coherent? } To measure coherence we propose to use a similarity measure based on the Hausdorff distance. Haussdorf distance is known to capture geometric features of trajectories, in particular, their shape~\cite{Besse2015review}. The Hausdorff distance $d_H$ is the maximum distance from any coordinate in a trajectory to the closest coordinate in the other:
$d_H(T_1,T_2) = \max\{\sup_{v \in T_1} d(v,T_2), \sup_{v' \in T_2}d(T_1,v') \}$.
In particular, we compute the following metrics.
\begin{itemize}[noitemsep,topsep=0pt]
    \item Agent Coherence (A-coherence): For a given referent $r$ with the same perspective for all agents, measure the mean pairwise similarity between each agent's utterance.
    \item Perspective Coherence (P-coherence): For a given agent and a given referent $r$, measure the mean pairwise similarity between utterances produced from different perspectives 
    \item Referent Coherence (R-coherence): For a given agent, measure the mean pairwise similarity between utterances produced for different referents.
\end{itemize}

\textit{Are the emergent signs compositional? } To measure the compositionally of the utterances, we introduce a topographic score based on the Hausdorff distance $\rho$. $\rho$ quantifies how an utterance denoting a compositional referent made of feature $i$ and $j$ ($u(r_{ij})$) is actually closer to the utterances denoting isolated features $u(r_i)$ or $u(r_j)$ than the utterance naming other compositional referents ($u(r_{xy})$, $x\neq i, y\neq j$). For a detailed derivation of metric $\rho$, see \ap\ref{sup:topo_comp}.

\newpage

\section{CURVES - Contrastive Utterance-Referent associatiVE Scoring}

\begin{figure*}[t!]
\centering 
\includegraphics[width=0.7\textwidth]{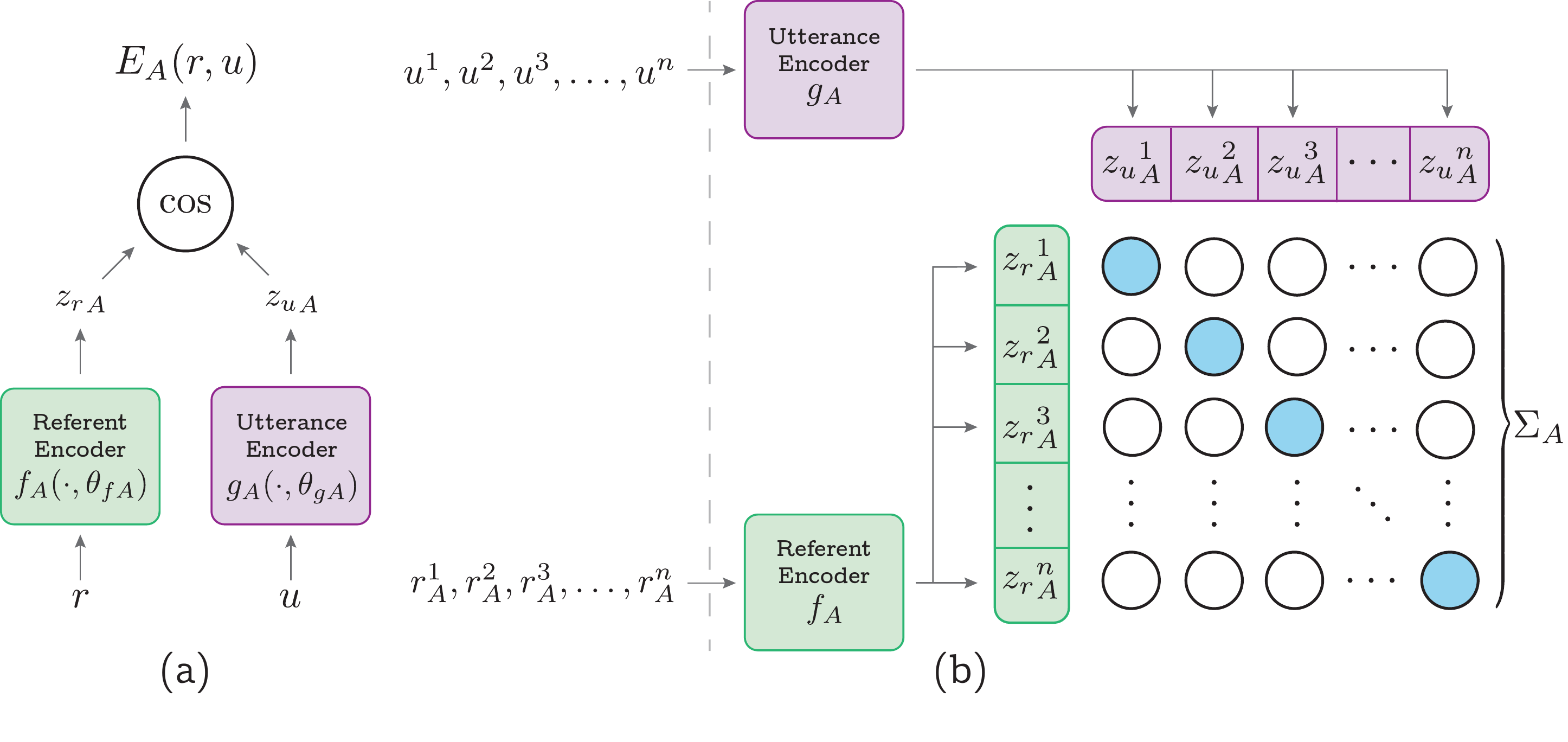}
\caption{(a) \textbf{Agents's dual encoder architecture.} Referents and utterances are mapped to a share latent space. The energy between a referent $r$ and an utterance $u$ is computed as the cosine similarity between their respective embeddings. (b) \textbf{Cosine similarity matrix update from collected samples.} Agents compute the energy for all referents and utterances they collected to form the squared matrix $\Sigma_A$. During contrastive updates agents maximize blue circles and minimize white ones.}
\label{fig:curves_encoders}
\end{figure*}

\curves is an energy-based approach that relies on two mechanisms:
\begin{enumerate}[noitemsep,topsep=0pt]
\item  The contrastive learning of an energy landscape $E(r,u)$, defined as the cosine similarity between utterance and referent embeddings.
\item The generation of an utterance that maximizes the energy for a given target referent $r^\star_S$.
\end{enumerate}

\textbf{Agents modules and interactions. } Each agent $A\in\{A_1,A_2\}$ perceives utterances and referents using two distinct CNN encoders $f_A$ (for referents) and $g_A$ (for utterances)\footnote{when referents are one-hot vectors $f_A$ is a fully-connected network. Parameters for both encoders are given in Suppl. table~\ref{table:hyperparameters}.}. $f_A$ and $g_A$ map referents and utterances in a shared $d$-dimensional latent space: $f_A(\cdot,\theta_{fA}): \mathcal{R}_m \rightarrow \mathbb{R}^d$ and $g_A(\cdot,\theta_{gA}): \mathcal{U} \rightarrow  \R^d$ such that $z_{rA}=f_A(r)$ and $z_{uA}=g_A(u)$, as displayed in Figure~\ref{fig:curves_encoders}(a). The agent then computes the energy landscape as: $E_A(r,u) = \cos(f_A(r),g_A(u))$.

A given referential game unfolds as follows. Agents have randomly attributed roles, for instance, $A_1$ is the speaker $A_1\leftarrow S$ and $A_2$ is the listener $A_2 \leftarrow L$. The speaker is given a context $\tilde{R}_S$ and a target referent perceived as $r^\star_S$ to produce an utterance $\hat{u}$ intending to approach the utterance $u^\star$ that maximizes $E_S(r^\star_S,u)$. The listener observes $\hat{u}$ and selects referent $\hat{r}$ in context $\tilde{R}_L$ that maximizes $E_L=(r,\hat{u})$:
\begin{equation}
\left\{
\begin{split}
    \hat{u} &\approx u^\star= \underset{u \in \mathcal{U}}{\textrm{argmax }} E_S(r^\star_S,u) \\
    \hat{r} &= \underset{r \in \tilde{R}_L}{\textrm{argmax }} E_L(r,\hat{u}) \end{split}
\right.
\label{eq:ut_gen_ref_sel}
\end{equation}
The outcome of the game is then $o = \mathbbm{1}_{[\hat{r}=r^\star]} - b$ where $b$ is a baseline parameter representing the mean success across previous games.

\textbf{Contrastive representation learning in referential games. } 
For a given context $R$, agents are randomly assigned their roles and play $n=|R|$ games. During these $n$ games, roles are fixed and the speaker agent successively selects each referent of the context $\tilde{R}_S$ as the target $r^\star_S$. During interactions, the speaker collects data $\{(r_S^i, u^i, o^i)\}_{i=1,...,n}$ while the listeners observes $\{(u^i, r_L^i)\}_{i=1,...,n}$. From the collected data each agent can compute the squared cosine similarity matrices $\Sigma_A$ whose elements are $(\Sigma_A)_{i,j} = E_A(r_A^i, u^j)$ as shown in Figure~\ref{fig:curves_encoders}(b). Contrastive updates are then performed using the objective $J_A$ that applies \textit{Cross Entropy} ($CE$) on the $i$-th row and $i$-th column of $\Sigma_A$.
\begin{equation}
\small
        J_A(\Sigma_A,i) = \frac{CE((\Sigma_A)_{i,1:n},e_i) + CE((\Sigma_A)_{1:n,i},e_i)}{2}\\    
\end{equation}
$e_i$ being a one-hot vector of size $n$ with value 1 at index $i$. Depending on the role of the agent, $J_A$ is instantiated either as $J_S$ (speaker) or $J_L$ (listener). Thus, the speaker updates its representation using the outcomes $o_i$ of the games (reinforcing the successful associations while decreasing the unsuccessful ones):
\begin{equation}
        \underset{\theta_{f_S}, \theta_{g_S}}{\textrm{minimize } } \overset{n}{\underset{i = 1}{\sum}} o_i  J_S(\Sigma_S,i)
\end{equation}
On the other hand, the listener needs to make sure that the selection matches the speaker's referent
\cite{steels2015talkingheads} and hence always increases associations (no matter the games' outcomes):
\begin{equation}
    \underset{\theta_{f_L}, \theta_{g_L}}{\textrm{minimize } } \overset{n}{\underset{i = 1}{\sum}} J_L(\Sigma_L,i)
    \label{eq:listener_loss}
\end{equation}
Note that in Eq.~\ref{eq:listener_loss}, $r_L^i$ is the target referent perceived by the listener. This means that, at the end of the game, the speaker indicates the referent (as perceived by the listener) that they named. This retroactive pointing mechanism was employed in both early language game implementations~\cite{steels1999situated} and more recent ones~\cite{chaabouni2020compositionality,portelance2021emergence}. 


\textbf{Speaker's utterance optimization. } We distinguish two utterance generation strategies:
\begin{itemize}[noitemsep]
    \item The descriptive generation: in which the speaker agent only considers the target referent $r_S^\star$ to produce an utterance that maximizes the cosine similarity between the embeddings of $r_S^\star$ and an utterance produced by our sensory system $u = M(c)$ from motor command $c$. Since $M$ is fully differentiable, we inject the sensory-motor constraint in equation~\ref{eq:ut_gen_ref_sel} and seek for the optimal motor command $c^\star$ using gradient ascent:
\begin{equation}
c^\star = \underset{c \in \mathbb{R}^p}{\textrm{argmax }} E(r^\star_S, M(c))
\label{eq:descri_gen}
\end{equation}
    \item The discriminative generation: in which the speaker also perceives the context $\tilde{R}_S$ during production. This is achieved by finding the motor command that minimizes the cross entropy given a target referent $r^\star_S$ and its context $\tilde{R}_S$:
\begin{equation}
    c^\star = \underset{c\in \R ^p}{\textrm{argmin }} CE( \sigma_S, e_{r^\star_S})
    \label{eq:discri_gen}
\end{equation}
where $\sigma_S$ is the vector with coordinates $\sigma_{Si} = [E(r^i, M(c))]_{r^i\in \tilde{R}_S}$ and $e_{r^\star_S}$ is the one-hot vector of size $|\tilde{R}_S|$ with value 1 at the position of $r^\star_S$ in $\tilde{R}_S$. This discriminative generation process is only used at test time when investigating \curves's generalization capabilities. 
\end{itemize}

\section{Experiments and Results}

This section focuses first on \curves's communicative performances when agents interact in \greg (question 1 of the objective paragraph in Section~\ref{sec:prob_def}).  Training and testing metrics correspond to the mean and standard deviation computed from training pairs of agents on 10 seeds. We then analyze the properties of the emergent language, namely its coherence and compositional nature (questions 2 and 3 of the objective paragraph in Section~\ref{sec:prob_def}). Studies are carried out with one-hot, shared-visual, and unshared-visual referents.

\subsection{\curves communicative performance}
\label{sec:results_perf}
\textbf{Training performances. } In all three settings of the Graphical Referential Game (one-hot, visual-shared, and visual-unshared), agents succeed and achieve a perfect training success rate of 1.

\textbf{Generalization to compositional referents. } Table~\ref{tab:generalization_perf} exposes the generalization performances of agents evaluated on referents $r \in \mathcal{R}_5^2$. During an evaluation, the context is exhaustive and contains all the combinations of 2 features: $|R|=10$. We compare the success rates to a \textit{random} baseline where the listener always selects the referent $\hat{r}_L$ randomly no matter the utterance ($\textsc{sr}_{\text{random}}=0.1$). We also introduce a \textit{1-feature} baseline where the speaker produces an utterance $u$ that only denotes one of the two features contained in $r_S^\star$ and the listener randomly selects one of the four combinations containing the communicated feature ($\textsc{sr}_{\text{1-feat}}=0.25$). 

\begin{table}[!h]
\small
    \centering
    \begin{tabular}{|c|c|c|}
        \hline
        \textbf{Referents} & \textbf{Descriptive} \textsc{SR} & \textbf{Discriminative} \textsc{SR}  \\
        \hline
         One-hot & $0.99\pm0.01$ & $0.99\pm0.01$ \\
         Visual-shared & $0.57\pm0.04$ & $0.56\pm0.03$ \\
         Visual-unshared & $0.39\pm0.02$ & $0.40\pm0.02$ \\
         \hline
    \end{tabular}
    \caption{\textbf{Generalization performances. } Success rates evaluated on exhaustive context $|R|=10$ with referents $r \in \mathcal{R}_5^2$ for both generative (Eq.~\ref{eq:descri_gen}) and discriminative (Eq.\ref{eq:discri_gen}) utterance generation.}
    \label{tab:generalization_perf}
\end{table}

The success rates for all referent types are significantly higher than the baseline values suggesting that agents are indeed able to communicate about compositional referents. Generalization performances are nearly perfect with one-hot referents but they decrease in visual settings. This performance gap can be explained by the extra difficulty of adding inter-perspective variability to the multi-agent interaction dynamic during the contrastive learning of referent representations. The better success rates obtained in auto-learning (where a single agent plays both the speaker and the listener roles) provided in \ap\ref{sup:auto_social_perf} seem to corroborate this hypothesis. Surprisingly, we observe that success rates for descriptive (Eq.~\ref{eq:descri_gen}) and discriminative (Eq.\ref{eq:discri_gen}) generation are very similar. This suggests that optimizing utterances so as to minimize their energy between non-targeted compositional referents ($r \in R, r \neq r^\star$) does not improve generalization performances.

\subsection{Analysis of the structure of the emergent language}

\begin{figure*}[!h]
    \centering
    \begin{tabular}{@{}c@{}c@{}c@{}}
    \includegraphics[width=0.3\textwidth]{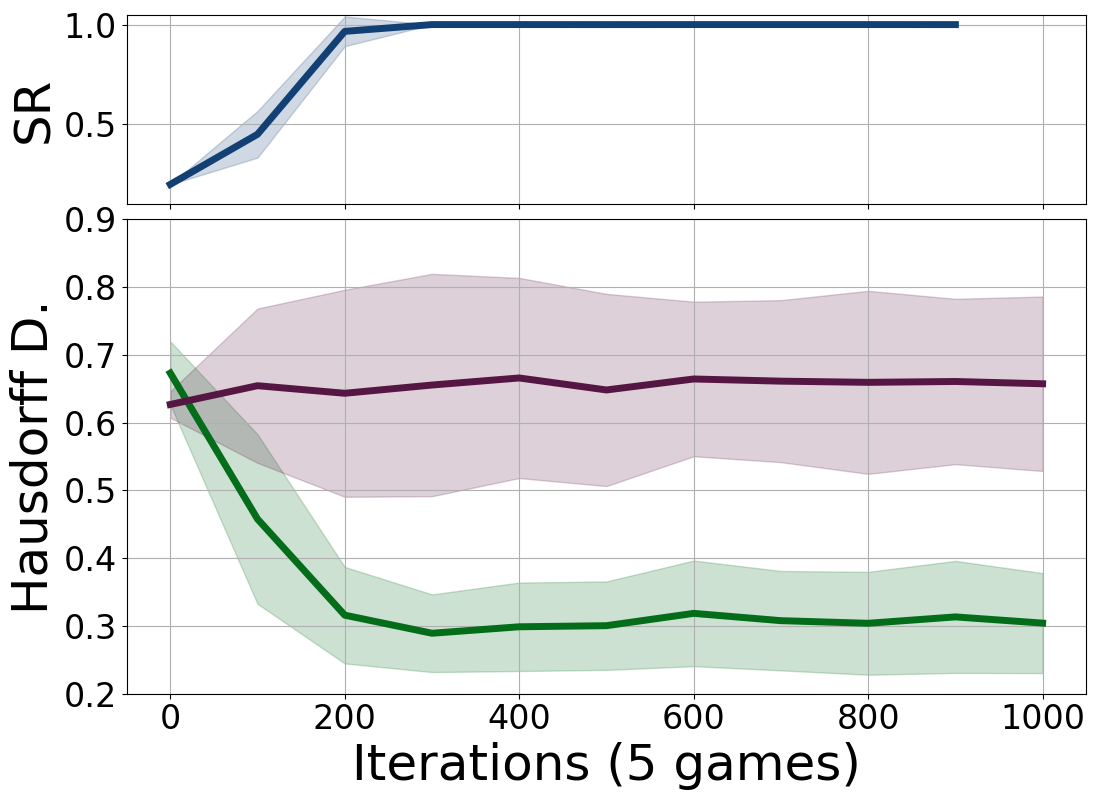} &  \includegraphics[width=0.3\textwidth]{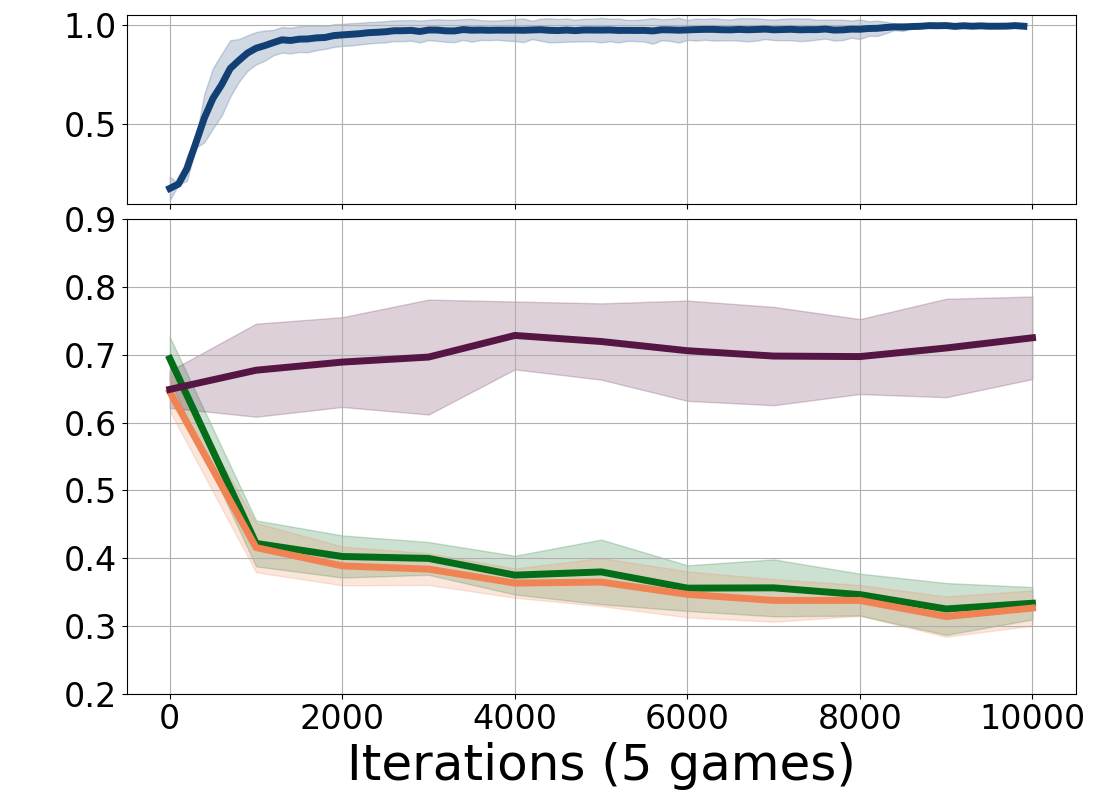} &
    \includegraphics[width=0.3\textwidth]{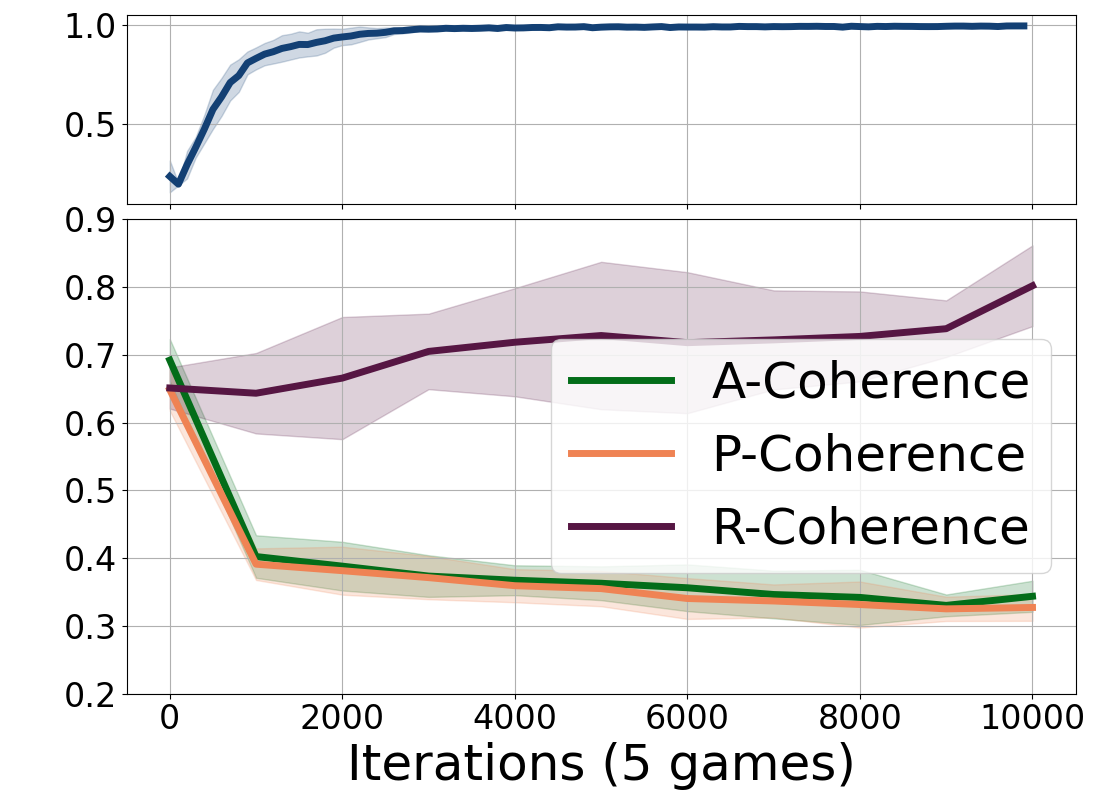}\\
    (a) & (b) & (c)
    \end{tabular}
    \caption{\textbf{Training success rate (SR) and Coherence distances} (a) one-hot referents (b) visual-shared referents (c) visual-unshared referents.}
    \label{fig:conv_coher}
\end{figure*}

\textbf{Coherence. }  Figure~\ref{fig:conv_coher} displays the evolution of the inter-agent (A), inter-perspective (P), and inter-referent (R) coherence during training. A group starts to converge and succeed at the game when inter-agent and inter-perspective coherence distances decrease. This correlation is proof of emergent communication as it indicates that agents start agreeing on signs to denote referents. The constant (for one-hot referent) and increasing (for visual referents) values of the R-coherence suggest that agents use distinct signs to name referents.

\begin{figure}[!h]
    \centering
    \small
    \includegraphics[width=0.5\textwidth]{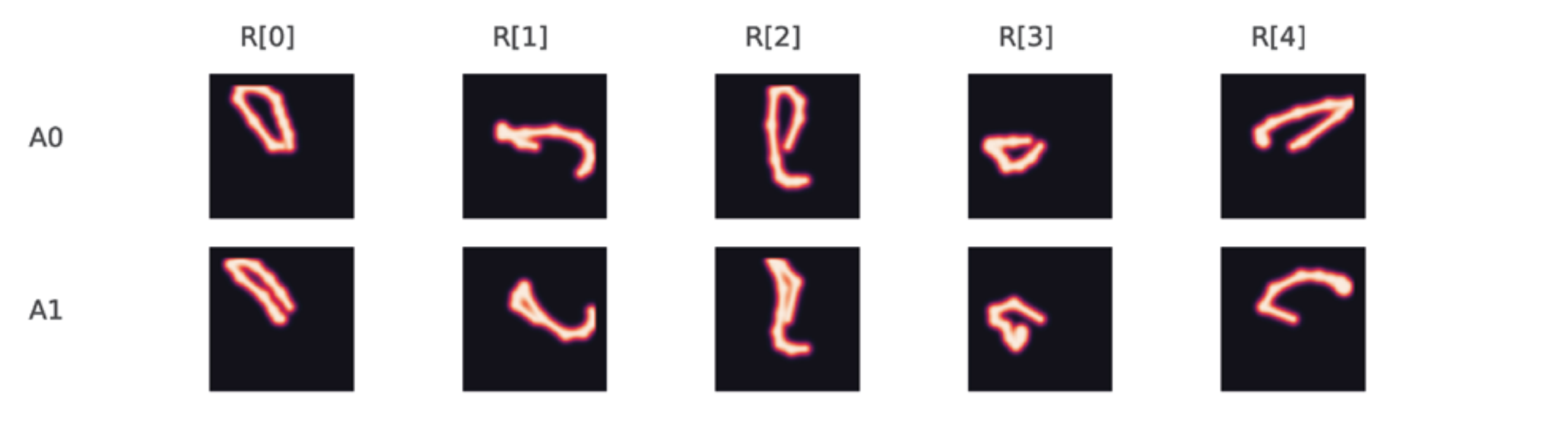}
    \vspace{-.2cm}
    \caption{\textbf{Instance of an emerging lexicon.} Utterances are produced by a pair of agents trained with unshared perspectives (1 seed). The perspective for each referent is chosen randomly.}
    \label{fig:lexicon_example}
\end{figure}

As displayed in Figure~\ref{fig:lexicon_example}, the language used by agents self-organizes around five distinct symbols. It is important to note that this self-organization arises from the production of continuous signals with no explicit communication of the five categories of visual referents. Other visualizations for one-hot and shared visual referents are available in \ap\ref{sup:lexicon_one_hot_shared}. We also provide illustrations of P-coherence in \ap\ref{sup:P-coherence}.

\begin{figure}[!h]
    \centering
    \small
    \begin{tabular}{c}
    \includegraphics[width=0.26\textwidth]{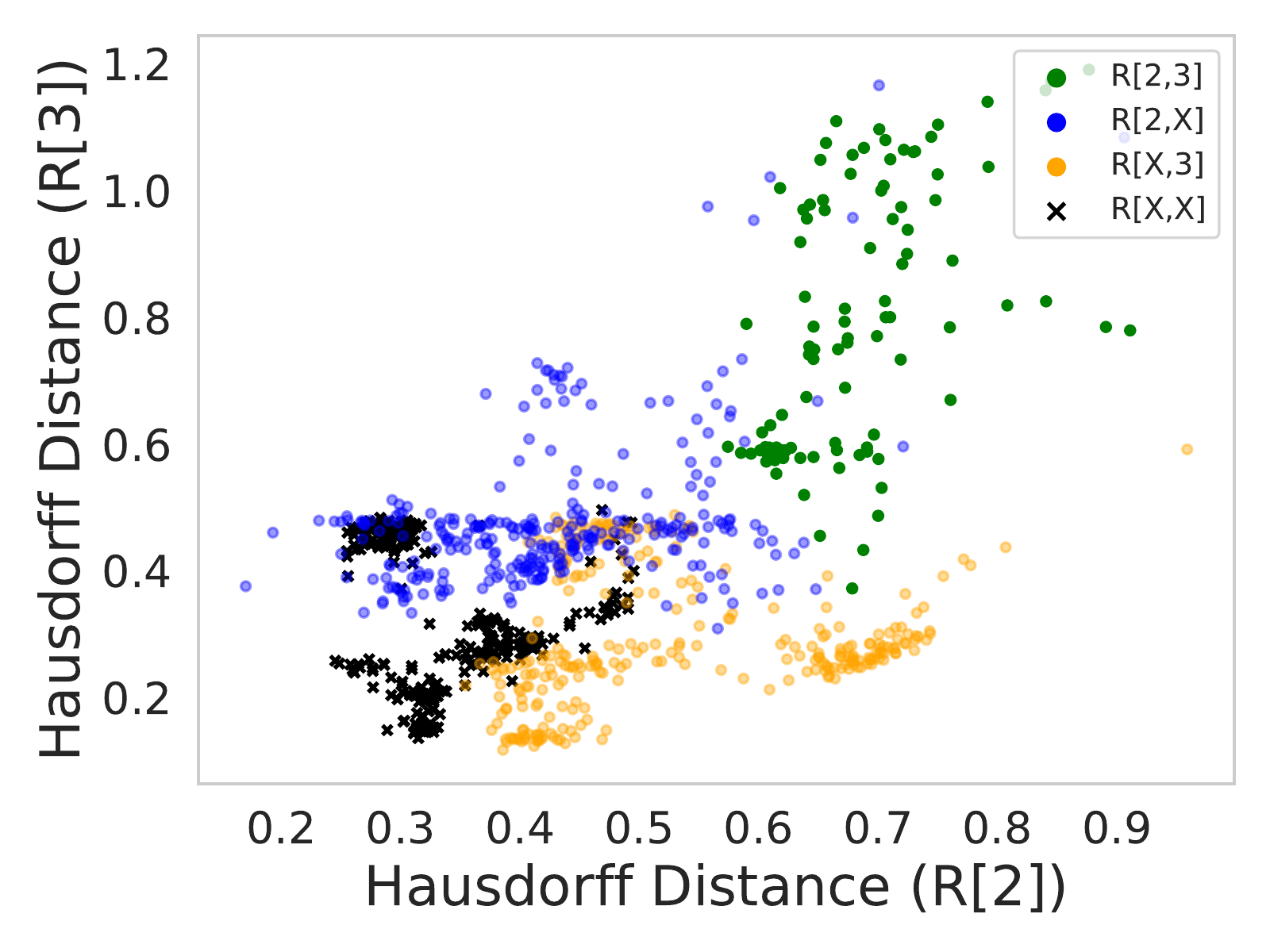} \\ 
    (a) ${\rho=-0.401}$ \\
    \includegraphics[width=0.26\textwidth]{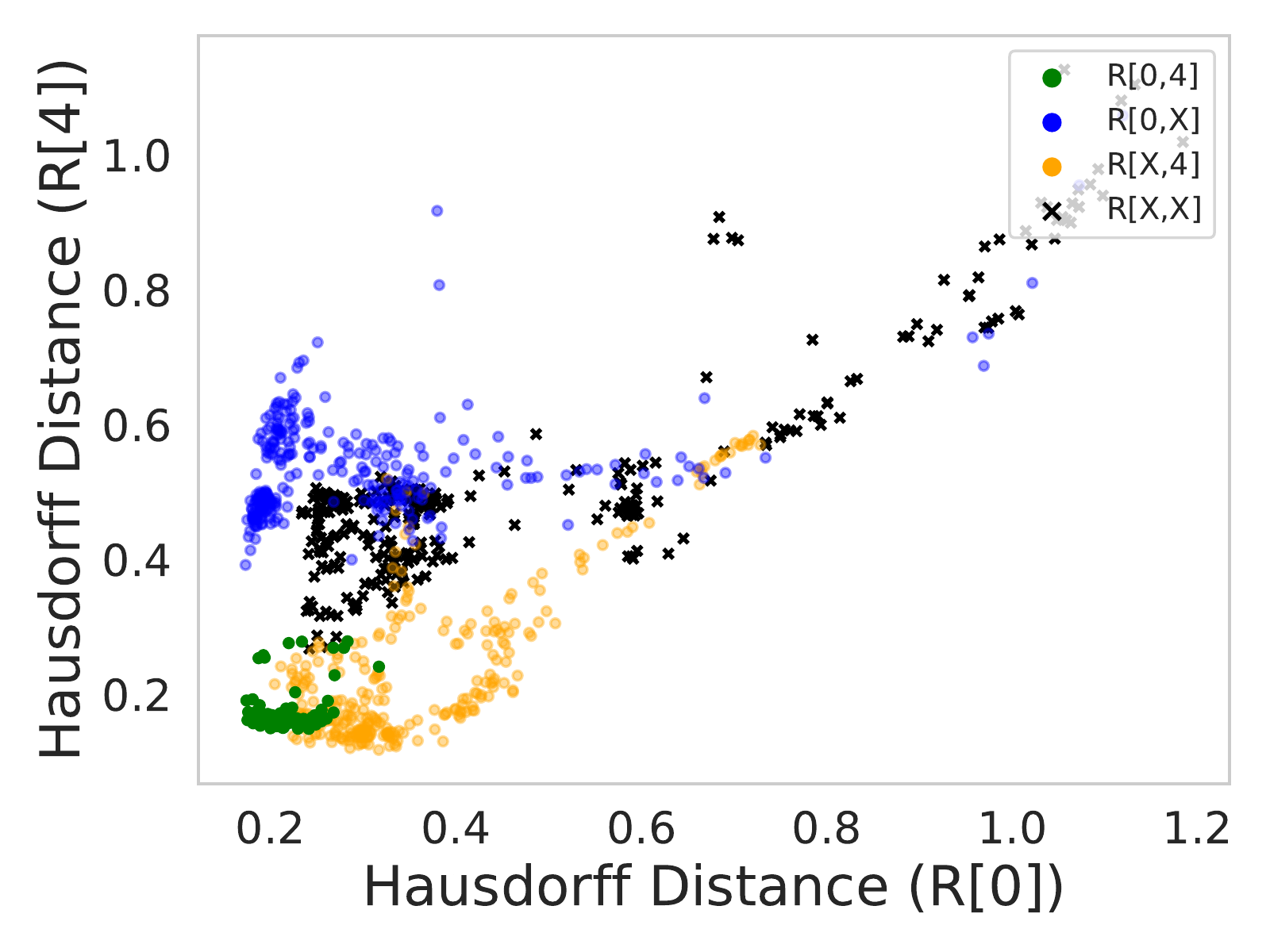} \\
     (b) ${\rho=0.147}$ 
    \end{tabular}
    \vspace{-0.2cm}
    \caption{\textbf{Topographic map examples for a single seed in one-hot referents setting}. Each utterance names a compositional referent and is colored in blue if it contains feature $i$ ($R[i, X]$), orange if it contains feature $j$ ($R[X, j]$), green if it contains both ($R[i, j]$), and black if it contains none ($R[X, X]$). (a) worst topographic score ${\rho=-0.401}$ ($i=2$ and $j=3$) (b) best topographic score $\rho=0.147$ ($i=0$ and $j=4$). }
    \label{fig:topo_maps}
\end{figure}

\begin{figure}[!h]
\centering
\includegraphics[width=0.25\textwidth]{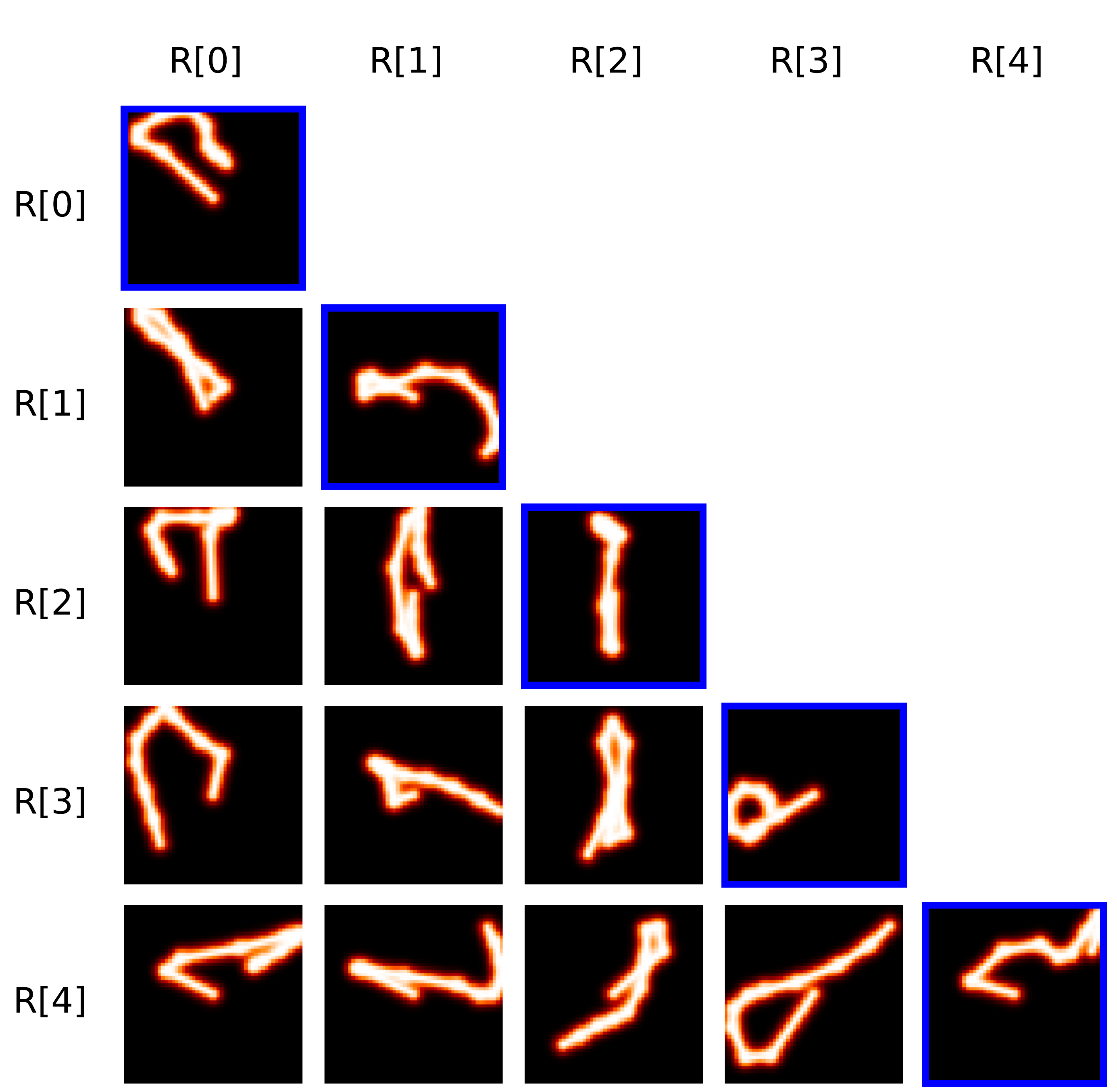}
\vspace{-0.2cm}
\caption{\textbf{Matrix of compositions. }Blue frames represent utterances for a perspective in $\mathcal{R}_5^1$, others for perspectives in $\mathcal{R}_5^2$ }
\label{fig:compo_matrix}
\end{figure}

\paragraph{Compositionality.}  In Section~\ref{sec:results_perf}, we showed that agents achieve a near-perfect success rate at naming compositions of one-hot features at test time. Is this successful communication reflected by a compositional structure in the produced signs? To investigate this question we propose the topographic maps associated with their topographic scores in Figure~\ref{fig:topo_maps}. Each point in a topographic map is an utterance naming a compositional referent $r\in \mathcal{R}_5^2$ and has coordinate $(d_H(u(r_i), \cdot), d_H(u(r_j), \cdot))$. Utterances at the bottom left of the topographic maps are therefore simultaneously close to the two utterances naming the isolated features. All the topographic maps are available in Figure~\ref{fig:sup_topo_one_hot} of \ap\ref{sup:topo_maps}. They show that for a minority of compositions (3 out of 10), the utterances naming the composition of two features are not close in Haussdorf distance to the utterances naming the two isolated features ($\rho < 0$).  This indicates that proximity in Haussdorf distance is not a necessary condition for agents to generalize on compositional referents. The matrix of composition provided in figure~\ref{fig:compo_matrix} illustrate that it is indeed very difficult to infer a composition rule from the generated utterances. 

\newpage

\begin{figure}[!h]
    \centering
    \includegraphics[width=0.48\textwidth]{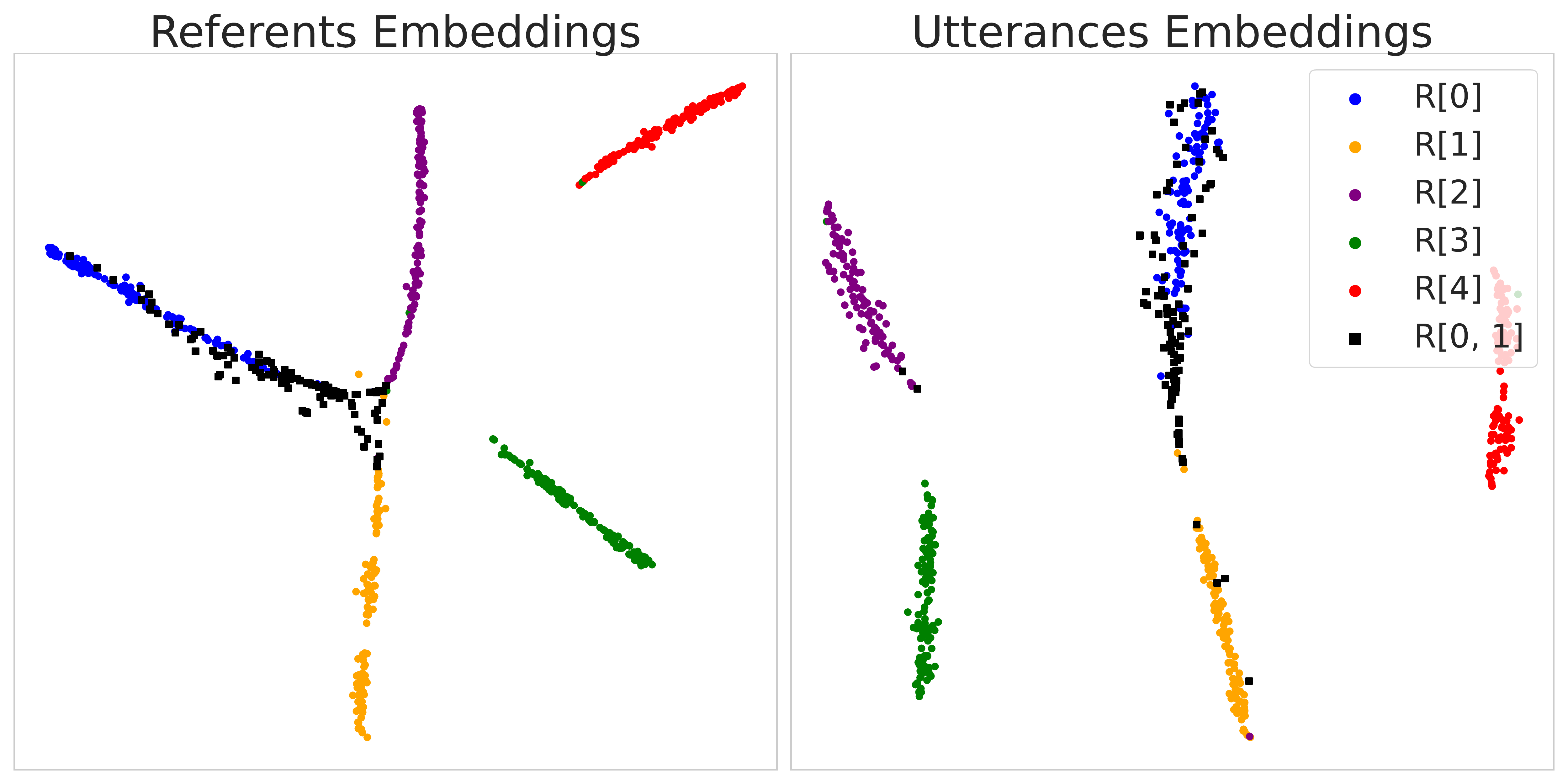}
    \caption{\textbf{T-sne of utterance and referent embeddings.} Embeddings are computed for 100 perspectives in the visual-unshared setting. Additional t-snes are provided in \ap\ref{sup:tsnes}.}
    \label{fig:tsne}
\end{figure}

Despite the fact that we cannot perceive the compositional structure of emerging signs, the internal representations of agents seem to leverage compositional mechanisms. The t-snes provided in Figure~\ref{fig:tsne} shows that the embeddings for both compositional referents and the utterances naming them are close to their constituents. 

\begin{figure}[!h]
\centering
 \includegraphics[width=.45\textwidth]{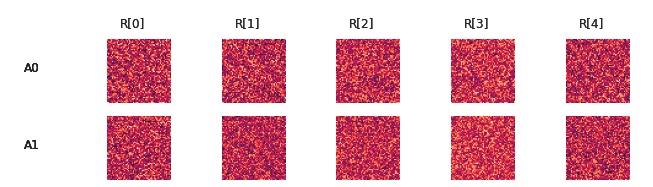}
        \captionof{figure}{\textbf{Emerging lexicon without motion primitives.} Utterances naming referents with unshared perspectives.}
        \label{fig:lexicon_ablation}
\end{figure}

\paragraph{Conclusion. } If the Haussdorf distance does not enable us to identify compositional rules in the production of utterances, it is particularly relevant for describing their coherence. This paper, therefore, provides the first step toward understanding the mechanisms at hand for the emergence of structure in self-organizing languages. The structural analysis we present sheds light on the importance of studying ecological systems. 
Indeed, agents directly optimizing utterances in pixel space can negotiate a successful communication protocol (as indicated in table~\ref{tab:lexicon_ablation}) but the absence of structure in the resulting lexicon (illustrated in figure~\ref{fig:lexicon_ablation}) prevents us from analyzing the properties of utterances. 

\begin{table}[!h]
\centering
        \small
        \begin{tabular}{|l|c|c|}    
        \hline
        & \textbf{SR}$_{\text{train}}$ & \textbf{SR}$_{\text{test}}$\\ \hline
        One-hot & $0.99 \pm 0.01$  & $0.96\pm0.02$  \\
        Visual-shared & $0.99 \pm 0.01$  & $0.55\pm0.03$  \\
        Visual-unshared & $0.99 \pm 0.01$ & $0.41 \pm 0.02$\\
        \hline
        \end{tabular}
        \captionof{table}{\textbf{Training and generalization success without DMPs.} Utterances are generated in descriptive mode, and visual referents are seen from different perspectives.}
        \label{tab:lexicon_ablation}
\end{table}



\section{Discussion and future work}

This work formalizes \greg: a new ecological referential game where two agents must communicate via a continuous sensory-motor system imitating a robotic arm drawing sketches. To tackle \greg, we propose \curves: a contrastive representation learning algorithm inspired by early language game contrastive implementation that scales to high dimensional signals. \curves allows a group of two agents two converge on a shared graphical language in contexts where referents are one-hot vectors or images of MNIST digits. The representations that agents learn enable them to communicate about compositional referents never encountered during training. If the Haussdorf distance illustrates that emergent signs are coherent, it does not capture compositionality among them. 

Future work may leverage our ecological setup and algorithmic solution to experiment with and test a variety of hypotheses that influence structures in self-organizing sing systems. An analysis of the impact of the sensory-motor constraints on the topology of graphical signs could for instance provide valuable insight into the ecological factors facilitating the emergence of a compositional graphical language. Inspired by work on the cultural evolution of language~\citep{kirby2001spontaneous}, our setup can also serve as a basis to investigate and visualize the impact of other factors such as population dynamic or cognitive abilities of agents (with varying memory or perceptual systems). Finally, \curves is agnostic to the modality used to represent utterances. As such, it could tackle other sensory-motor systems.  The central element of \curves lies in the contrastive learning of utterance-referent associations. In our implementation, we optimize utterances by maximizing this energy via gradient ascent. Much like CLIP opened many avenues for multi-modal generation, we could plug in more complex generative strategies such as diffusion models~\cite{Rombach2021HighResolutionIS,Saharia2022PhotorealisticTD}. 

\section{Reproducibility statement}
We ensure the reproduciblity of the experiments presented in this work by providing our code.
Additional information regarding the methods and hyper-parameters can be found in \ap\ref{sup:training_proc}. Information about our derived metrics can also be found in \ap\ref{sup:topo_comp}. We ensure to display the variance of our experimental results by using 10 random seeds, reporting the standard deviation. 

\section{Acknowledgements}
Experiments presented in this paper benefitted from the HPC resources of IDRIS under the allocation 2022-[A0131011996] made by GENCI.
\newpage
\bibliography{example_paper}
\bibliographystyle{icml2023}

\newpage
\appendix
\onecolumn
\begin{center}
    \Large \textsc{Supplementary Material}
\end{center}
This Supplementary Material provides additional derivations, implementation details and results. More specifically: 
\begin{itemize}[noitemsep]
\item Section A provides supplementary implementation details in the form of:
    \begin{itemize}[noitemsep]
    \item Images of testing set of visual referents;
    \item Topographic score derivation;
    \item Training procedures and hyperparameters;
    \item Pseudo-code.
\end{itemize}

\item Section B provides supplementary results:
    \begin{itemize}[noitemsep]
    \item Auto-comprehension generalization performances;
    \item Additional Lexicons;
    \item Utterances examples across perspectives illustrating coherence;
    \item Topographic maps \& scores;
    \item Composition matrix examples;
    \item T-SNEs of embeddings;
\end{itemize}
\end{itemize}

\section{Supplementary Methods}

\subsection{Sensory-Motor System}
\label{sec:suppl_sensorymotor}
\textit{Dynamic Motion Primitives.} This subsection provides additional details about the implementation of the Dynamical Movement Primitives use to produce 2-dimensional trajectories. Our drawing system consists of a 2-dimensional system that mimics the motion of a pen in a plan. Each of the $x$ and $y$ positions of the pen is controlled by a DMP starting at the center of the image and parameterized by 10 weights. These weights are the parameters of the motion of a one-dimensional oscillator that generates a smooth trajectory of 10 points. The parameters of the two DMPs are given in table~\ref{tab:dmp_parameters}.

\begin{table}[!h]
    \centering
      \def\arraystretch{1.5}
    \begin{tabular}{|c|c|}
    \hline
        \textbf{Parameter} & \textbf{Value}\\
        \hline
         Number of weights & 10  \\
         Delta time & 0.1 \\
         Number of points & 10 \\
         Weights range & [-500, 500]\\
         Position Init. & 0\\
         \hline
    \end{tabular}
    \caption{DMP parameters for each of the two coordinate motions}
    \label{tab:dmp_parameters}
\end{table}

\textit{Sketching Library. } Trajectories obtained with the DMPs are then mapped to a 52x52 grid which is converted to an image with the \texttt{raster} and \texttt{softor} functions of the sketching library~\cite{Mihai2021DifferentiableDA}. The drawing thickness parameter is fixed to $1\text{e}-2$.

\newpage
\subsection{Testing Set}

\begin{figure}[h!]
\centering 
 \begin{tabular}{ccccc}
     \includegraphics[width=0.1\textwidth]{"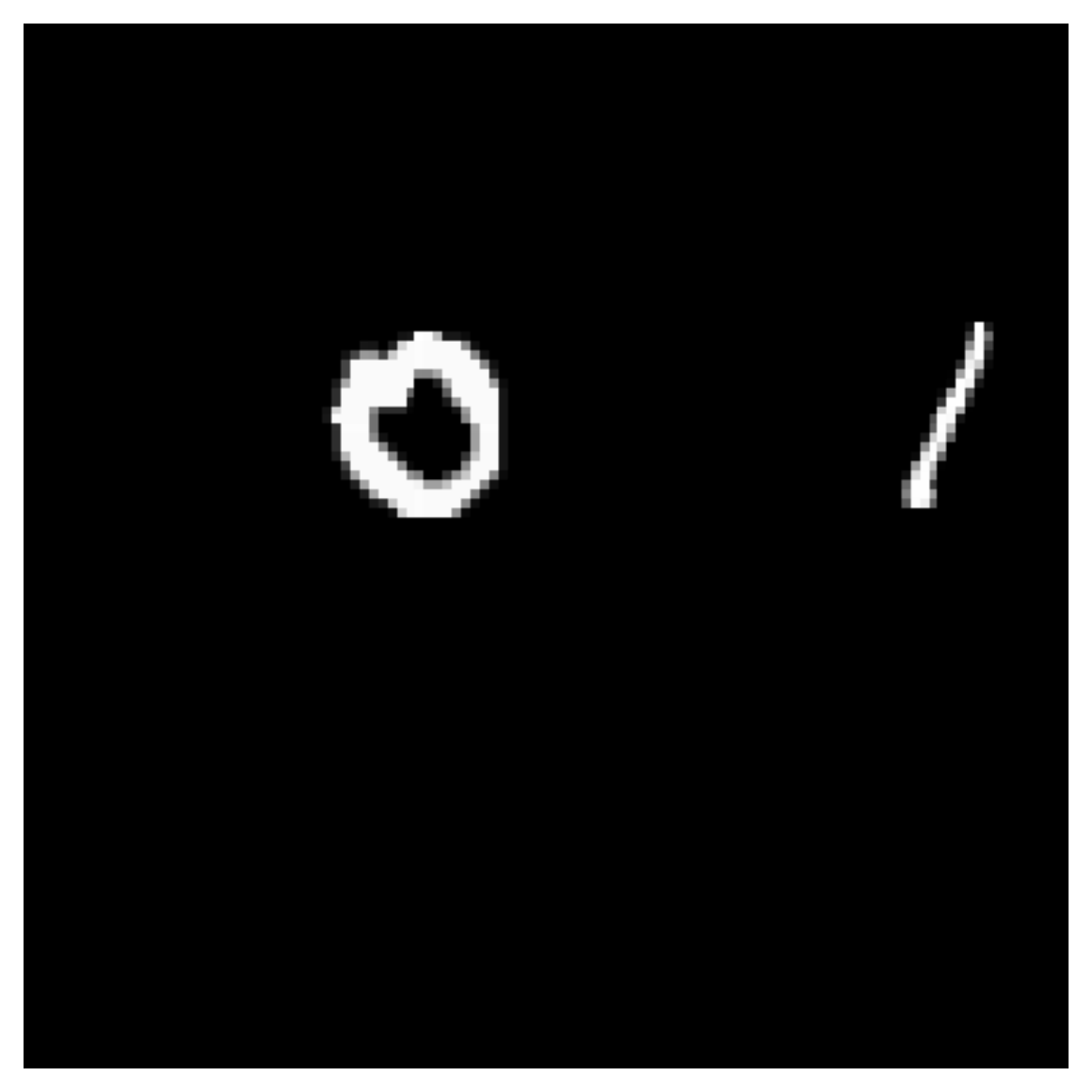"} &
     \includegraphics[width=0.1\textwidth]{"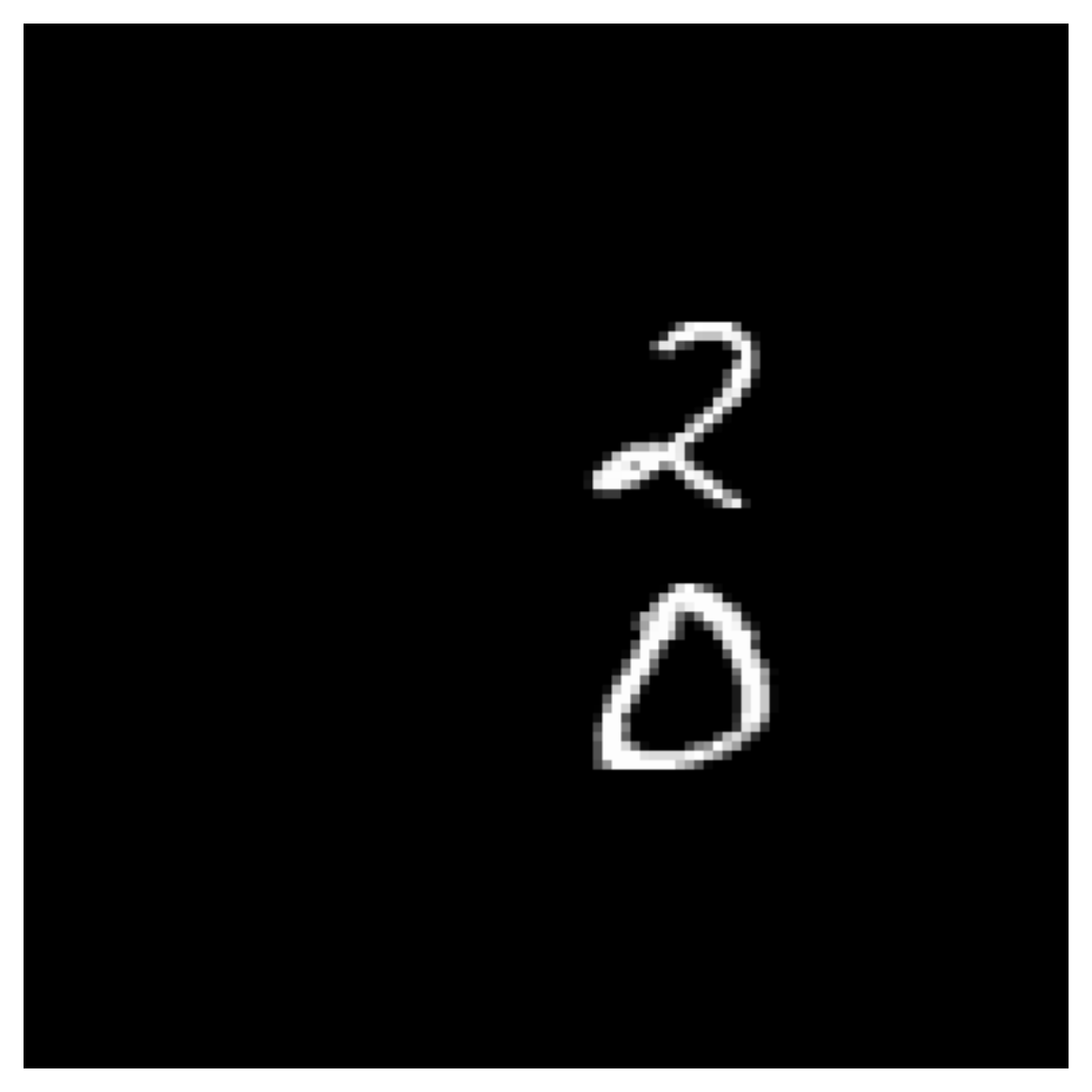"} &
     \includegraphics[width=0.1\textwidth]{"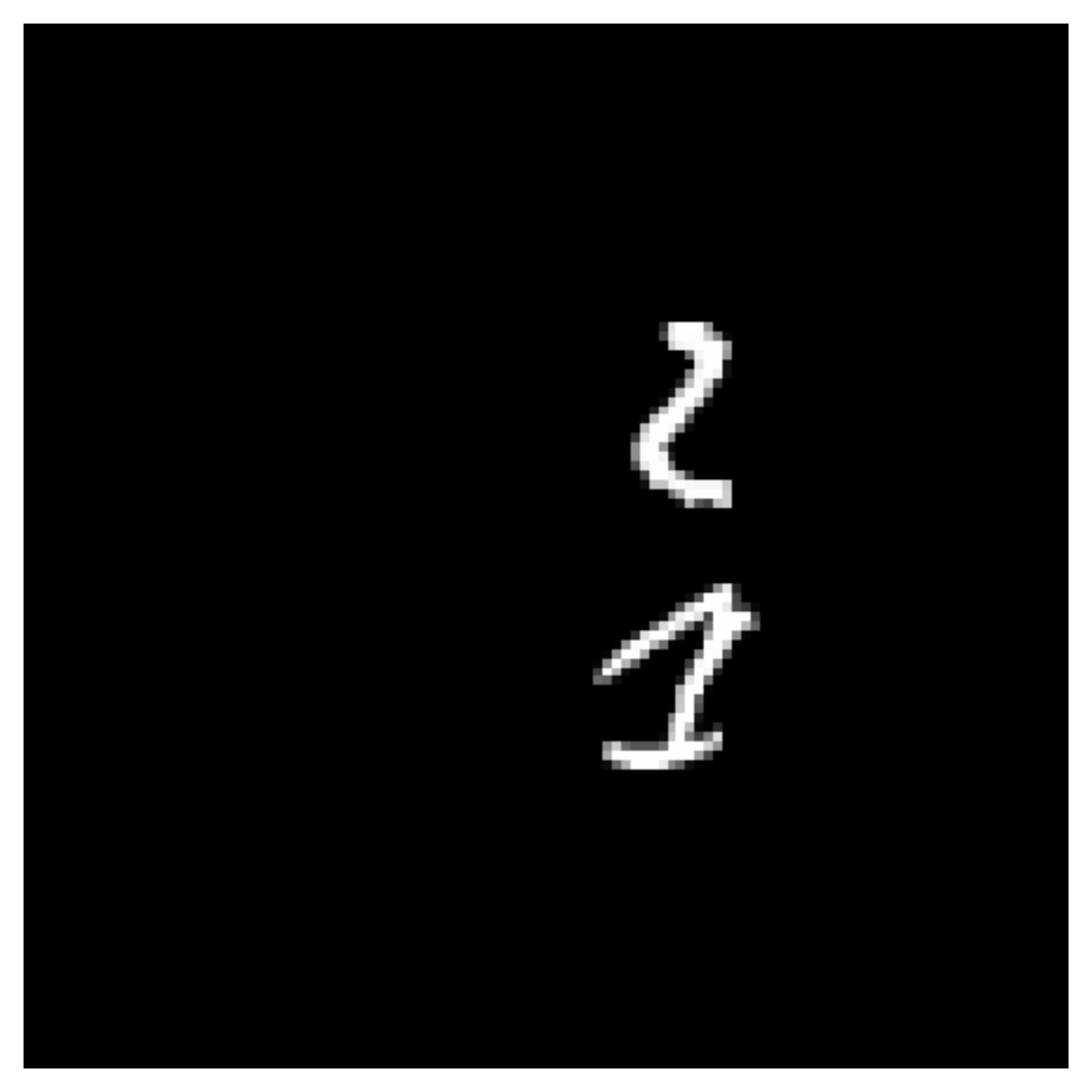"} &
     \includegraphics[width=0.1\textwidth]{"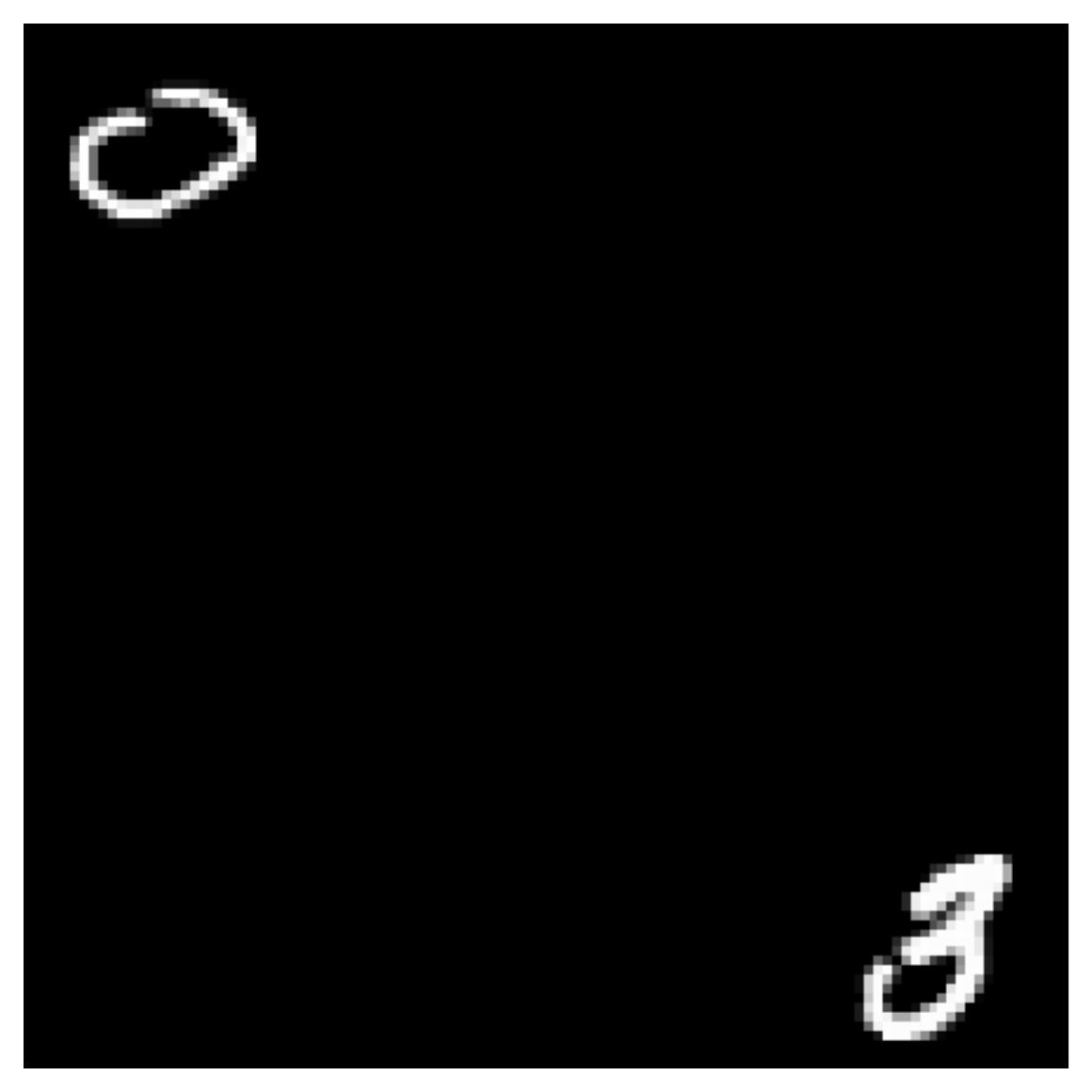"} &
     \includegraphics[width=0.1\textwidth]{"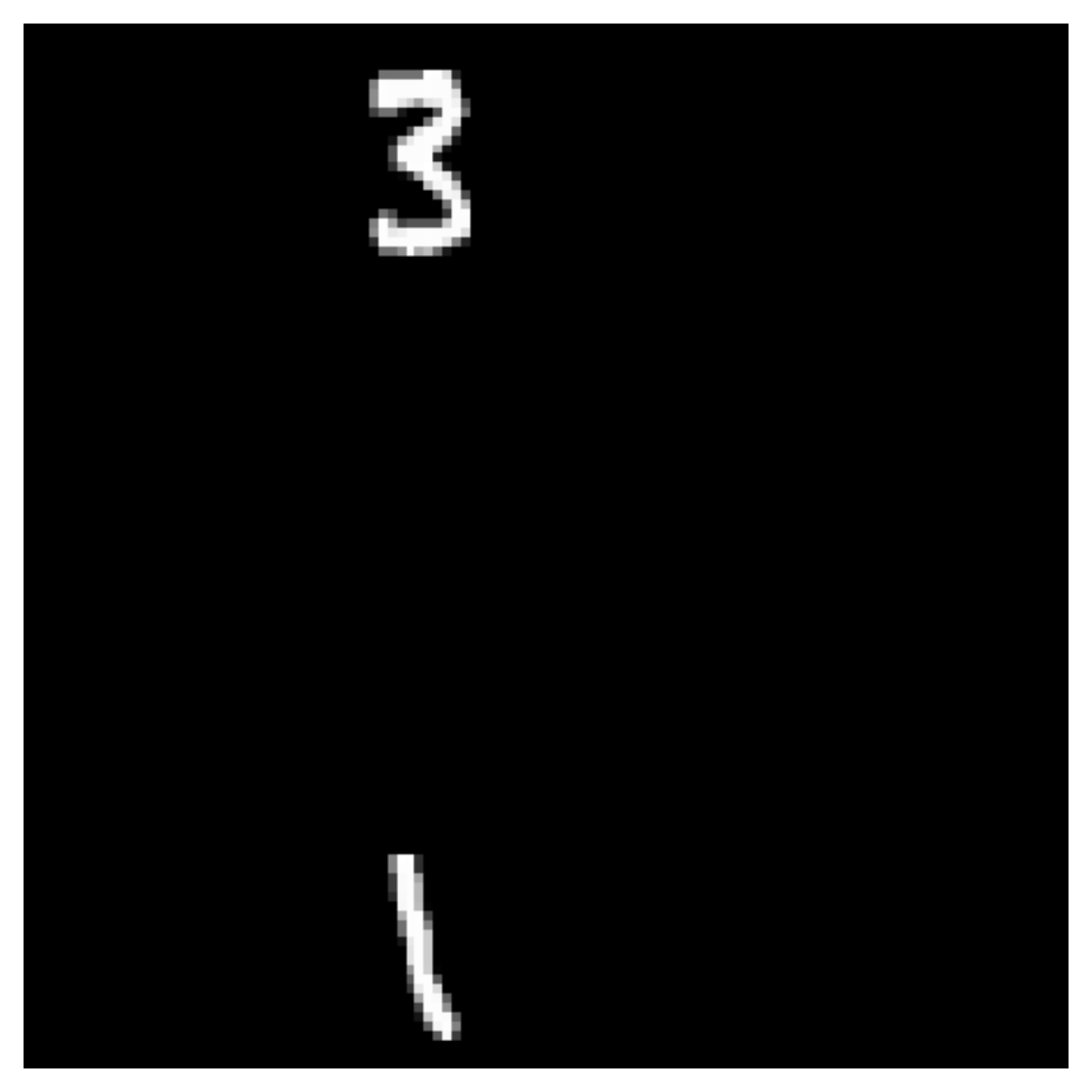"} \\
     \includegraphics[width=0.1\textwidth]{"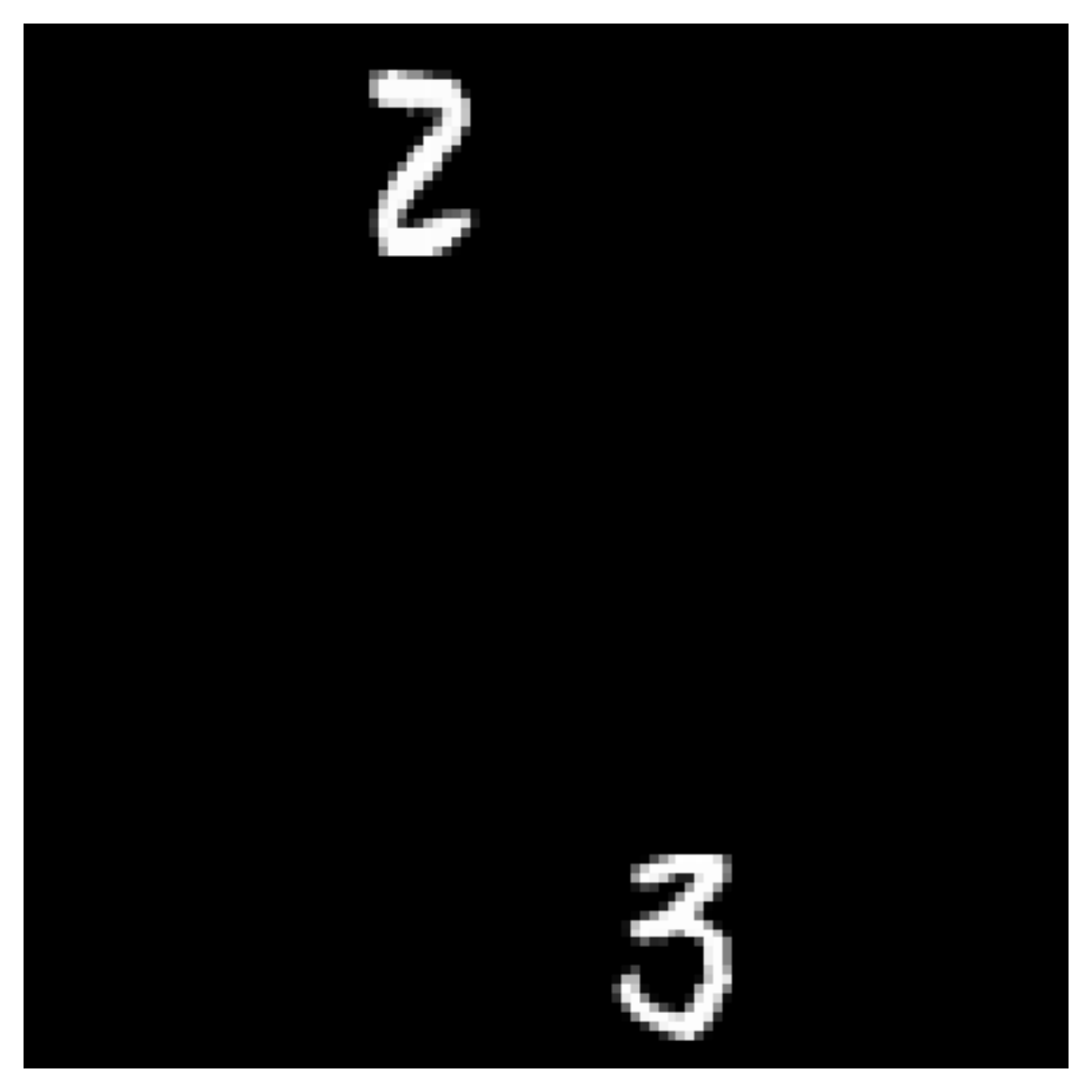"} &
     \includegraphics[width=0.1\textwidth]{"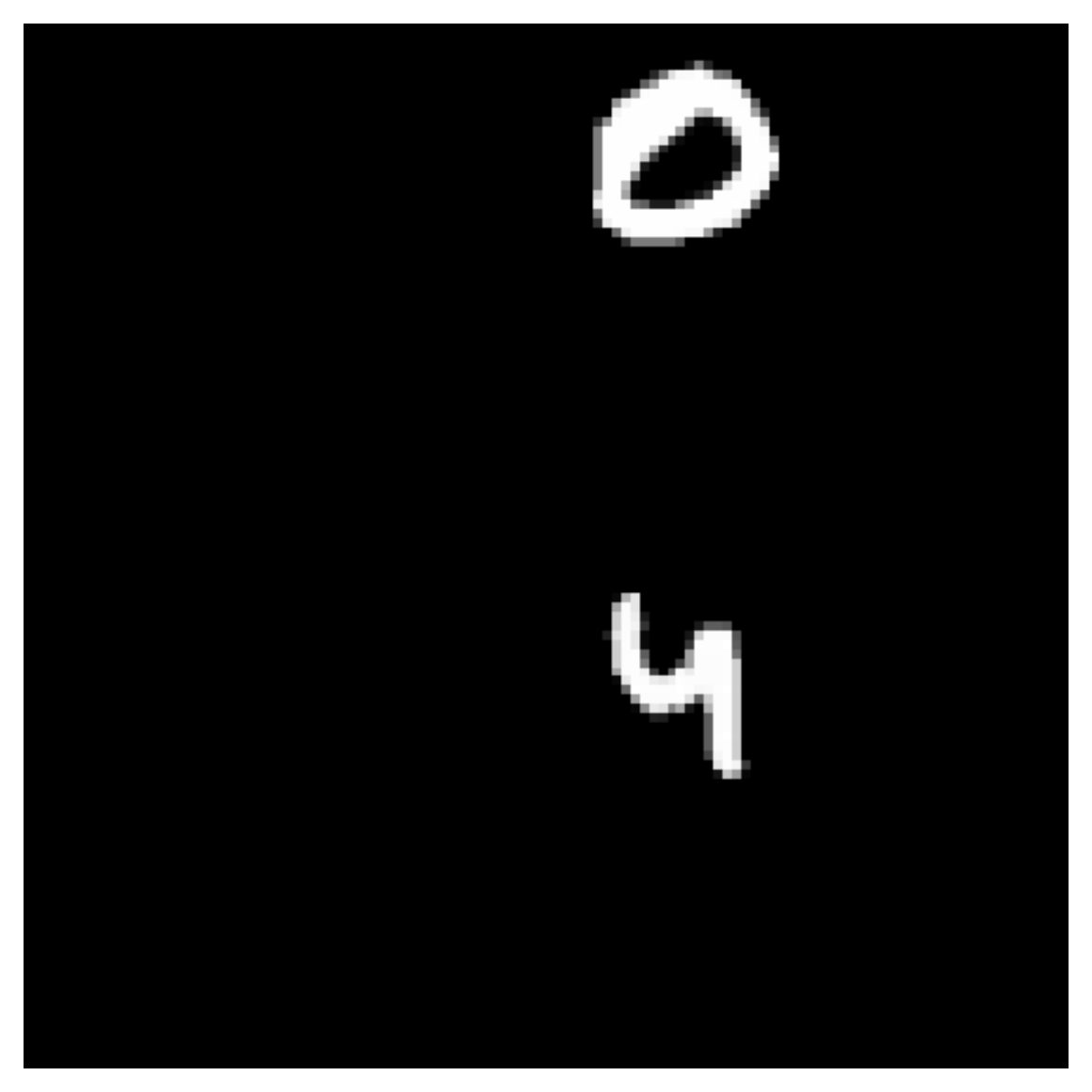"} &
     \includegraphics[width=0.1\textwidth]{"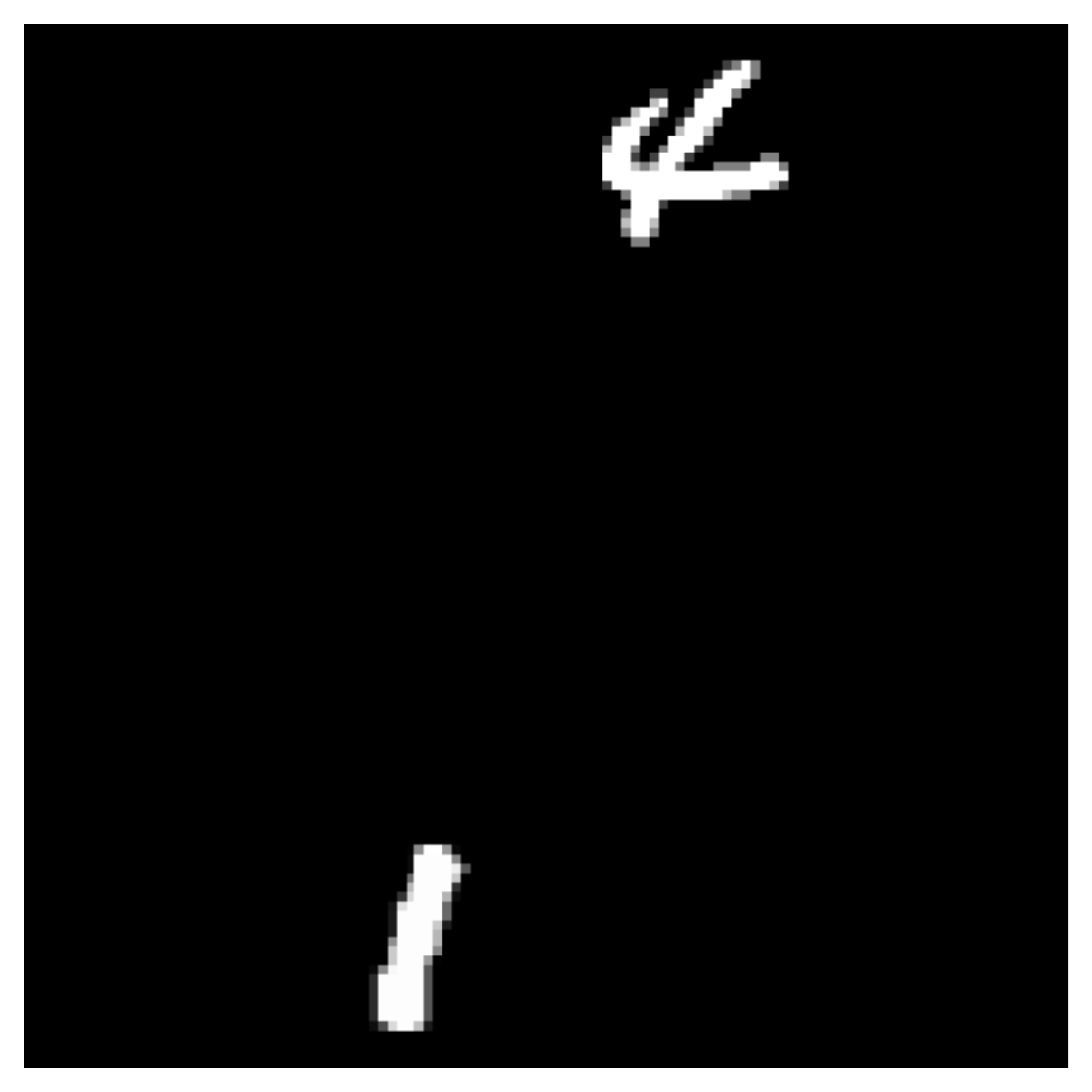"} &
     \includegraphics[width=0.1\textwidth]{"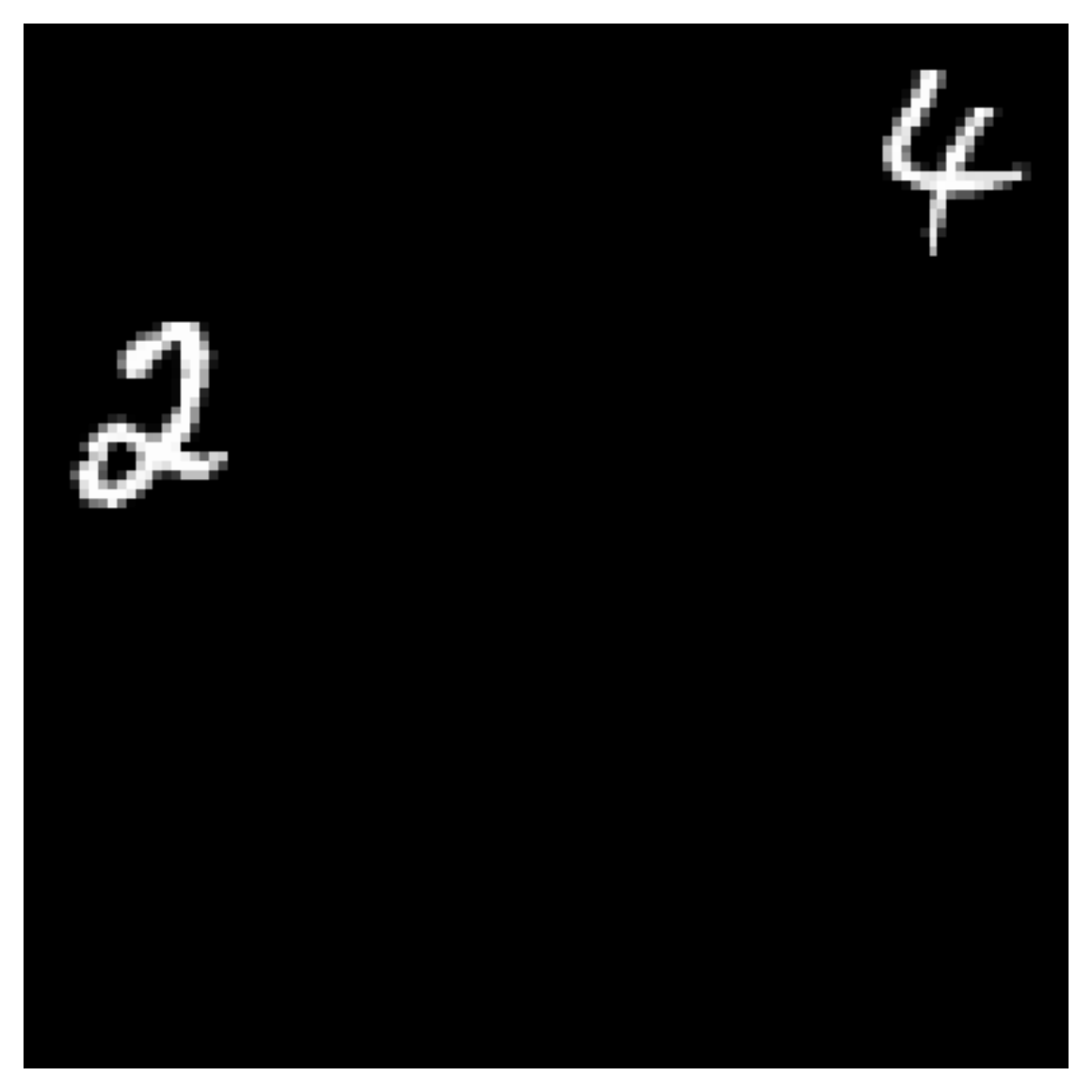"} &
     \includegraphics[width=0.1\textwidth]{"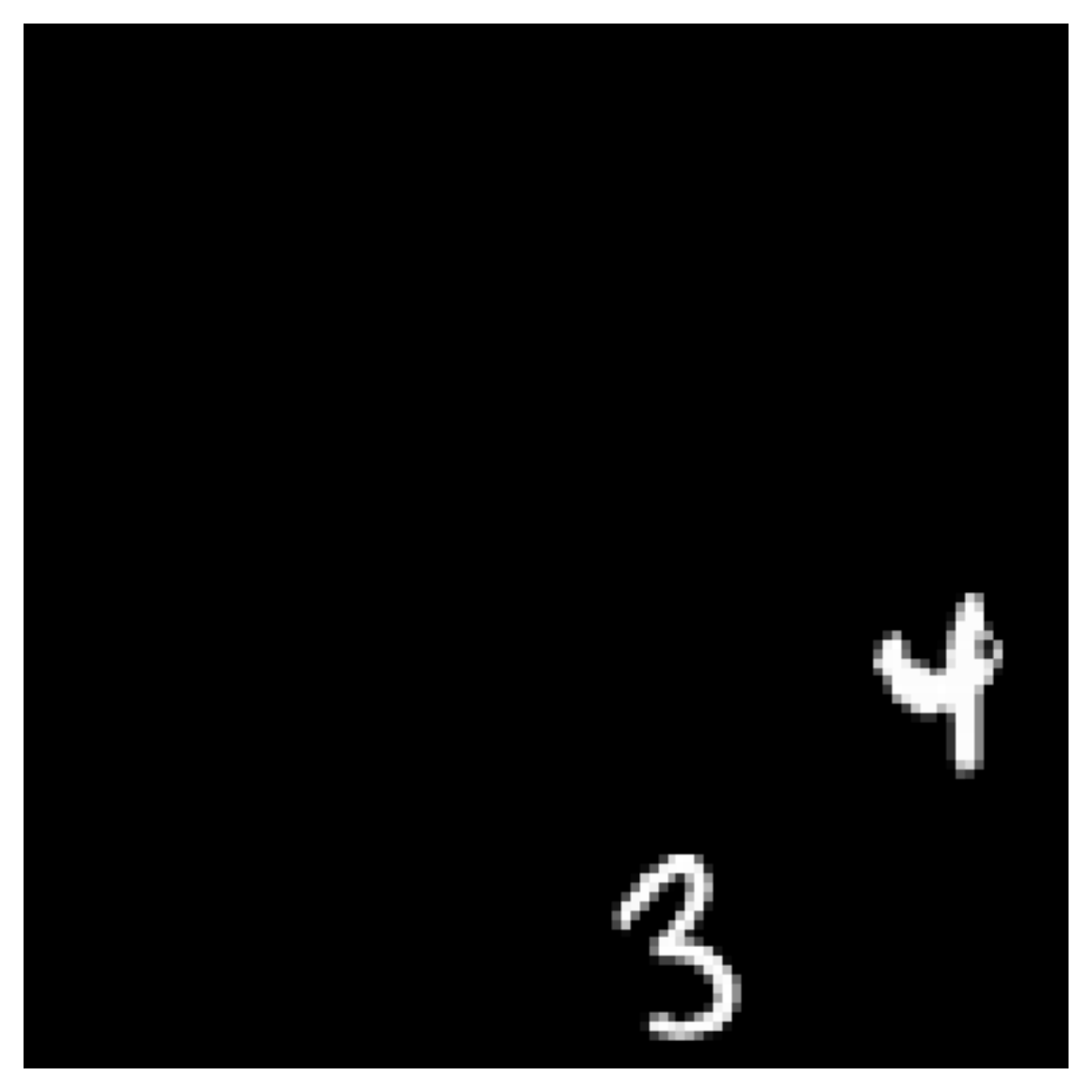"}
 \end{tabular}1

\caption{\textbf{Perspective instances of the testing set} $\mathcal{R}_5^2$.}
\label{fig:suppl_referents}
\end{figure}

Figure~\ref{fig:suppl_referents} displays examples of compositional referents made of 2 features.

\label{sup:ref_comp}
\subsection{Topographic Score}
\label{sup:topo_comp}
To evaluate the compositionality of the emerging language we define the topographic score:
\begin{equation}
    \rho_{ij} = ||(O, h_{ij})||_2 - ||(O, h_k)||_2 \text{ with } k = \text{argmin}_{k\in \{i, j\}}||h_k,h_{ij}||_2)
    \label{eq:topo_score}
\end{equation}

It is obtained by computing the Hausdorff distance between the utterances denoting compositional referents with respect to both the utterance denoting the single feature $i$ ($d_H(u(r_i),\cdot)$)and the one denoting the single feature $j$ ($d_H(u(r_j),\cdot)$). To derive our metric, we define 4 groups of utterances denoting compositional referents. 
\begin{itemize}[noitemsep,topsep=0pt]
    \item $u(r_{ij})$ the utterances for referent made of feature $i$ and $j$.
    \item $u(r_{xj}, x\neq i)$ the utterances denoting referent made by composing feature $j$ with any other feature different than $i$
    \item $u(r_{iy}, y\neq j)$ the utterances denoting referent made by composing feature $i$ with any other feature different than $j$
    \item $u(r_{xy})$ the utterance denoting all other compositional referents in $\mathcal{R}^2_5$.
\end{itemize}
and compute their Hausdorff distances to $u(r_i)$ and $u(r_j)$.

\begin{figure}[h!]
    \centering
    \includegraphics[width=0.5\textwidth]{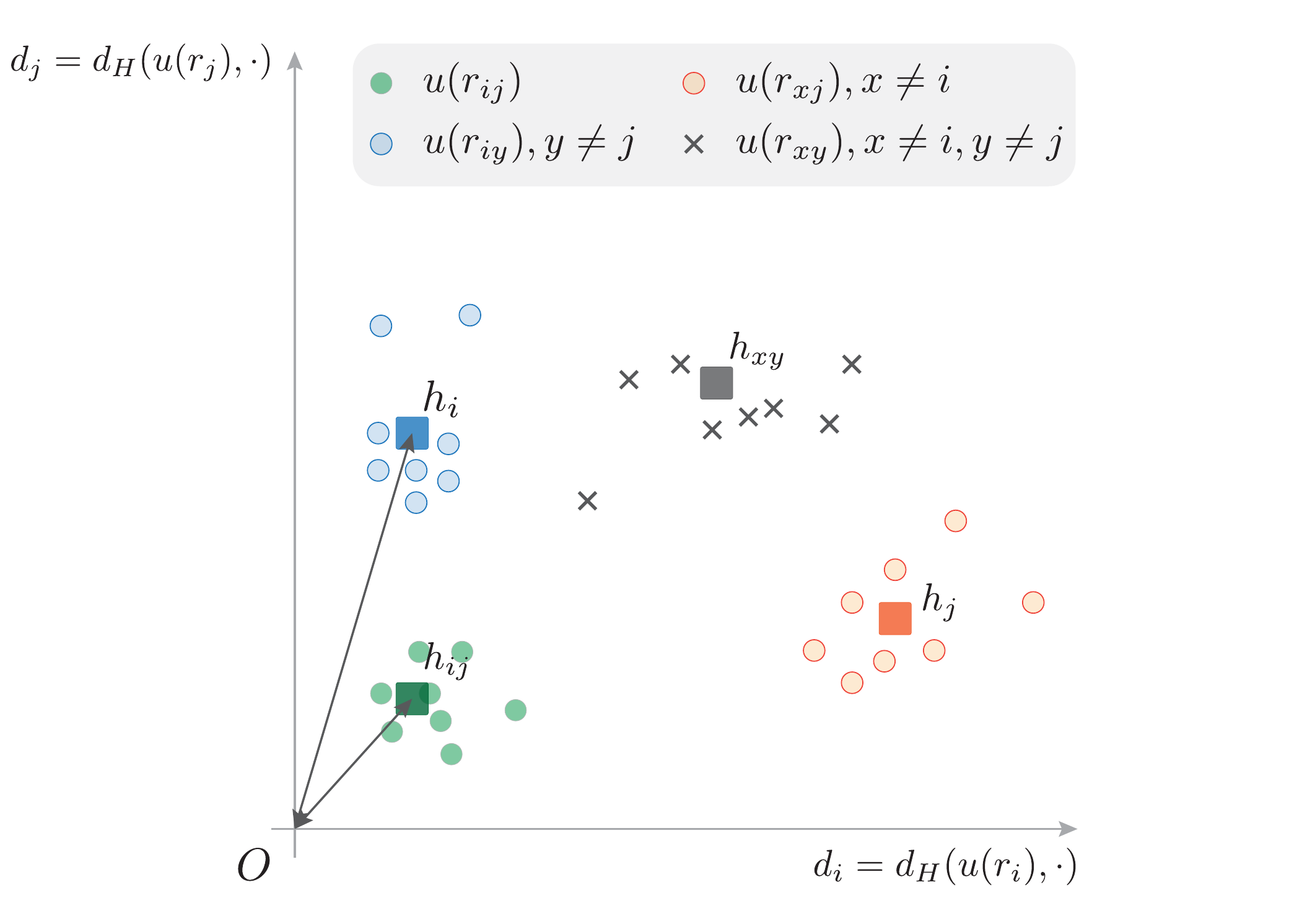}
    \caption{Idealized mapping of utterances denoting compositional referents in the plan representing distances to utterances naming isolated features $i$ and $j$.}
    \label{fig:topo_distances_intuitive}
\end{figure}

As displayed in Figure~\ref{fig:topo_distances_intuitive}, if utterances $u(r_{ij})$ are compositional we expect them to be at the same time close to $u(r_{i})$ and close to $u(r_{j})$ and hence to land in the bottom left corner of the distance graph. Moreover, they should be closer to the origin than $u(r_{xj})$ and $u(r_{iy})$. To quantify to what extent it is the case we compute the barycenter of each group $h_i, h_j, h_{ij}$ and $h_{xy}$ and compute "how closer to the origin" is the compositional barycenter $h_{ij}$ compared to its closest barycenter using equation~\ref{eq:topo_score}.

\newpage
\subsection{Training procedure and hyperparameters}
\label{sup:training_proc}
Agents have two separate encoders based on the same model architecture described in Table.\ref{table:hyperparameters}. Each agent performs association updates with a single step of gradient descent, using its own Adam optimizer with a learning rate of $1e^{-4}$. To allow faster convergence, agents perform an association update between an abstract referent $r^\star_A$ and an utterance $u$ by using a batch of 64 perspectives $\{\Phi(r^\star_A)\}_{i \in [1,64]}$. From a cognitive science perspective, this is comparable to an agent "walking around" an object to better understand how different perceptions relate to the same object. From a computer science perspective, this is similar to the self-supervised framework of SimCLR~\citep{Chen2020ASF}, where agents learn representation by contrastively aligning the embeddings of an input with these of the same transformed input.

\begin{table}[h!]
\centering 

\def\arraystretch{1.5}
\begin{tabular}{ |c|c|} 
 \hline
 \textbf{Layer} & \textbf{Activation}\\
 \hline
 Conv2D(filters=8, stride=2, padding=1) & ReLU\\ 
 \hline
 Conv2D(filters=16, stride=2, padding=1) & ReLU\\ 
 \hline
 Conv2D(filters=32, stride=2, padding=0) & ReLU\\ 
 \hline
 Linear(128) & ReLU\\
 \hline
 Linear(32) & None\\
 \hline
\end{tabular}

\caption{\textbf{Model architecture used for both the referent and utterance Encoders.} (when referents are one-hot vectors, the 3 Conv2D layers are replaced by a Linear layer with ReLu activation)}

\label{table:hyperparameters}
\end{table}

While the drawing pipeline is fully differentiable, it is highly sensitive to local minima. Thus, we solve equation~\ref{eq:descri_gen} in the descriptive case or equation~\ref{eq:discri_gen} in the discriminative scenario by simultaneously performing gradient descent on a batch of $64$ randomly initialized command vectors over 100 iterations, using a newly initialized Adam optimizer each time with a learning rate of $1e^{-2}$.

\subsection{Pseudo-code}
\textbf{ }
\begin{algorithm}
\caption{Speaker's Utterances}\label{alg:sp}
\begin{algorithmic}

\REQUIRE perceived referents $\tilde{R}_S$, speaker's referent encoder $f_S$, speaker's utterance encoder $g_S$, sensory-motor system $M$

\STATE $Z_r \gets f_S(\tilde{R}_S) $
\STATE $c \sim \textrm{Uniform}()$
\FOR{$i \textrm{ in range} (N_{\textrm{production}})$}
    \STATE $U_S \gets M(c)$
    \STATE $Z_u \gets g_S(U)$
    \STATE $S \gets \textrm{sim}_{\textrm{cos}}(Z_r,Z_u)$
    \STATE $\mathcal{L} \gets \textrm{mean(diag(S))} * (-1)$
    \STATE GD step on $c$ to minimize $\mathcal{L}$
\ENDFOR
\STATE \textrm{\textbf{Return} }$M(c)$
\end{algorithmic}
\end{algorithm}

\begin{algorithm}
\caption{Listener's Selections \& Binary Outcomes}\label{alg:ls}
\begin{algorithmic}

\REQUIRE perceived referents $\tilde{R}_L$, produced utterances $U_S$, listener's referent encoder $f_L$, listener's utterance encoder $g_L$

\STATE $Z_r \gets f_L(\tilde{R}_L) $
\STATE $Z_u \gets g_L(U_S) $
\STATE $S\gets \textrm{sim}_{\textrm{cos}}(Z_r,Z_u)$
\STATE $t \gets \textrm{argmax}(S, \textrm{axis=}1)$
\STATE $o \gets \textbf{0}$
\FOR{$i \textrm{ in range} (N_{\textrm{referents}})$}
    \STATE $o_i \gets \mathds{1}_{[t_i = i]}$
\ENDFOR
\STATE \textrm{\textbf{Return} }$o$
\end{algorithmic}
\end{algorithm}

\begin{algorithm}
\caption{Agents's Association Losses}\label{alg:al}
\begin{algorithmic}

\REQUIRE perceived referents $\tilde{R}_A$, produced utterances $U_A$, outcomes $o$, agent's referent encoder $f_A$, agent's utterance encoder $g_A$

\STATE $Z_r \gets f_A(\tilde{R}_A) $
\STATE $Z_u \gets g_A(U_A) $
\STATE $S\gets \textrm{sim}_{\textrm{cos}}(Z_r,Z_u)$
\STATE $\mathcal{L}_0 \gets CE(S, \textrm{reduction=False})$
\STATE $\mathcal{L}_1 \gets CE(S^{\top},\textrm{reduction=False})$
\STATE $\mathcal{L} \gets (\mathcal{L}_0 + \mathcal{L}_1)/2$
\IF{$A = \textrm{"S"}$} 
    \STATE $\mathcal{L} \gets (\mathcal{L} \cdot o) / N_{\textrm{referents}}$
\ELSE
    \STATE $\mathcal{L} \gets (\mathcal{L} \cdot \textbf{1}) / N_{\textrm{referents}}$
\ENDIF 

\STATE $\textrm{\textbf{Return} } \mathcal{L}$
\end{algorithmic}
\end{algorithm}

\newpage
\section{Supplementary Results}

\subsection{Auto-comprehension generalization performances}
\label{sup:auto_social_perf}

\def\arraystretch{1.5}
\begin{table}[!h]
\centering
    \begin{tabular}{|c|c|c|}
     \hline
     \textbf{Ref.} & \textbf{Auto} & \textbf{Social}\\
     \hline
     One-hot & $0.997 \pm 0.005$ & $0.991 \pm 0.015$ \\
     Visual-shared & $0.862 \pm 0.034$ & $0.559 \pm 0.027$  \\
     Visual-unshared & $0.425 \pm 0.016$ & $0.388 \pm 0.02$  \\
     \hline
    \end{tabular}
    \caption{Descriptive Success Rate}
    \label{tab:disc_auto_social}
\end{table}

\def\arraystretch{1.5}
\begin{table}[!h]
\centering
    \begin{tabular}{|c|c|c|}
     \hline
     \textbf{Ref.} & \textbf{Auto} & \textbf{Social}\\
     \hline
     One-hot & $0.997 \pm 0.005$ & $0.992 \pm 0.009$ \\
     Visual-shared & $0.812 \pm 0.019$ & $0.567 \pm 0.034$  \\
     Visual-unshared & $0.466 \pm 0.019$ & $0.404 \pm 0.019$  \\
     \hline
     
    \end{tabular}
    \caption{Descriminative Success Rate}
    \label{tab:desc_auto_social}
\end{table}

We define the \textbf{Auto} performance metric as the communicative success rate, on test set, for language games involving a single agent playing as both the speaker and listener. We compare \textbf{Auto} and \textbf{Social} performances (the latter involving pairs of different agents, as done until now) in Tables~\ref{tab:desc_auto_social}~\&~\ref{tab:disc_auto_social}.

\subsection{Additional Lexicons}
\label{sup:lexicon_one_hot_shared}

\begin{figure}[h!]

\centering
\includegraphics[trim={5cm 1cm 1cm 2cm},clip,width=0.85\textwidth]{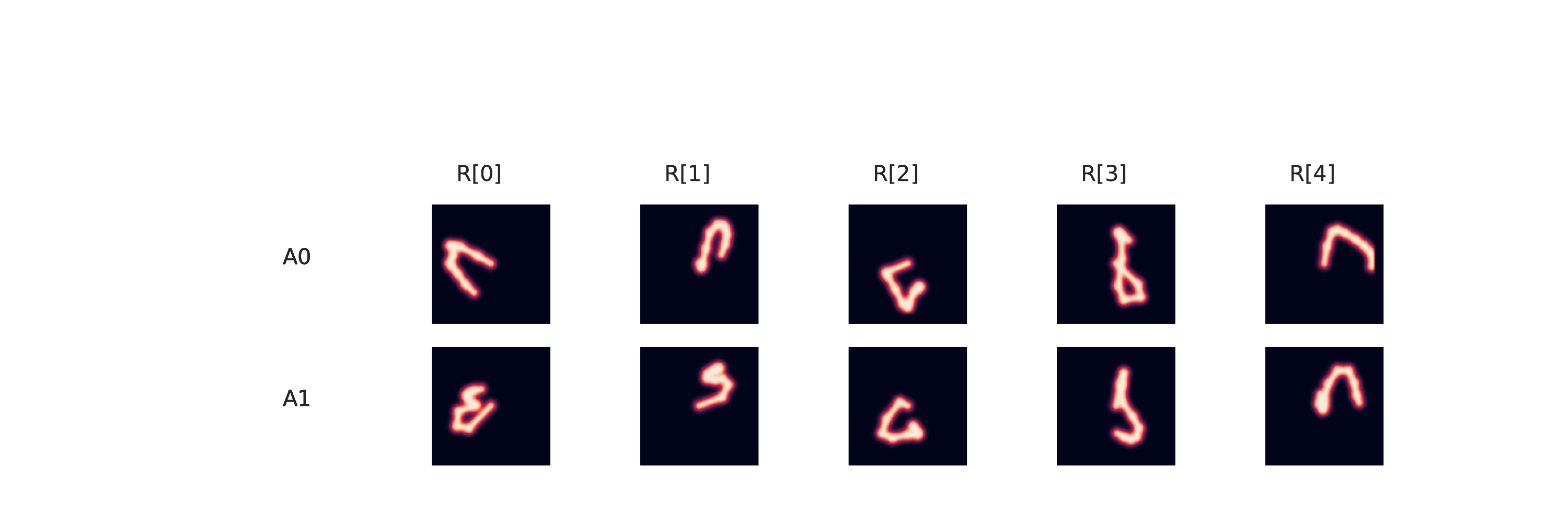}
\caption{\textbf{Instance of an emerging lexicon.} (Utterances are naming visual-shared referents).}
\end{figure}

\begin{figure}[h!]

\centering
    \includegraphics[trim={5cm 1cm 1cm 2cm},clip,width=0.85\textwidth]{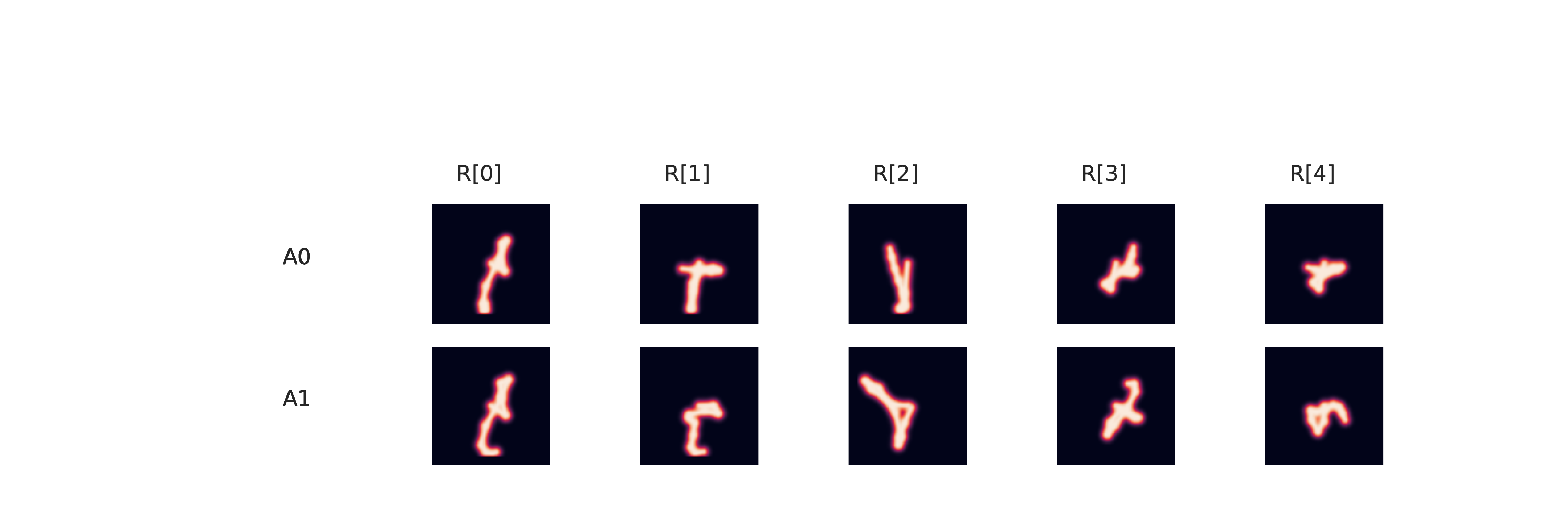}
    
\caption{\textbf{Instance of an emerging lexicon.} (Utterances are naming one-hot referents).}
\end{figure}

\newpage
\subsection{Utterances examples across perspectives illustrating coherence.}
\label{sup:P-coherence}

The following figures illustrate the P-coherence and A-coherence of an emerging lexicon (Visual-unshared) by displaying, for each referent in $R_1$, the descriptive utterance produced for 10 random perspectives.

\begin{figure}[h!]
\centering
    \includegraphics[trim={5cm 1cm 1cm 1cm},clip,width=1\textwidth]{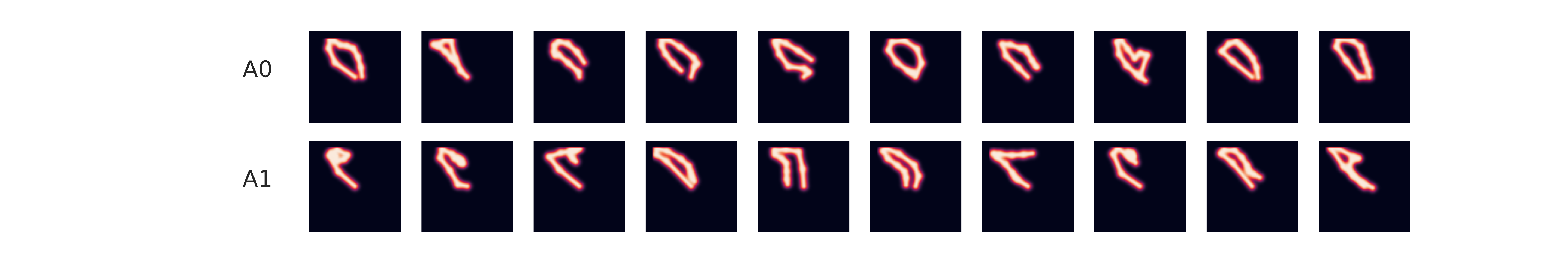}
\caption{Utterances examples for referent 0.}
\end{figure}

\begin{figure}[h!]
\centering
    \includegraphics[trim={5cm 1cm 1cm 1cm},clip,width=1\textwidth]{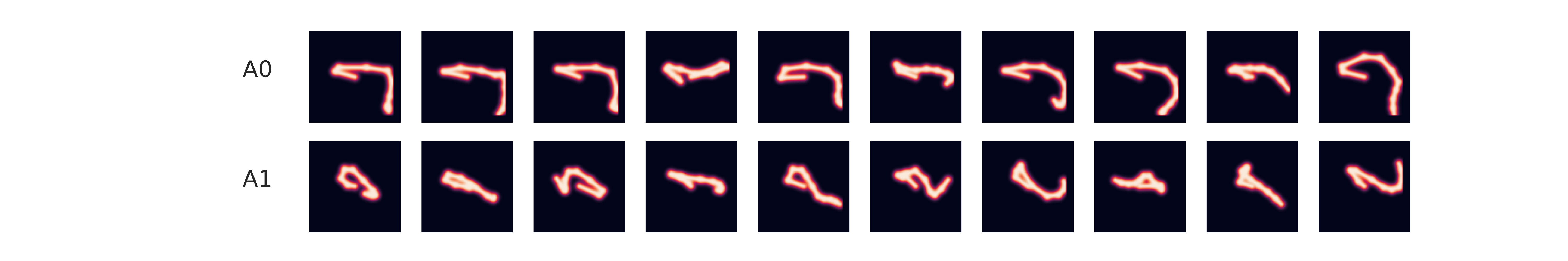}
\caption{Utterances examples for referent 1.}
\end{figure}

\begin{figure}[h!]
\centering
    \includegraphics[trim={5cm 1cm 1cm 1cm},clip,width=1\textwidth]{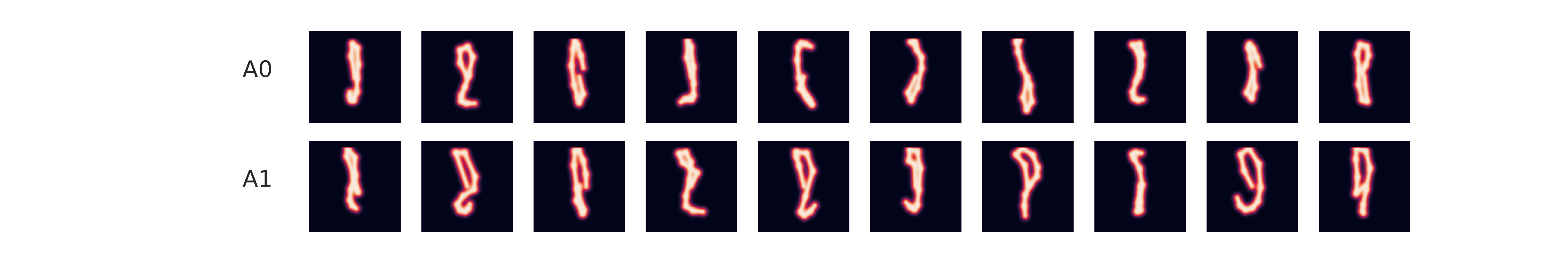}
\caption{Utterances examples for referent 2.}
\end{figure}

\begin{figure}[h!]
\centering
    \includegraphics[trim={5cm 1cm 1cm 1cm},clip,width=1\textwidth]{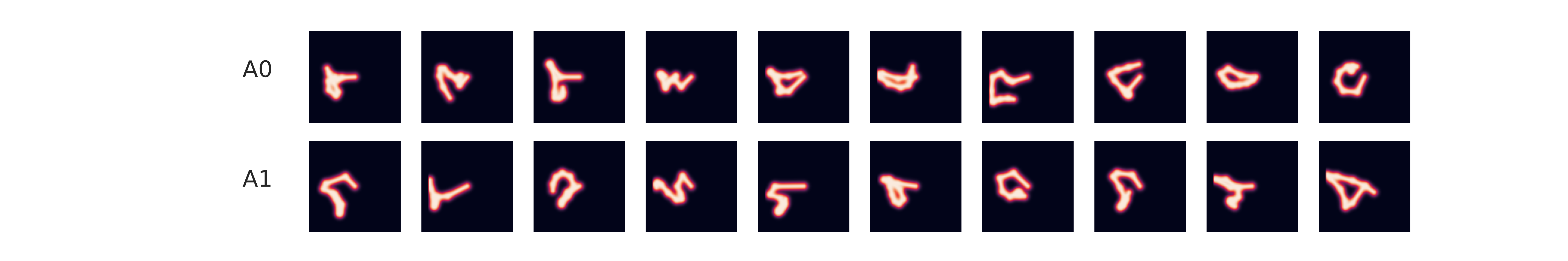}
\caption{Utterances examples for referent 3.}
\end{figure}

\begin{figure}[h!]
\centering
    \includegraphics[trim={5cm 1cm 1cm 1cm},clip,width=1\textwidth]{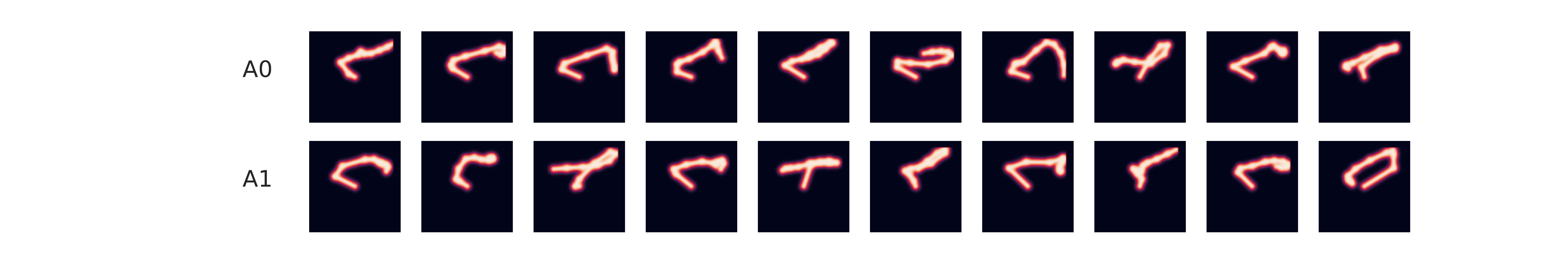}
\caption{Utterances examples for referent 4.}
\end{figure}

\newpage
\subsection{Topographic Maps \& Scores}
\label{sup:topo_maps}

\subsubsection{One-Hot}

\begin{figure}[!h]
    \centering
     \begin{tabular}{ccc}

     \includegraphics[width=0.3\textwidth]{figures/onehot_distance_plots/0_d=-0.401.pdf} & \includegraphics[width=0.3\textwidth]{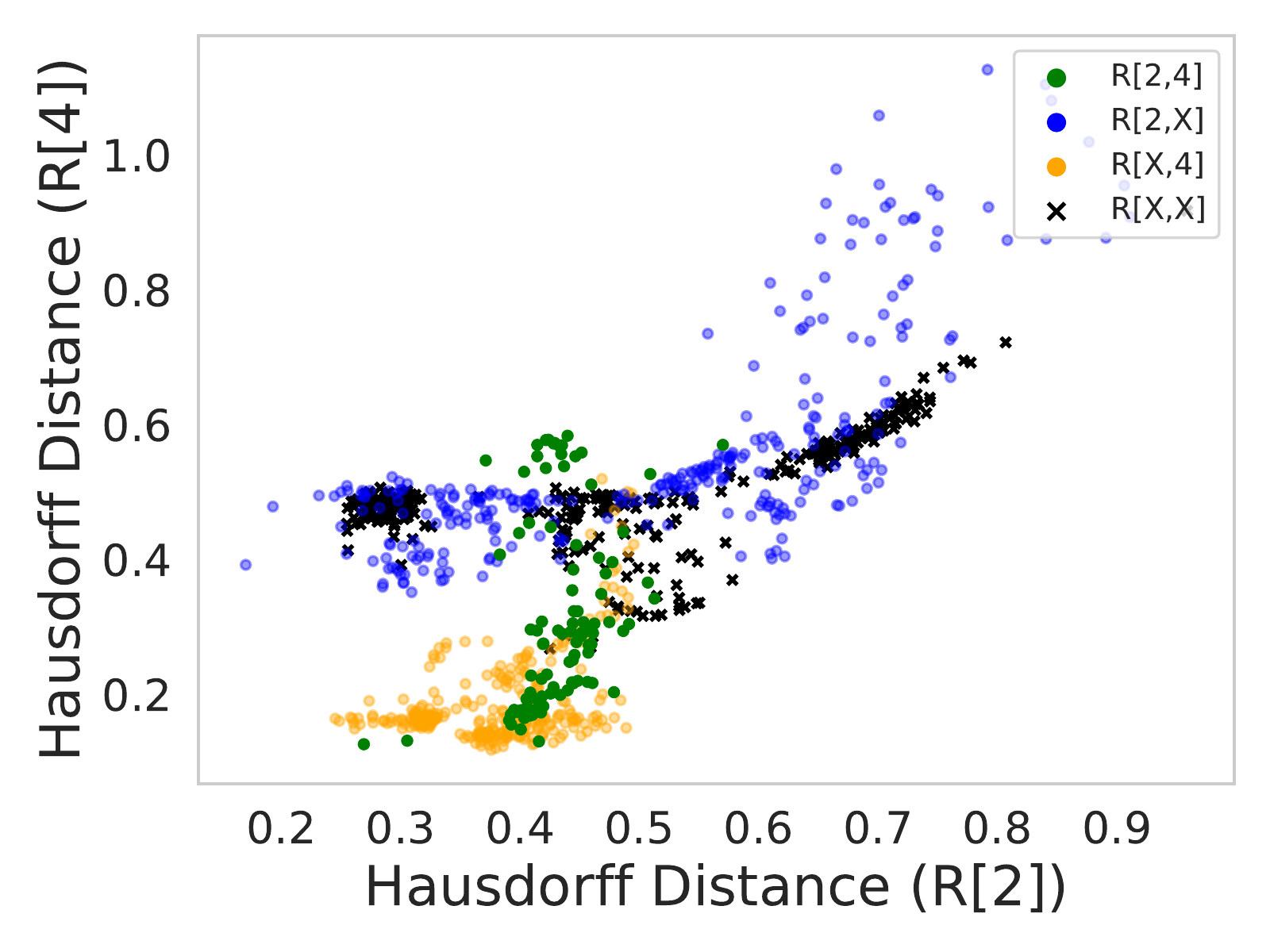} &  \includegraphics[width=0.3\textwidth]{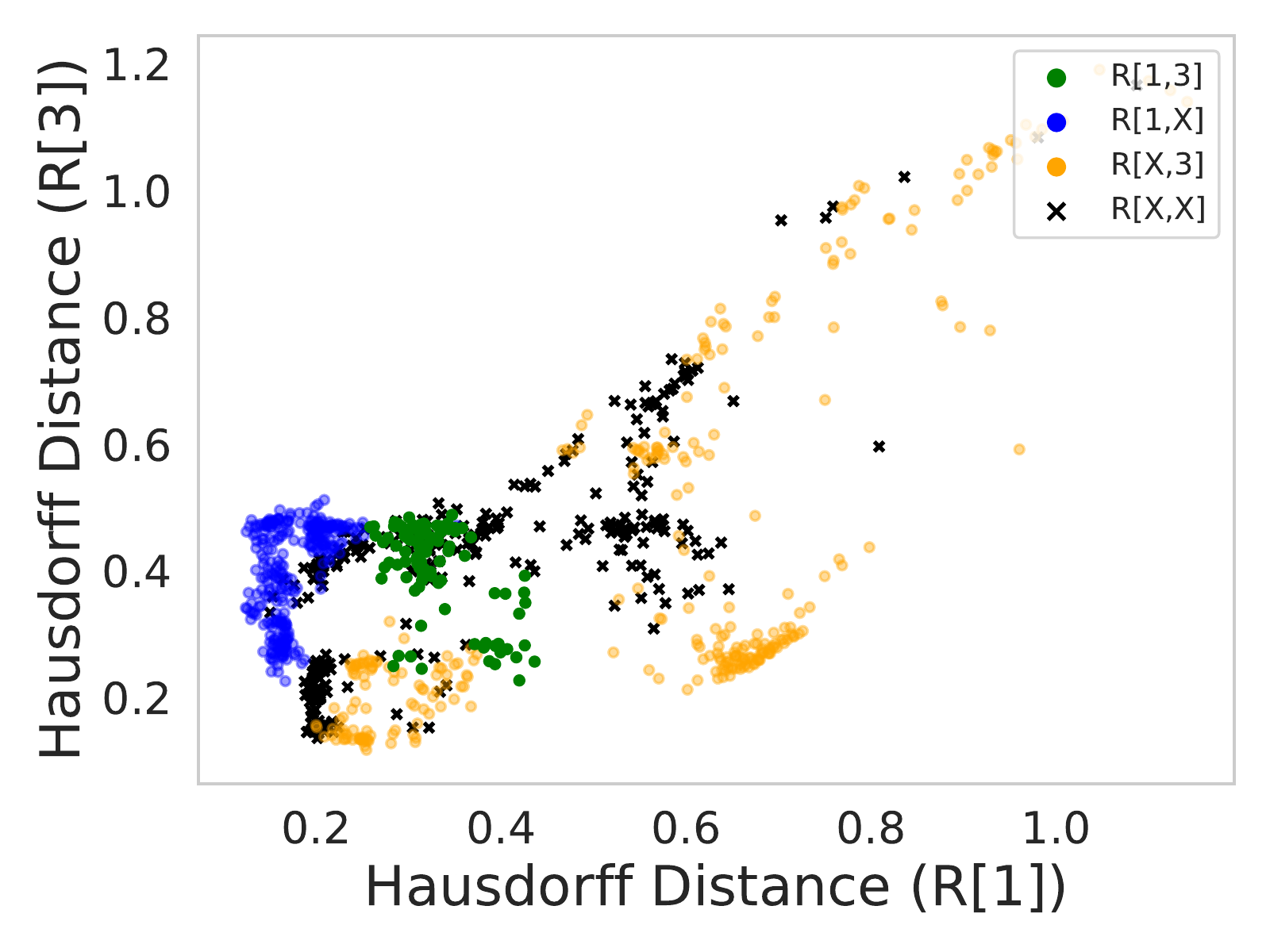}    \\
     $\rho=-0.401$ & $\rho=-0.111$ & $\rho=-0.085$                                                                \\
     \includegraphics[width=0.3\textwidth]{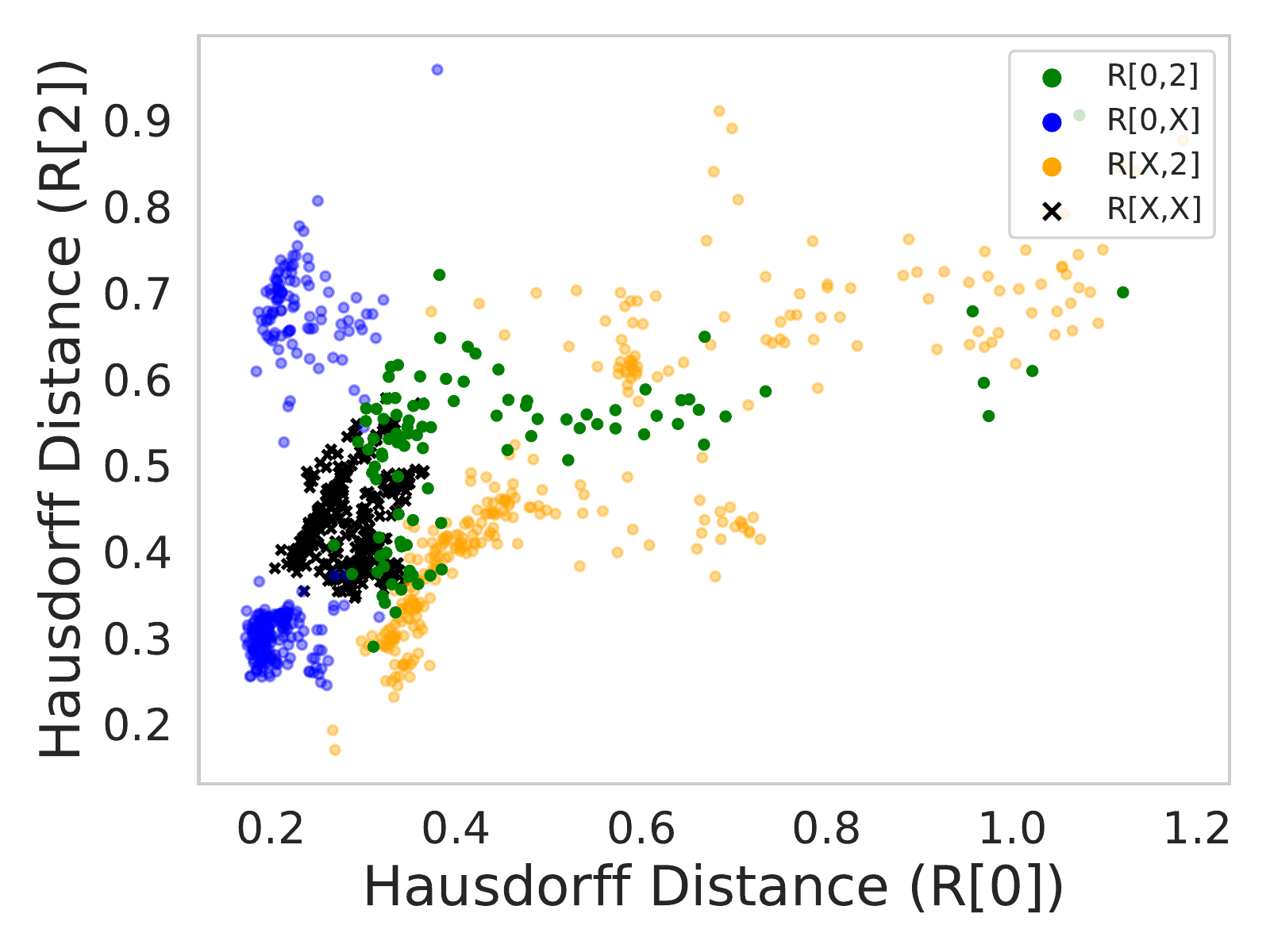} & \includegraphics[width=0.3\textwidth]{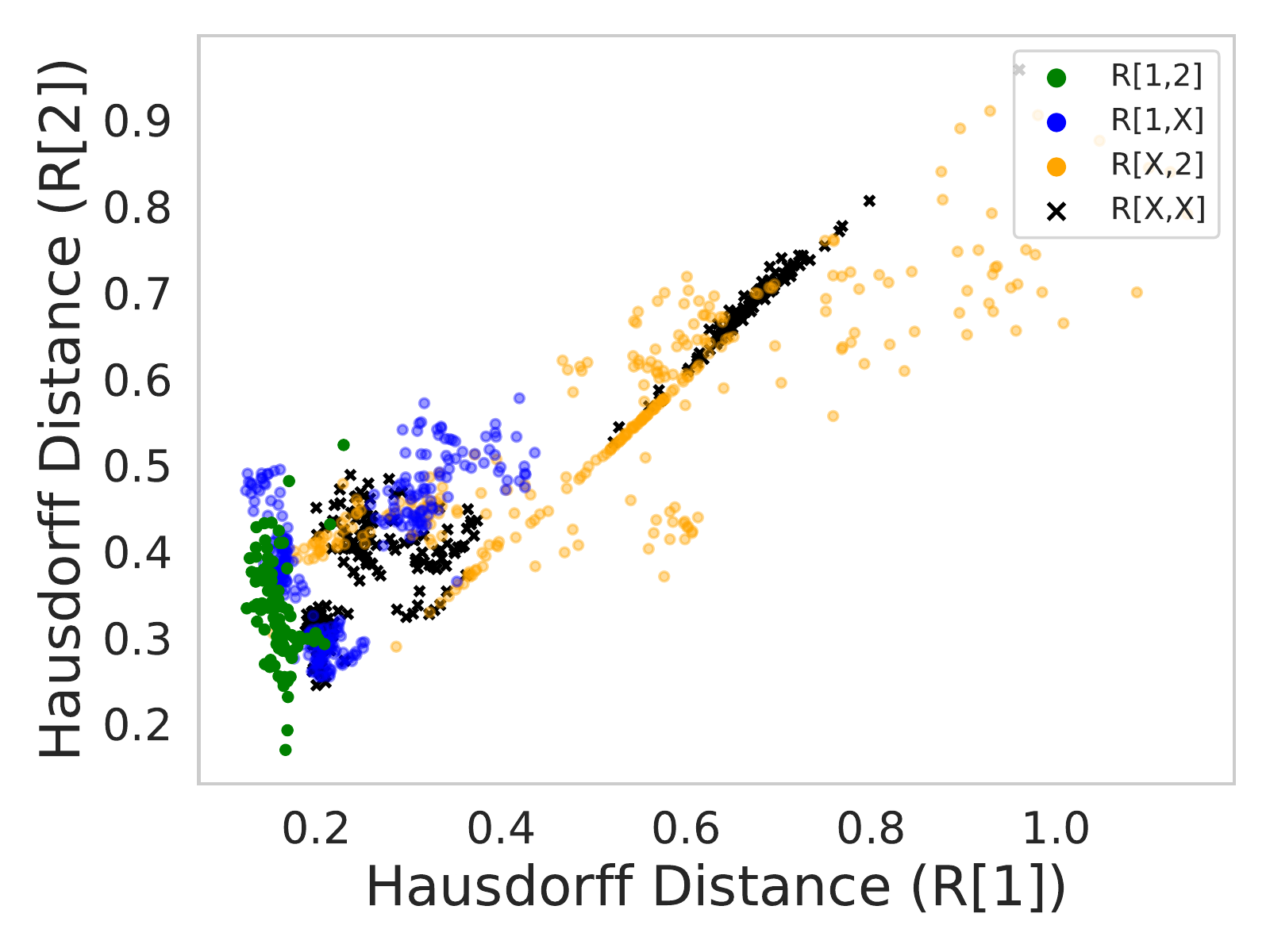} &  \includegraphics[width=0.3\textwidth]{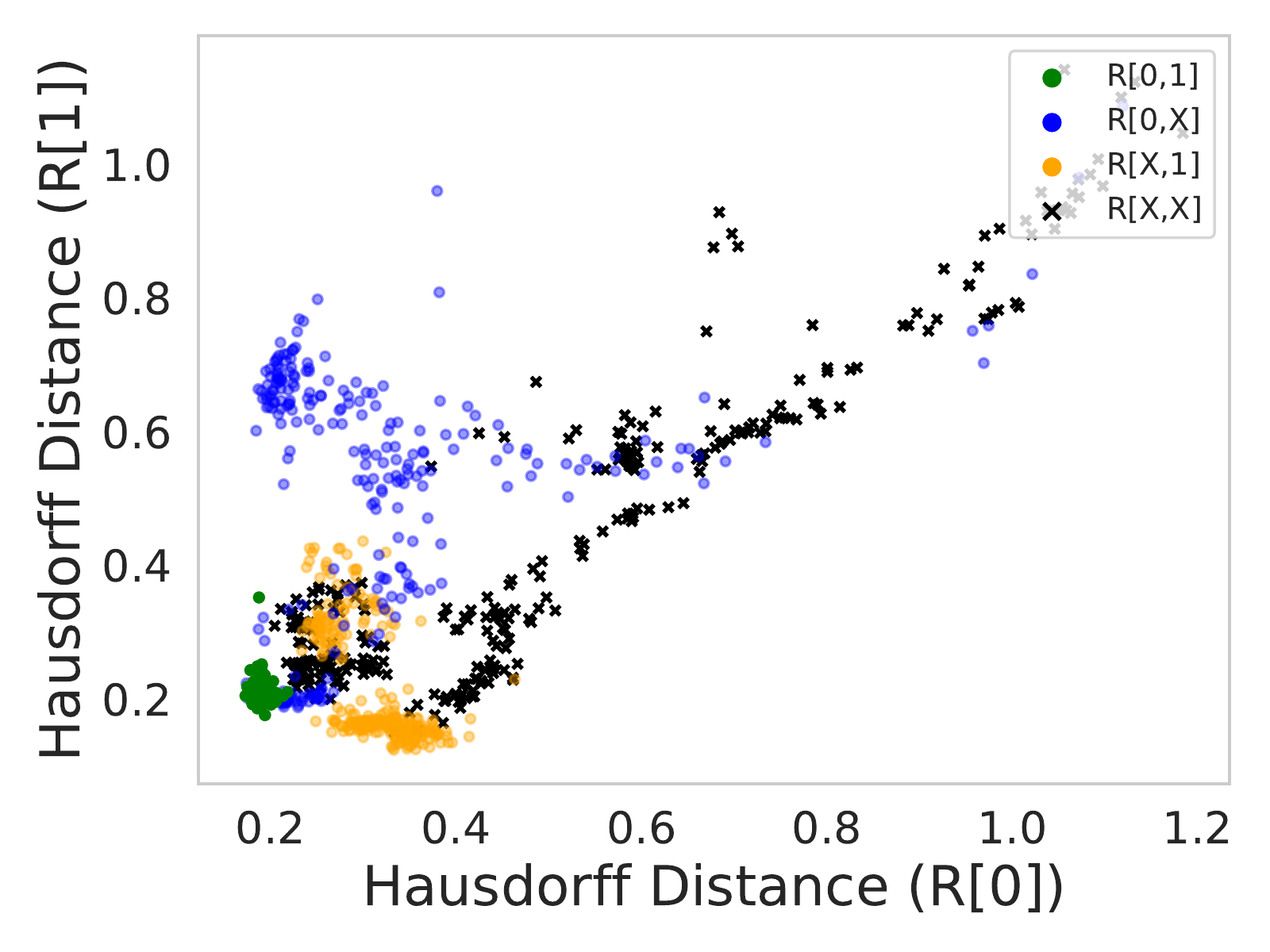}    \\
     $\rho=0.041$ & $\rho=0.09$ & $\rho=0.094$                                                                              \\
     \includegraphics[width=0.3\textwidth]{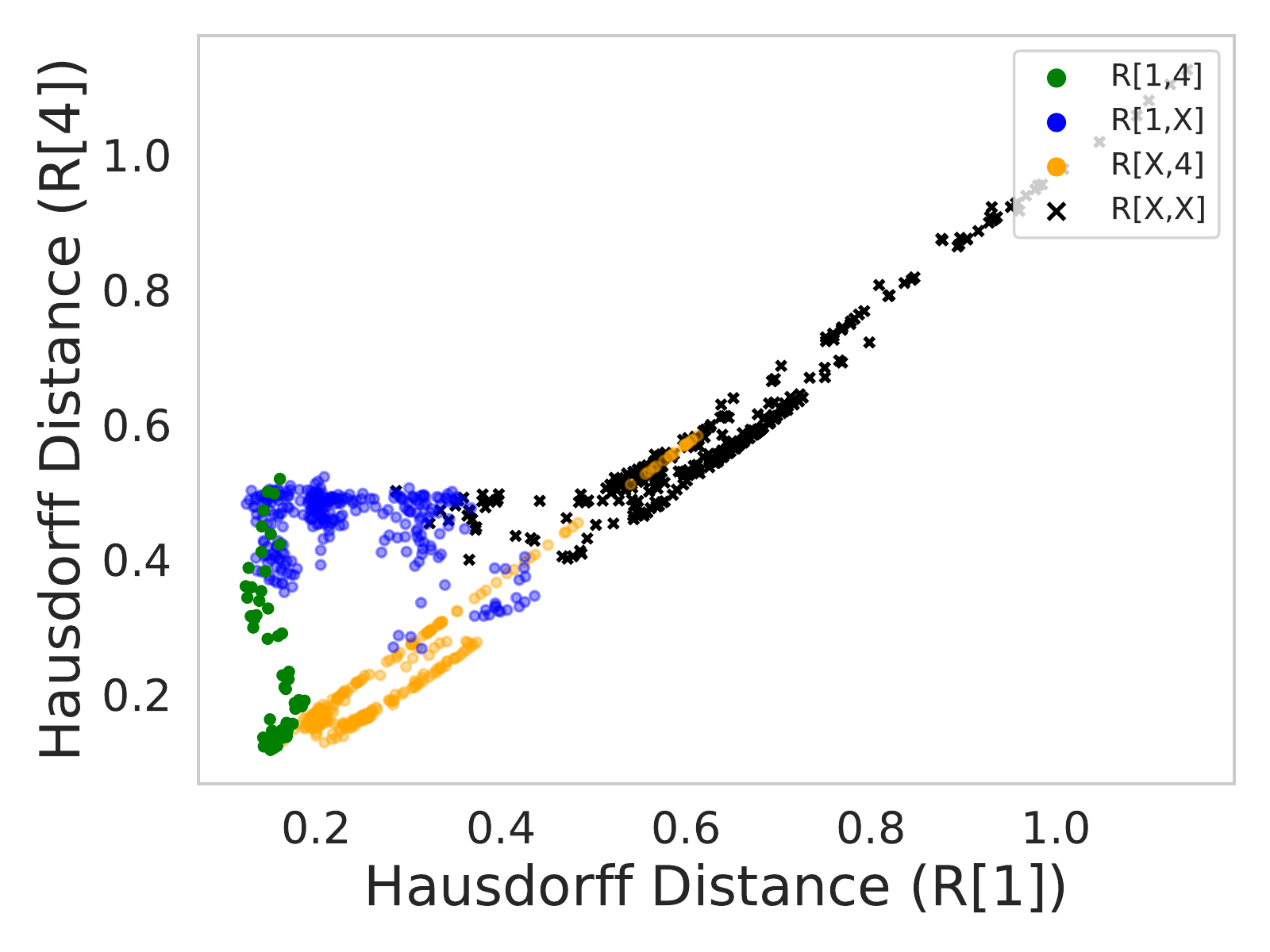} & 
     \includegraphics[width=0.3\textwidth]{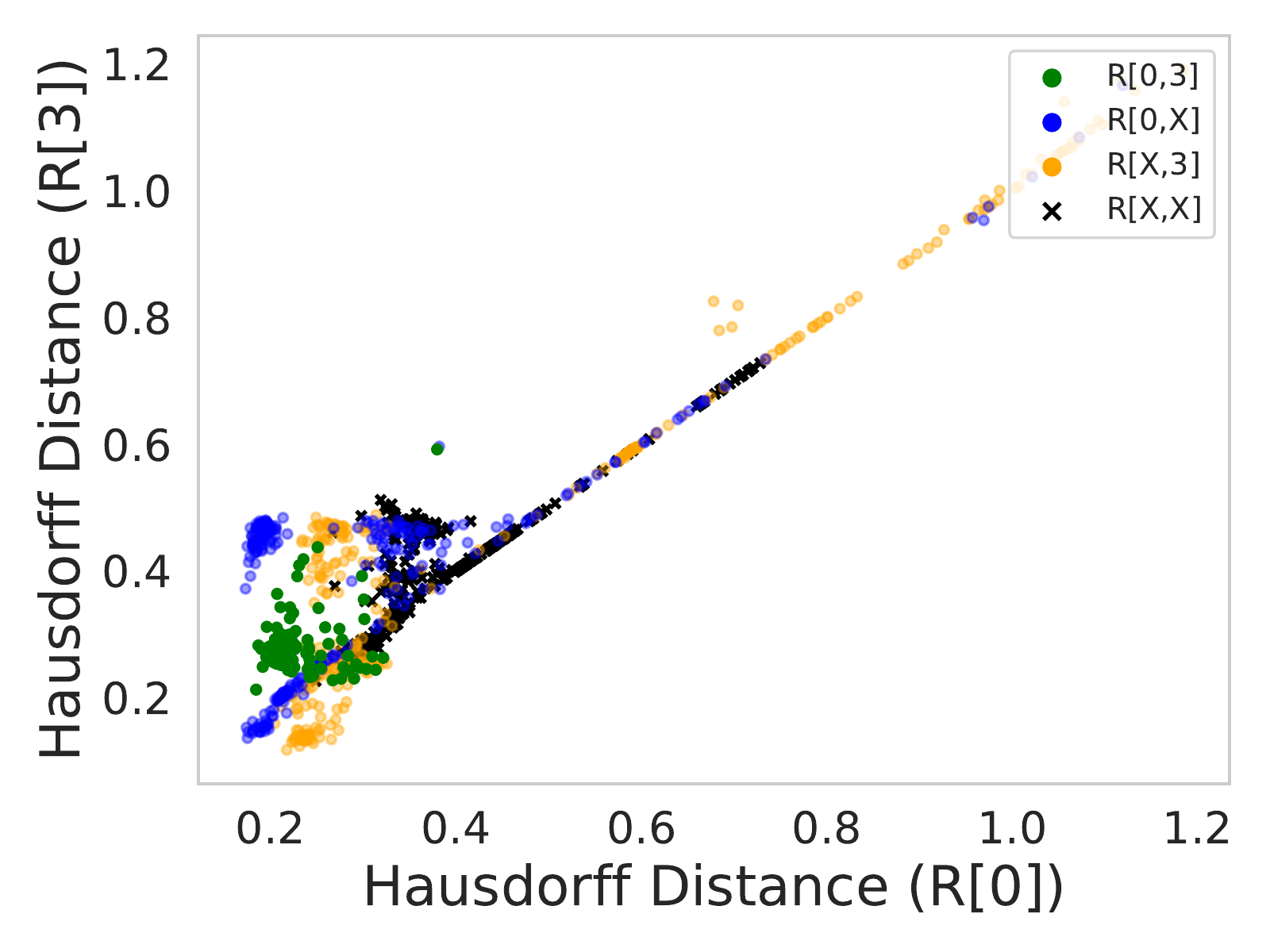} &  \includegraphics[width=0.3\textwidth]{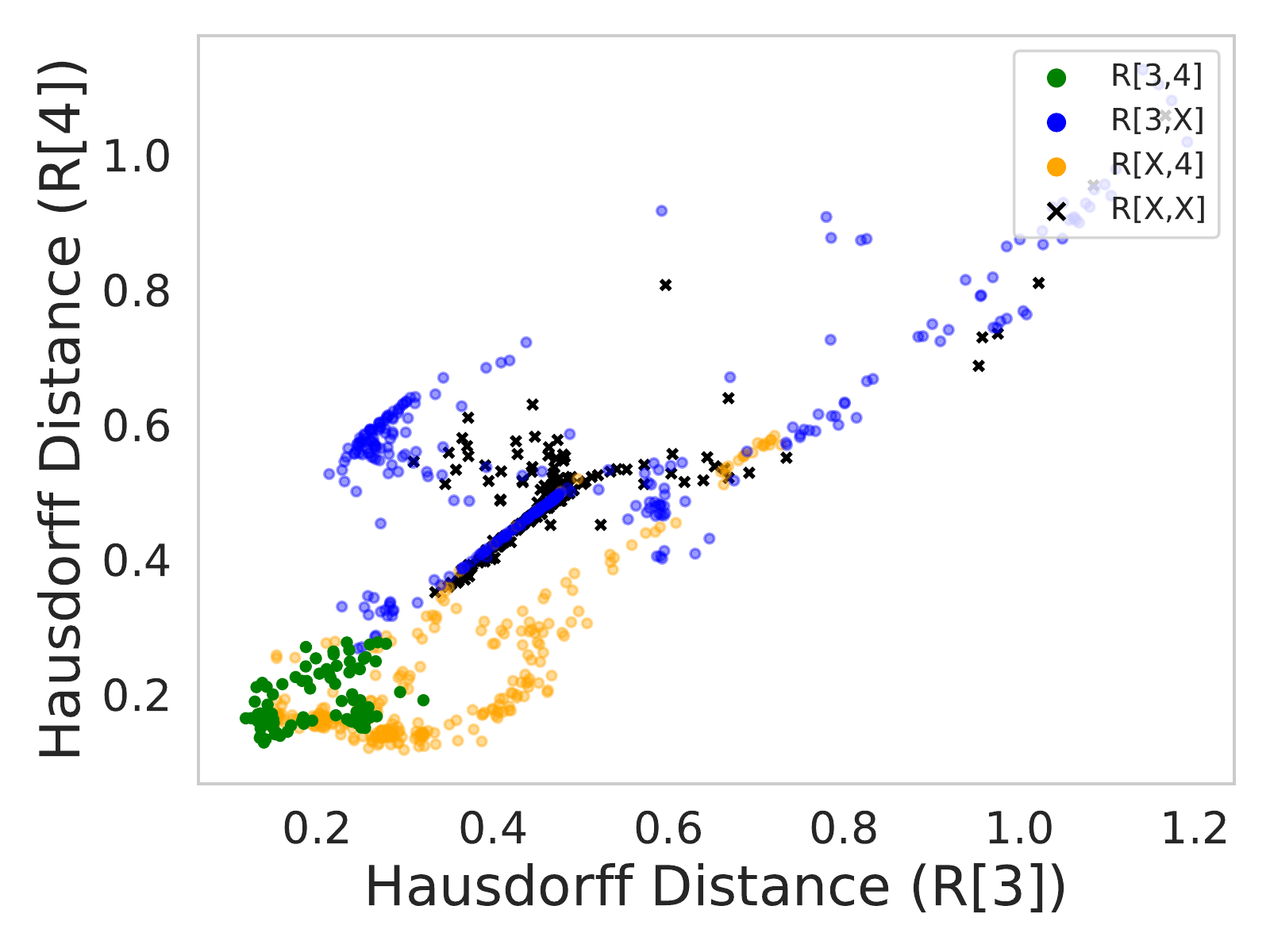}    \\
     $\rho=0.1$ & $\rho=0.114$ & $\rho=0.133$                                                                           \\
     \includegraphics[width=0.3\textwidth]{figures/onehot_distance_plots/9_d=0.147.pdf} & &    \\
     $\rho=0.147$ &  &                                                                                   \\
    
 \end{tabular}
    \caption{Topographic maps and their associated topographic scores for each combination of features with one-hot referents}
    \label{fig:sup_topo_one_hot}
\end{figure}

\newpage

\subsubsection{Visual - Shared Perspectives}

\begin{figure}[!h]
 \begin{tabular}{ccc}

     \includegraphics[width=0.3\textwidth]{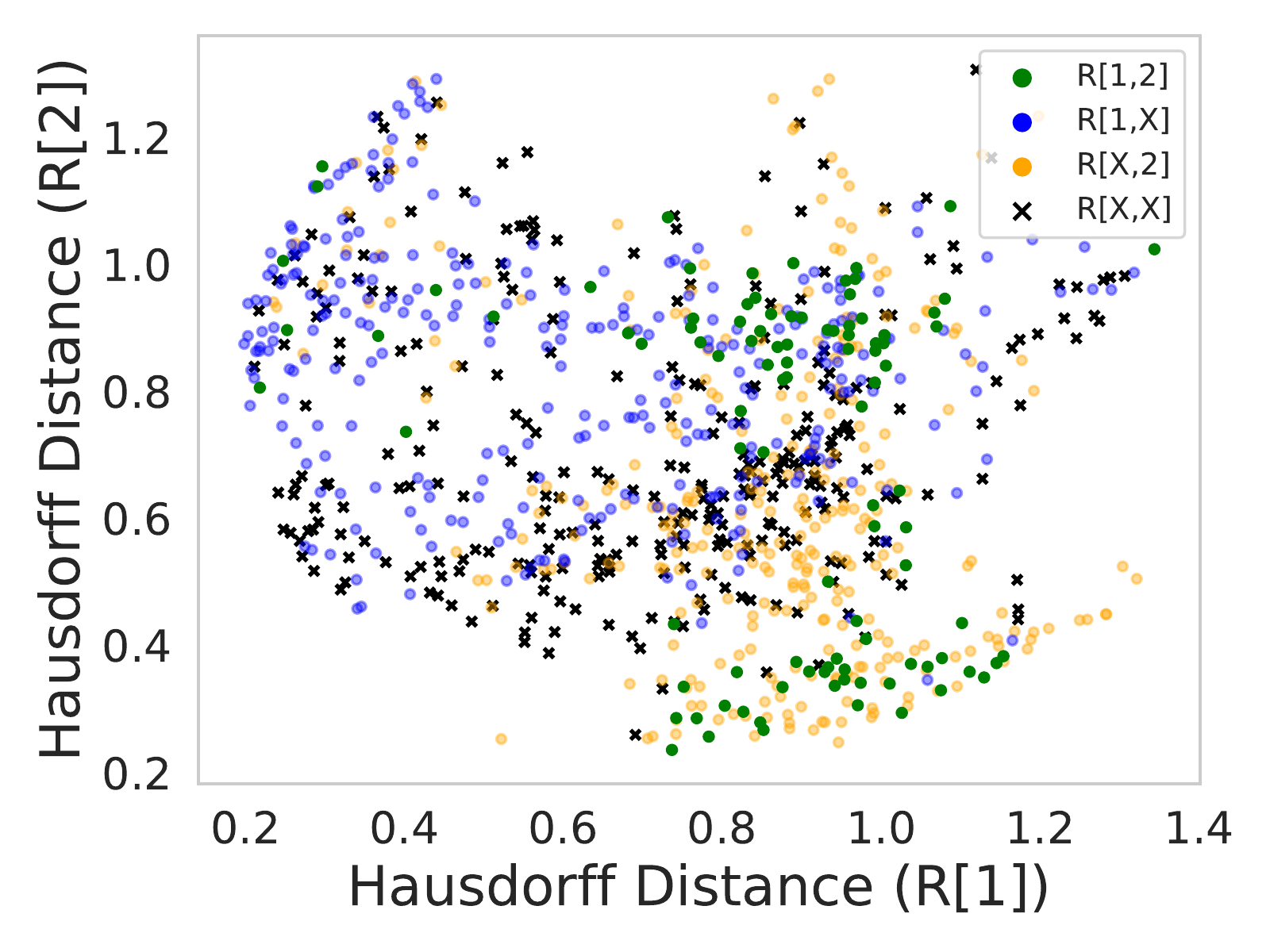} & \includegraphics[width=0.3\textwidth]{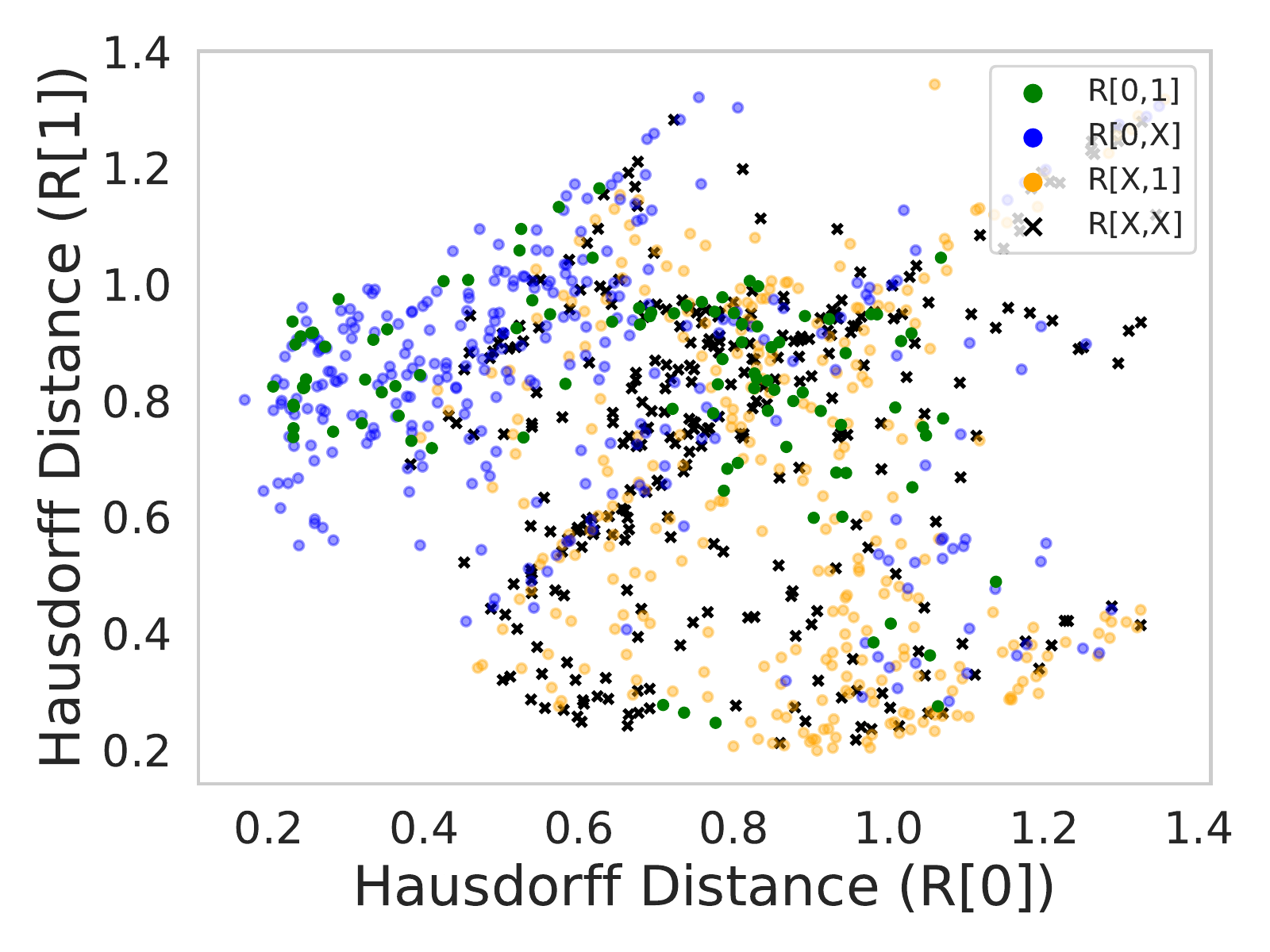} &  \includegraphics[width=0.3\textwidth]{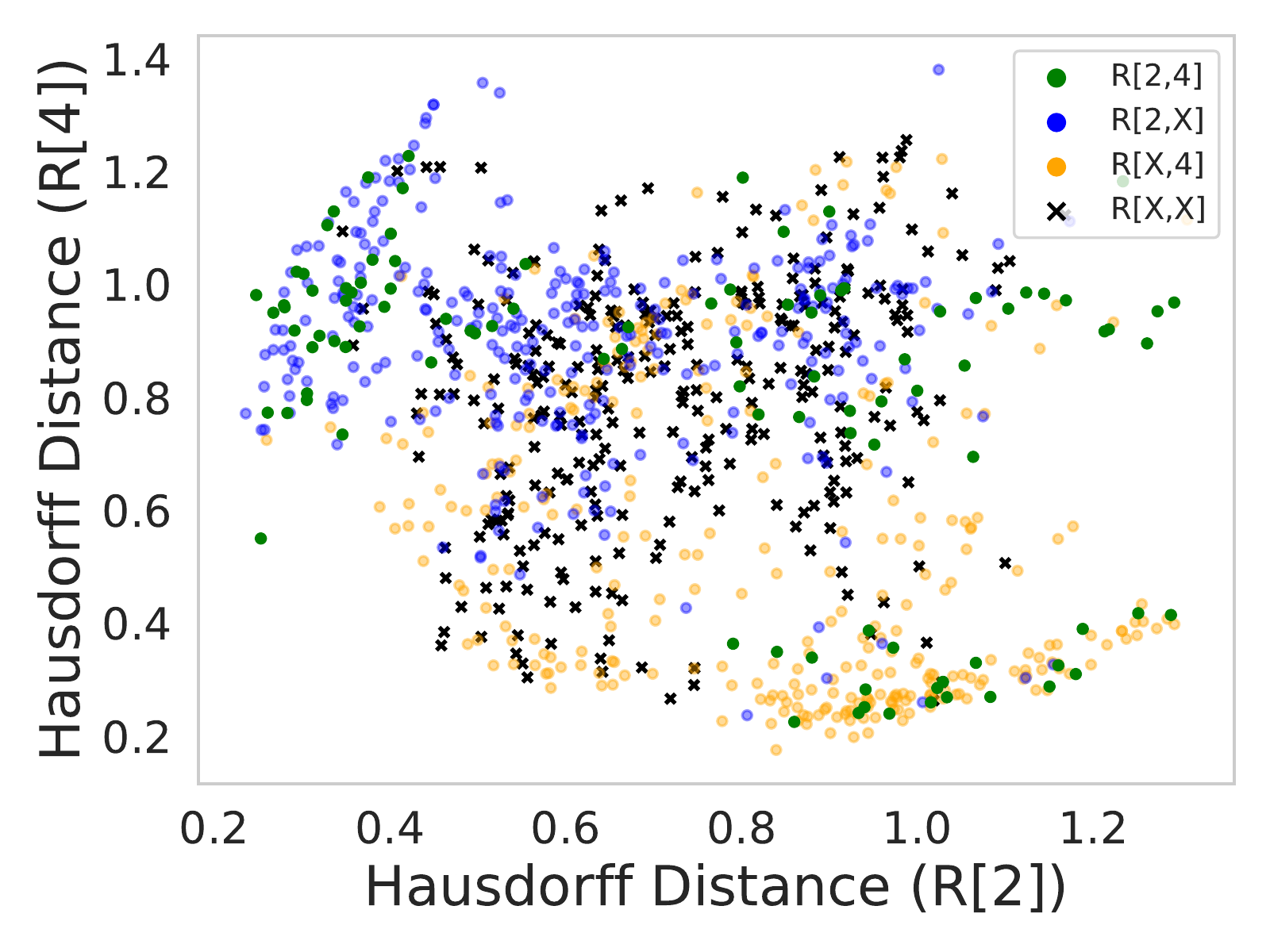}    \\
     $\rho=-0.047$ & $\rho=-0.044$ & $\rho=0.005$                                                                \\
     \includegraphics[width=0.3\textwidth]{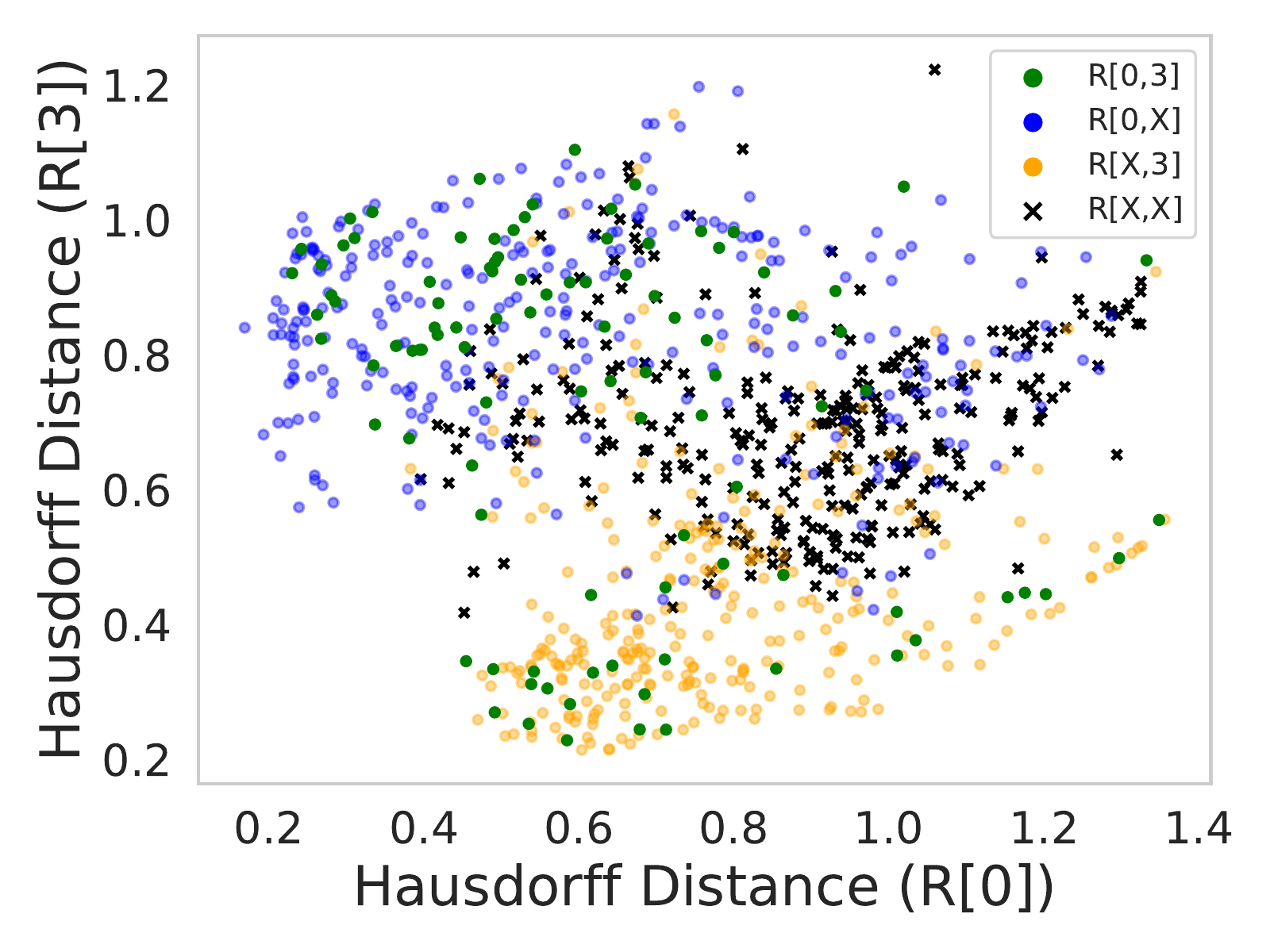} & \includegraphics[width=0.3\textwidth]{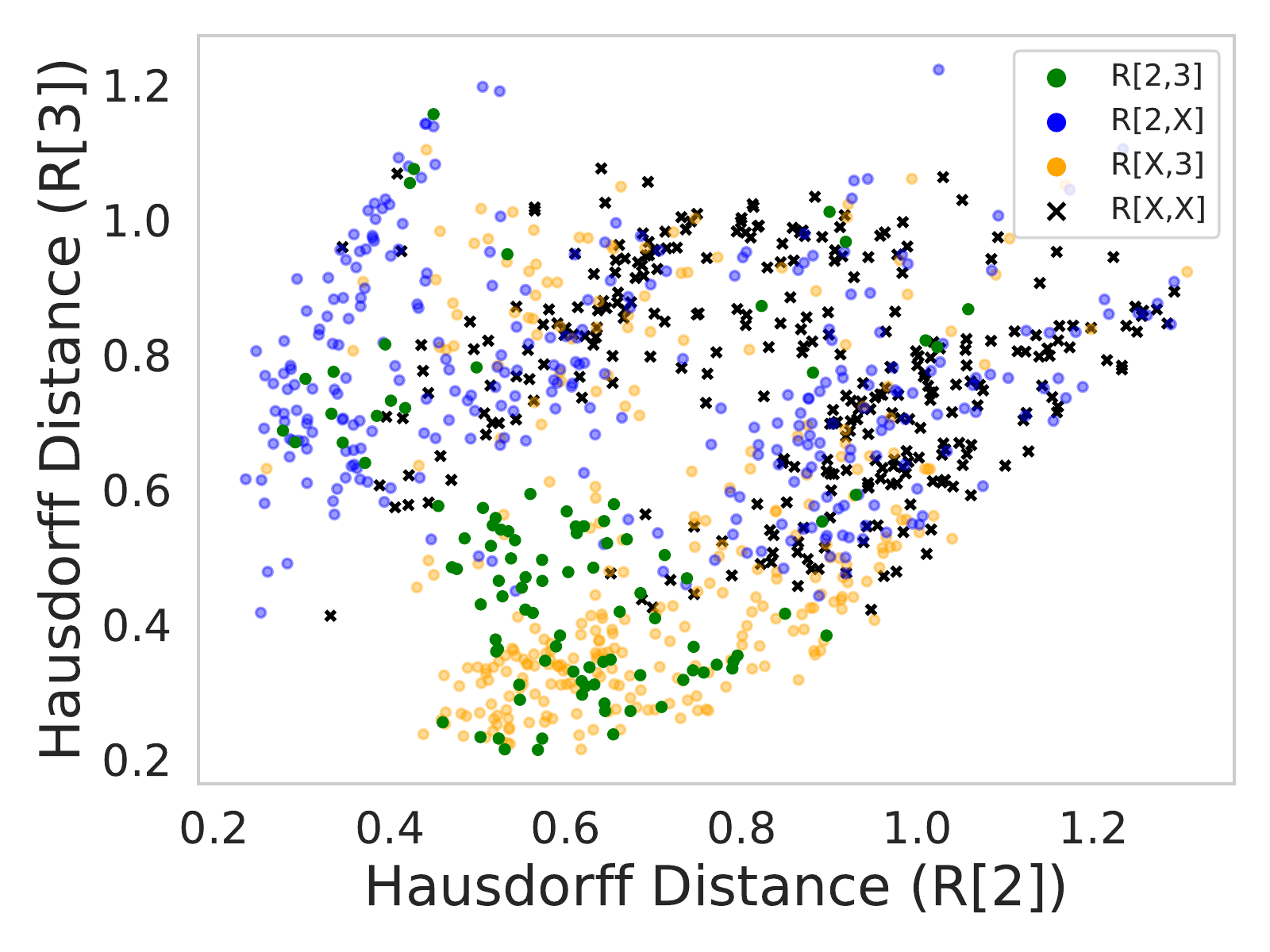} &  \includegraphics[width=0.3\textwidth]{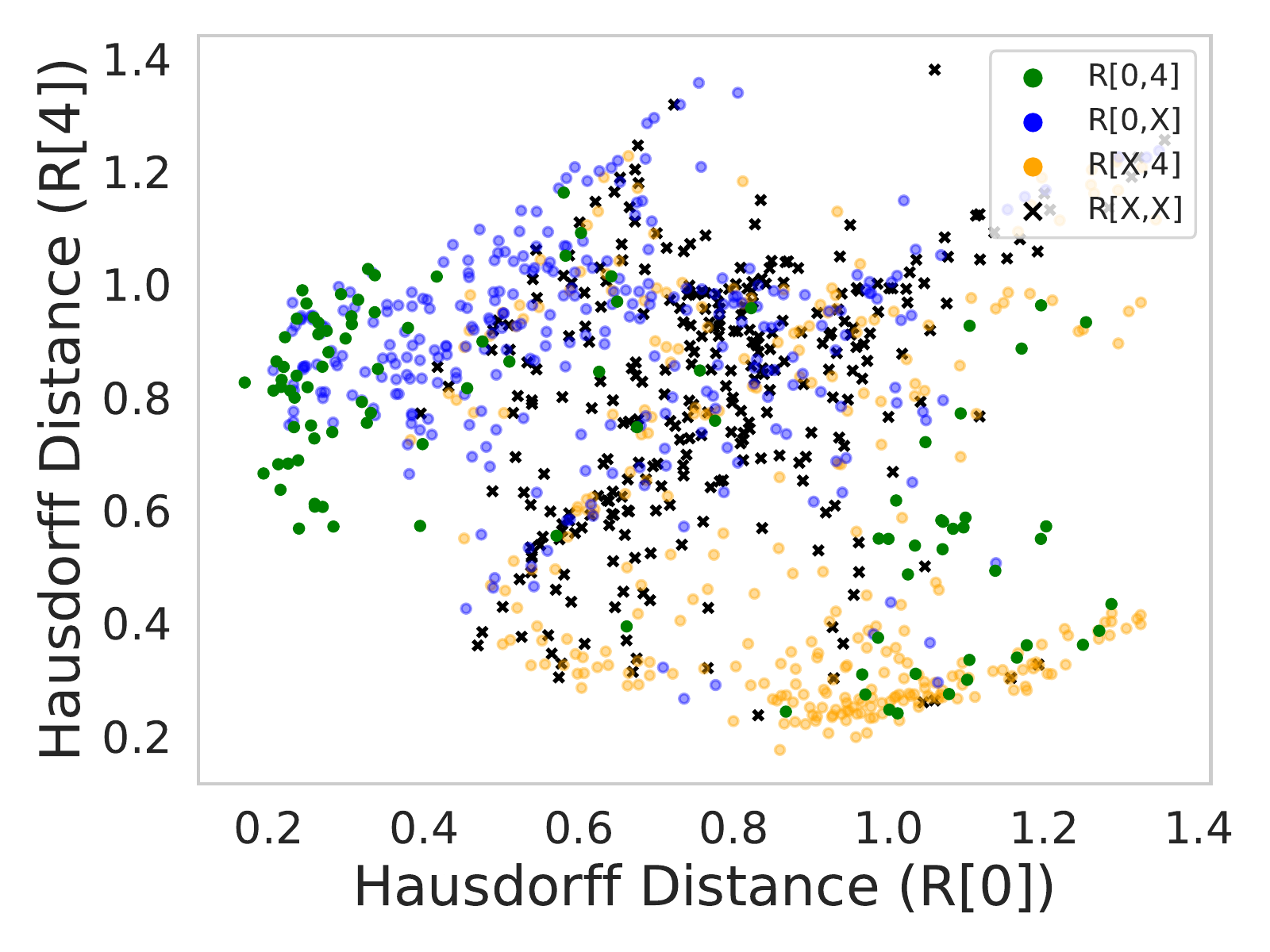}    \\
     $\rho=0.073$ & $\rho=0.081$ & $\rho=0.137$                                                                             \\
     \includegraphics[width=0.3\textwidth]{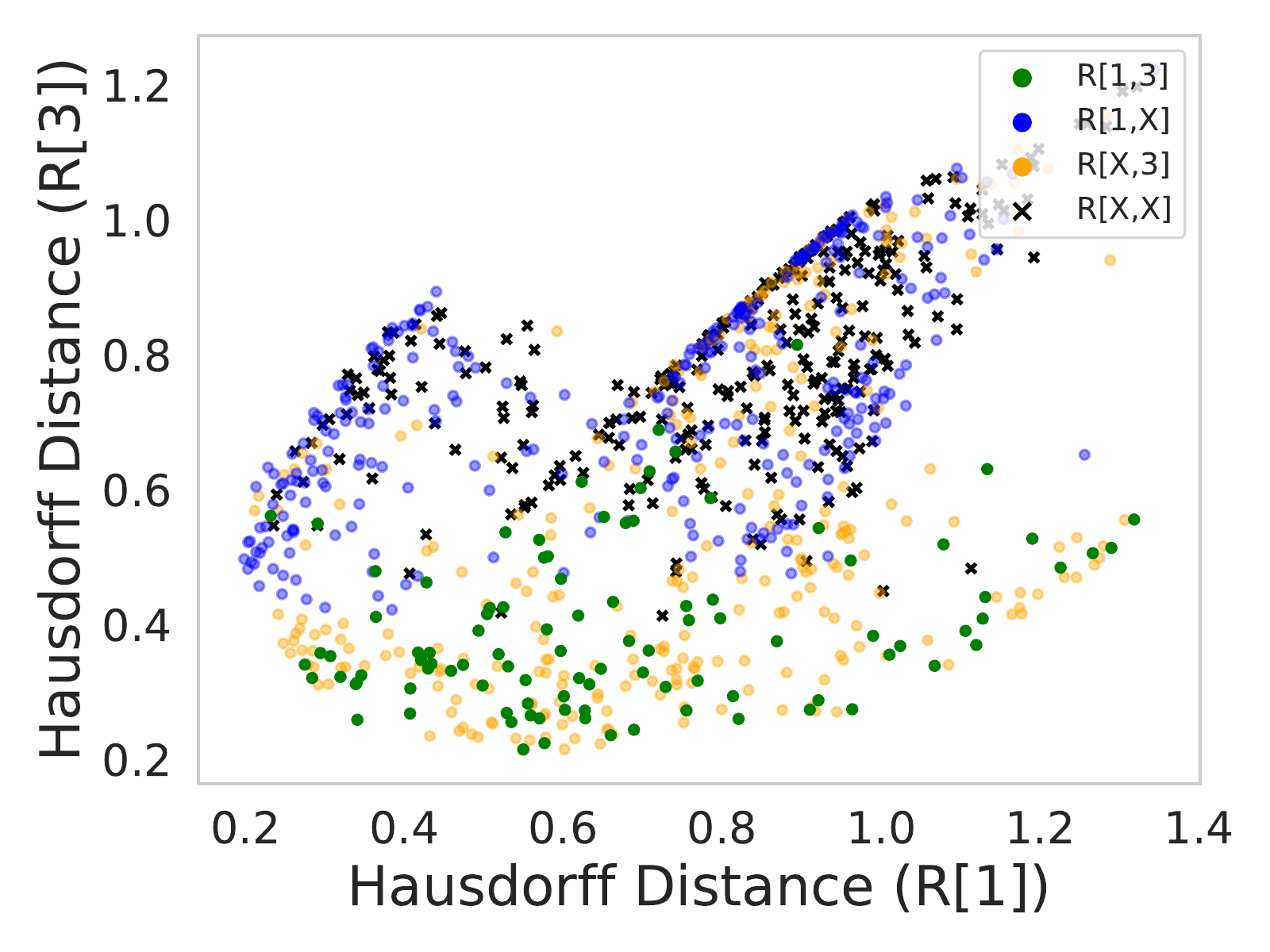} & 
     \includegraphics[width=0.3\textwidth]{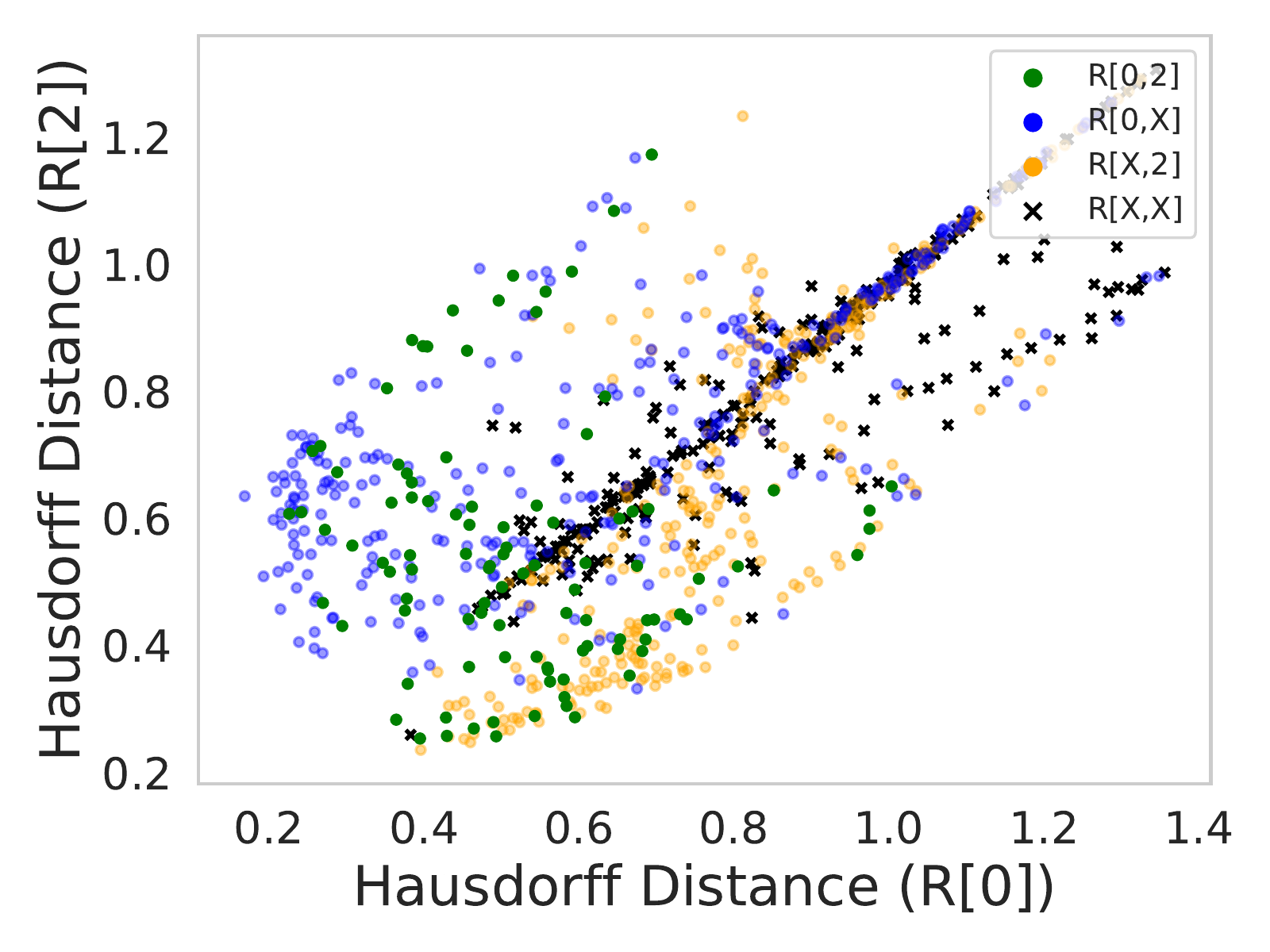} &  \includegraphics[width=0.3\textwidth]{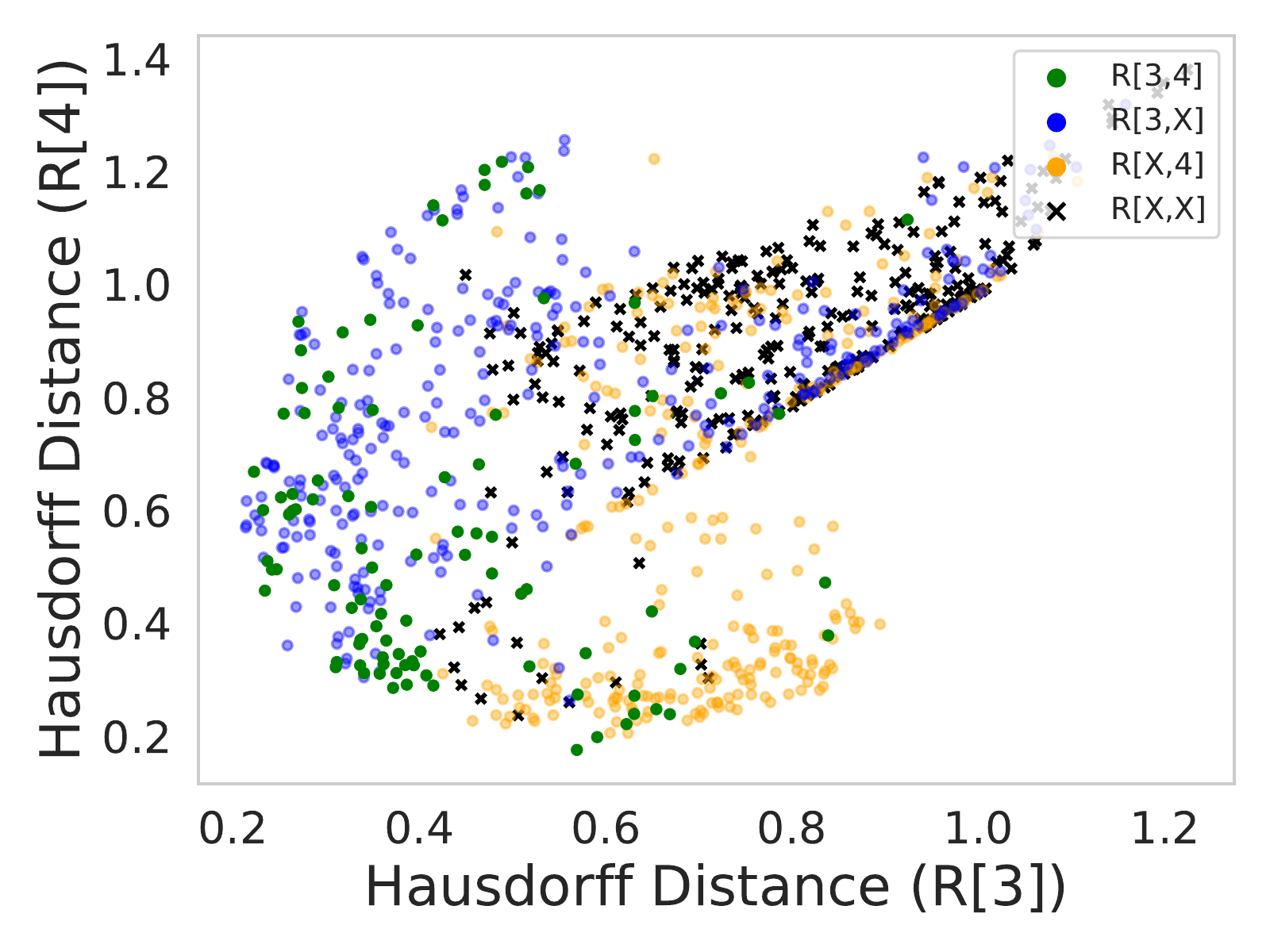}    \\
     $\rho=0.16$ & $\rho=0.215$ & $\rho=0.242$                                                                              \\
     \includegraphics[width=0.3\textwidth]{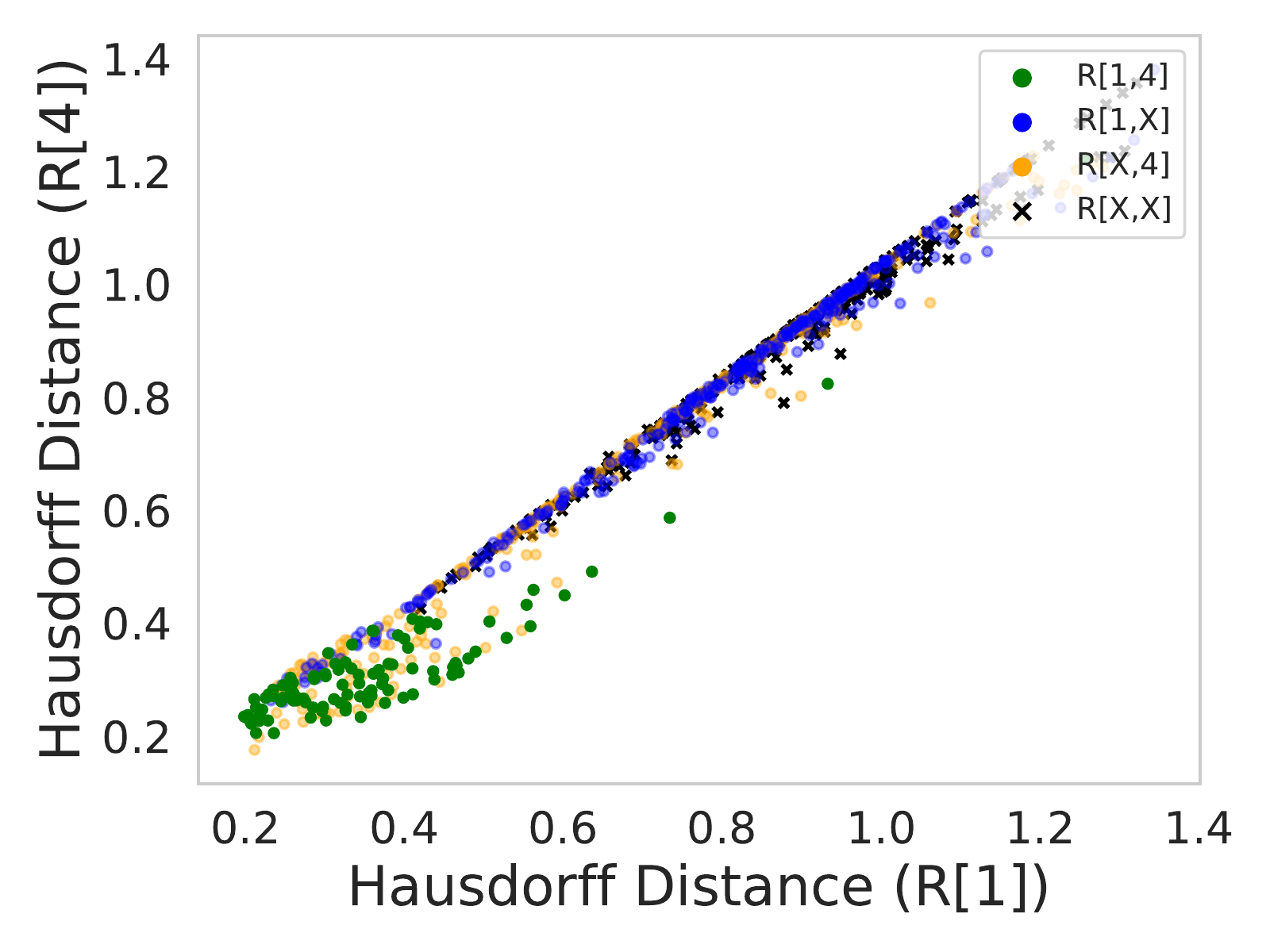} & &    \\
     $\rho=0.501$ &  &                                                                                   \\
    
 \end{tabular}
     \caption{Topographic maps and their associated topographic scores for each combination of features with shared-visual referents}
\label{fig:sup_topo_shared}
\end{figure}

\newpage

\subsubsection{Visual - Unshared Perspectives}

\begin{figure}[!h]
    \centering
 \begin{tabular}{ccc}

     \includegraphics[width=0.3\textwidth]{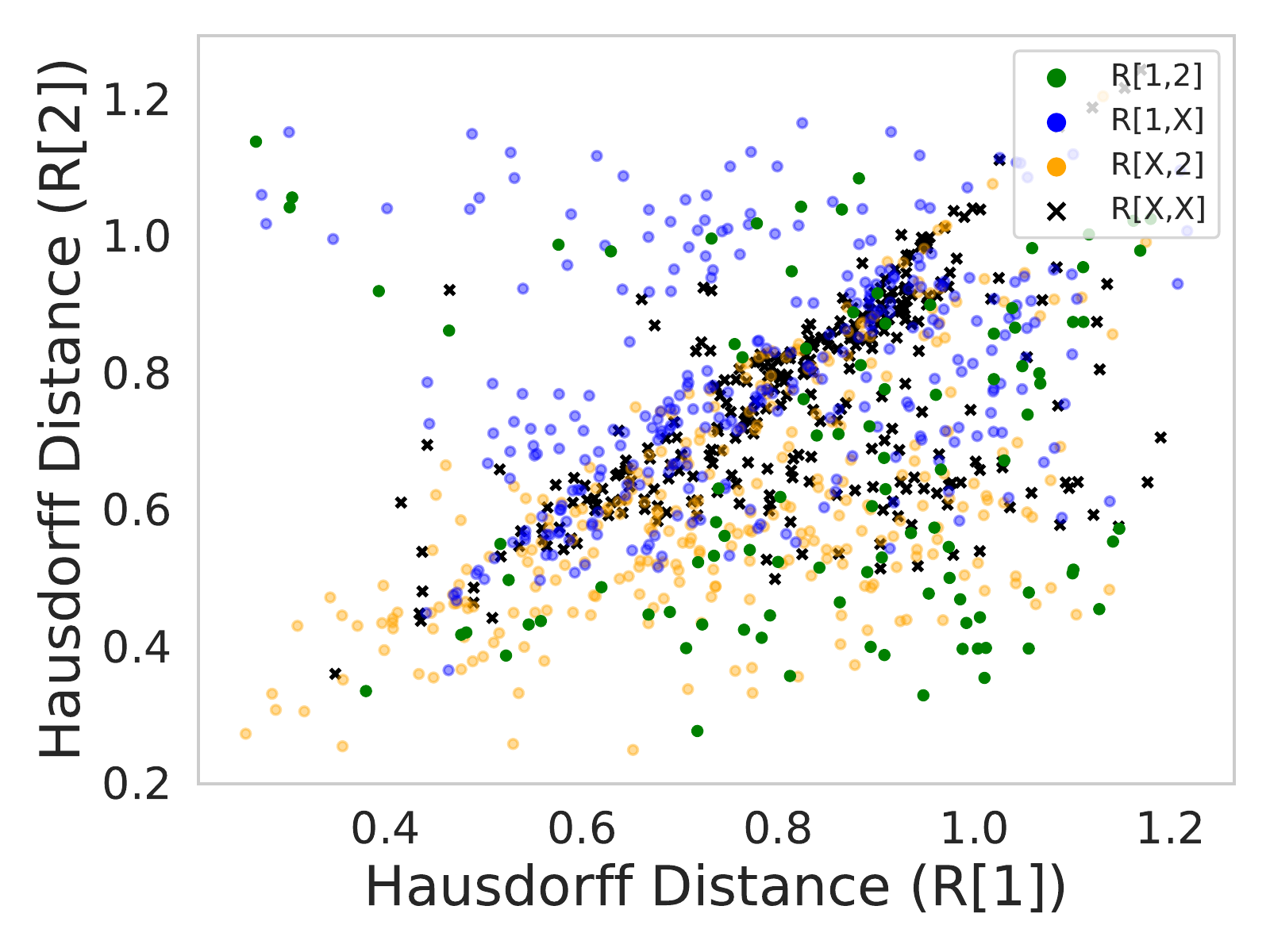} & \includegraphics[width=0.3\textwidth]{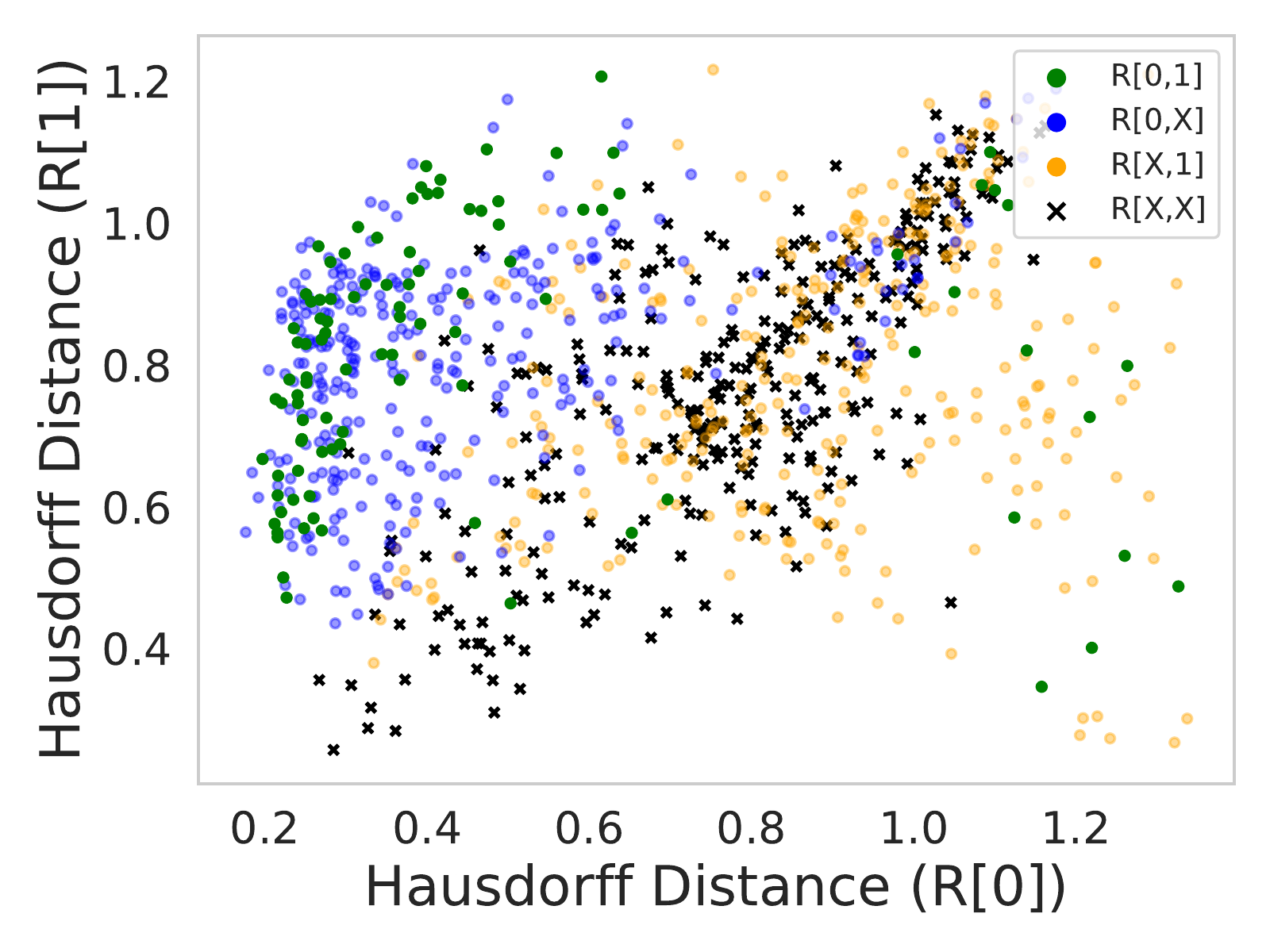} &  \includegraphics[width=0.3\textwidth]{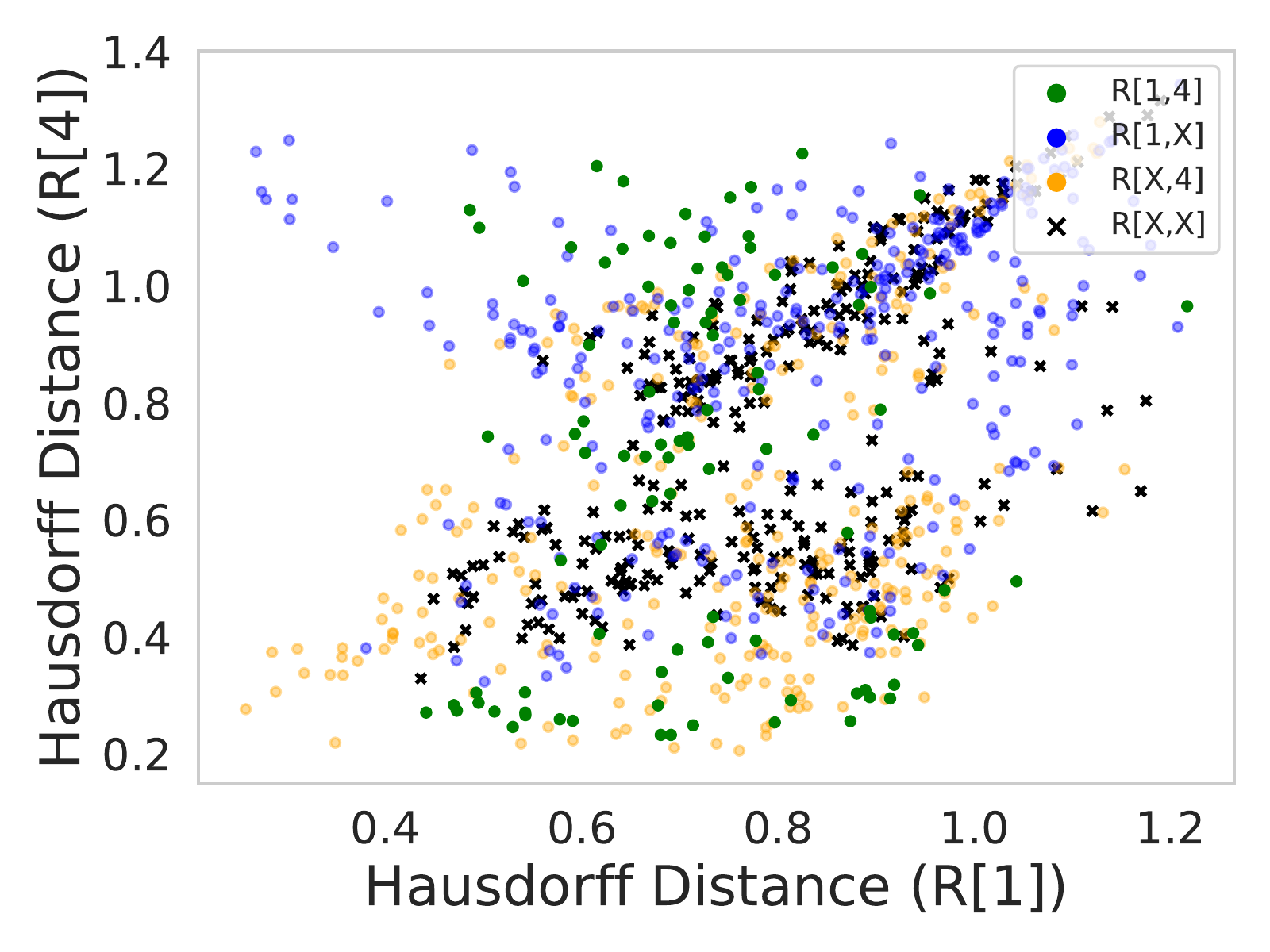}    \\
     $\rho=-0.113$ & $\rho=-0.016$ & $\rho=0$                                                                \\
     \includegraphics[width=0.3\textwidth]{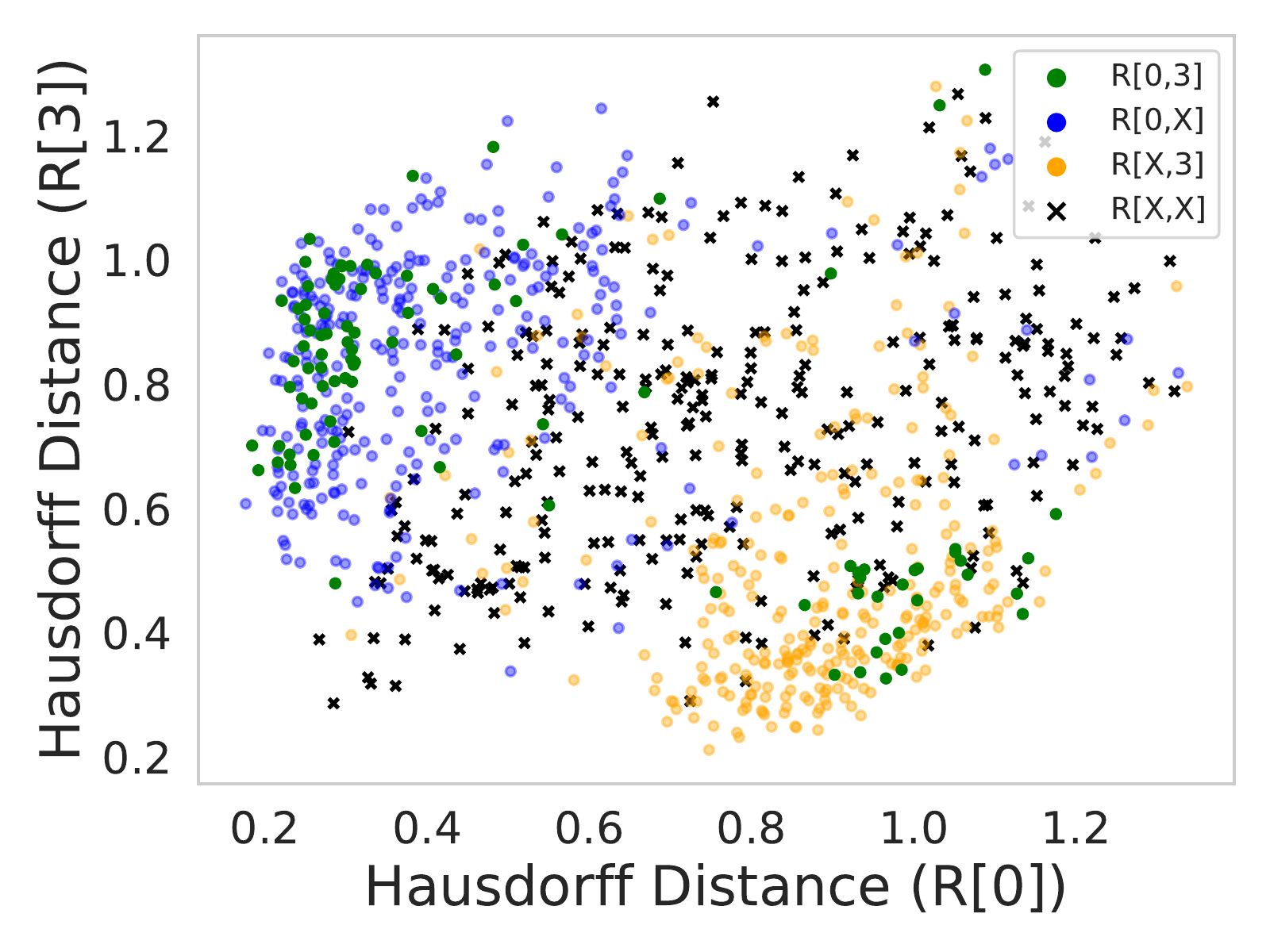} & \includegraphics[width=0.3\textwidth]{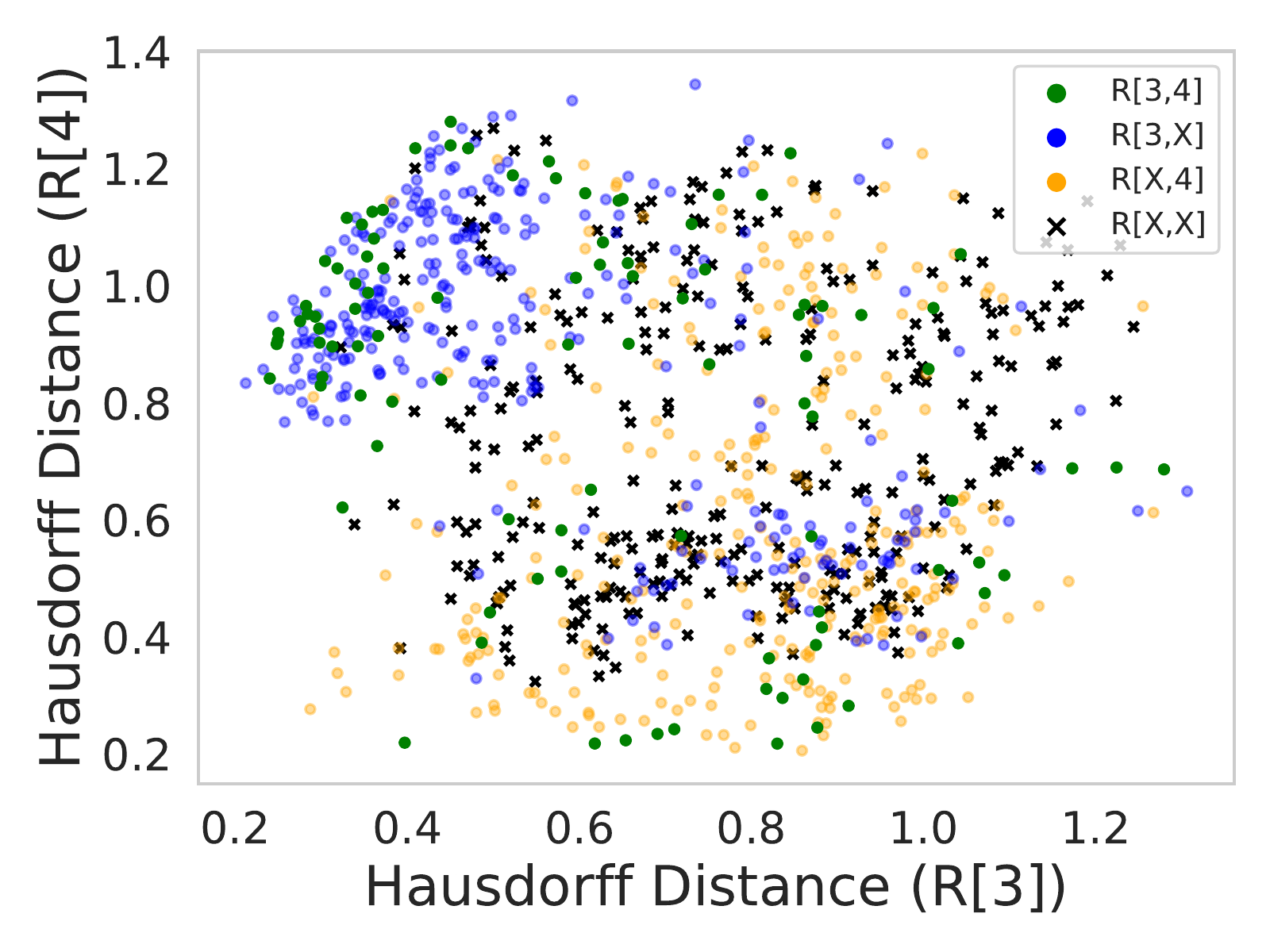} &  \includegraphics[width=0.3\textwidth]{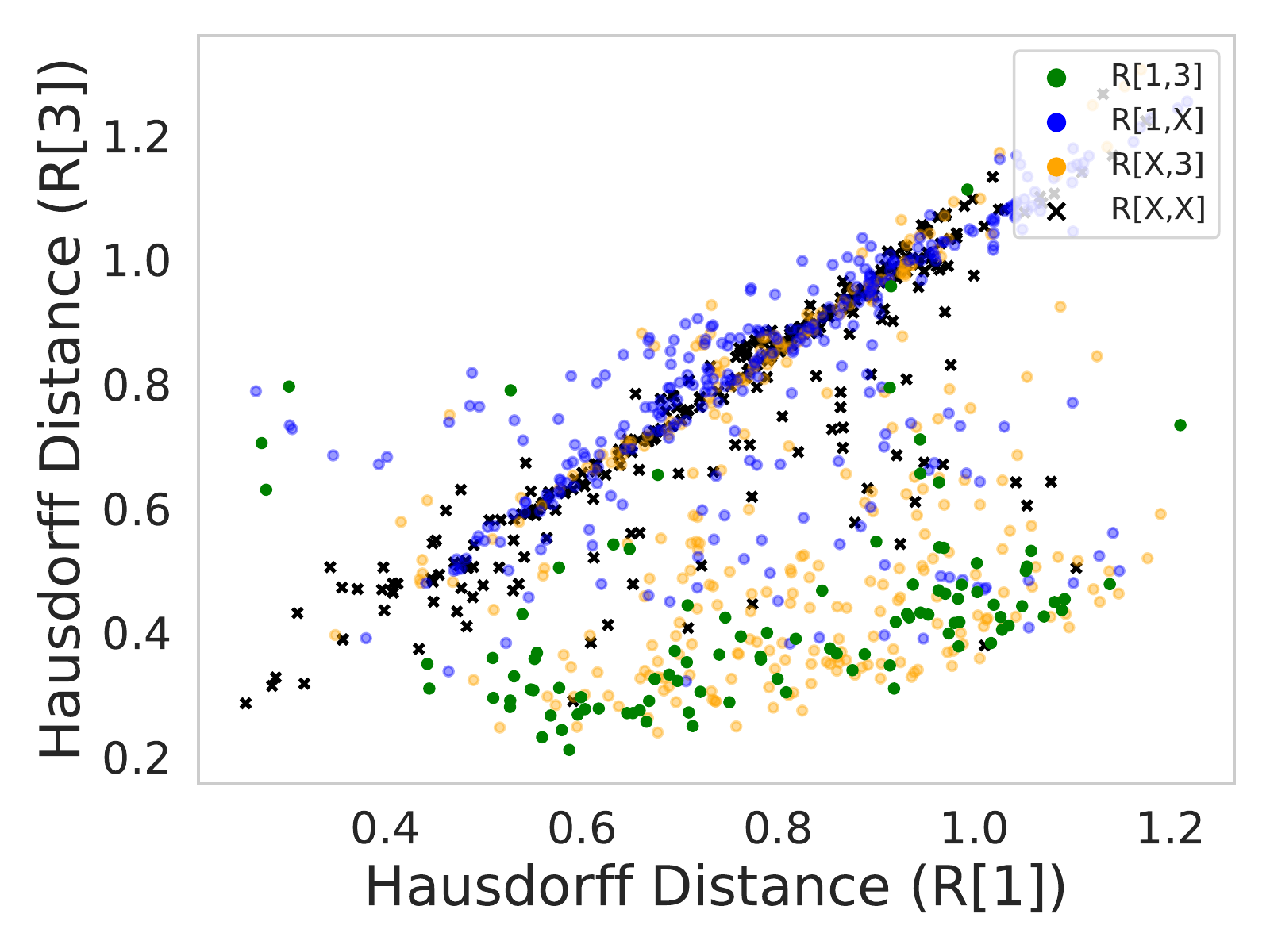}    \\
     $\rho=0.019$ & $\rho=0.026$ & $\rho=0.13$                                                                              \\
     \includegraphics[width=0.3\textwidth]{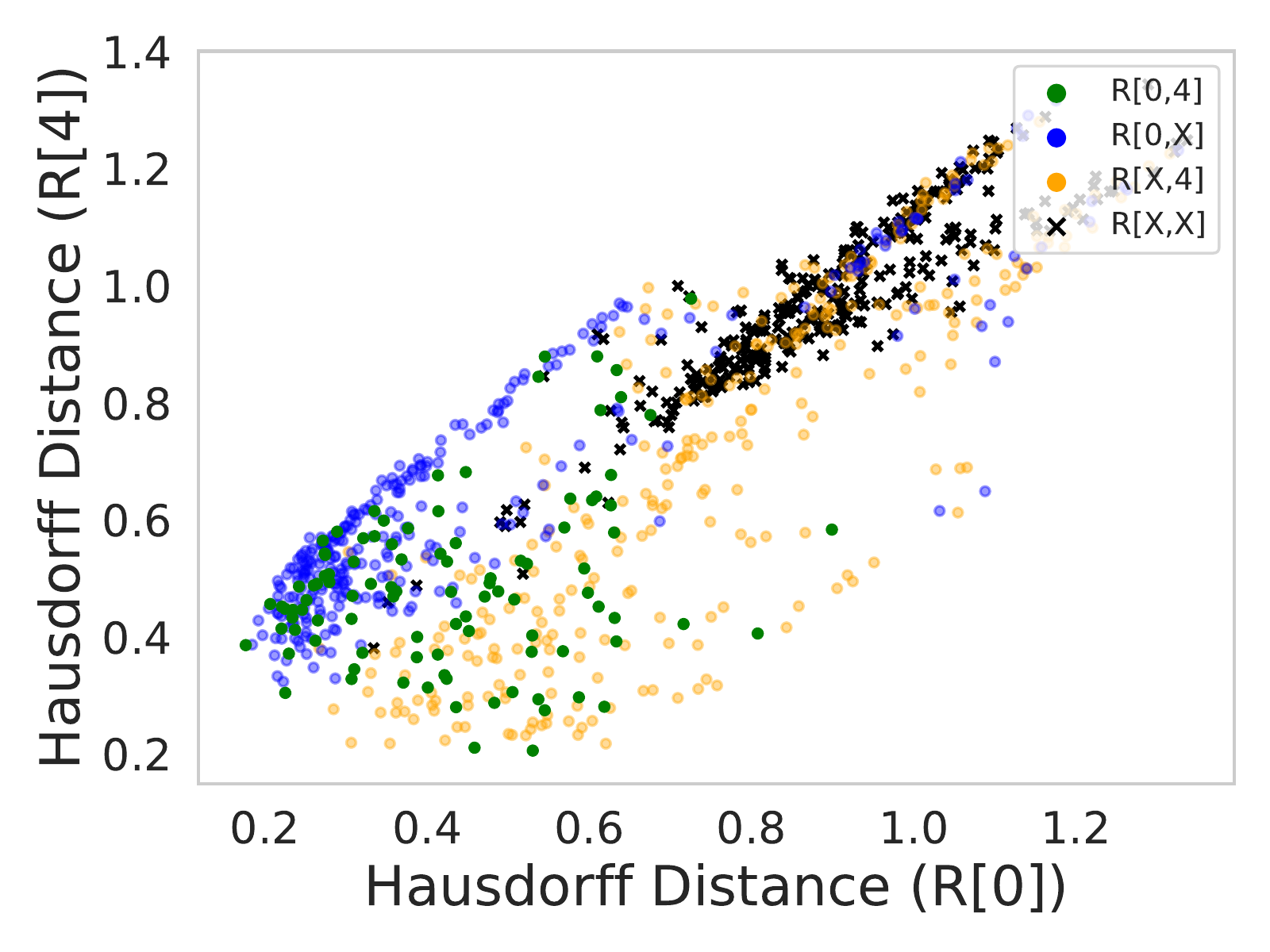} & 
     \includegraphics[width=0.3\textwidth]{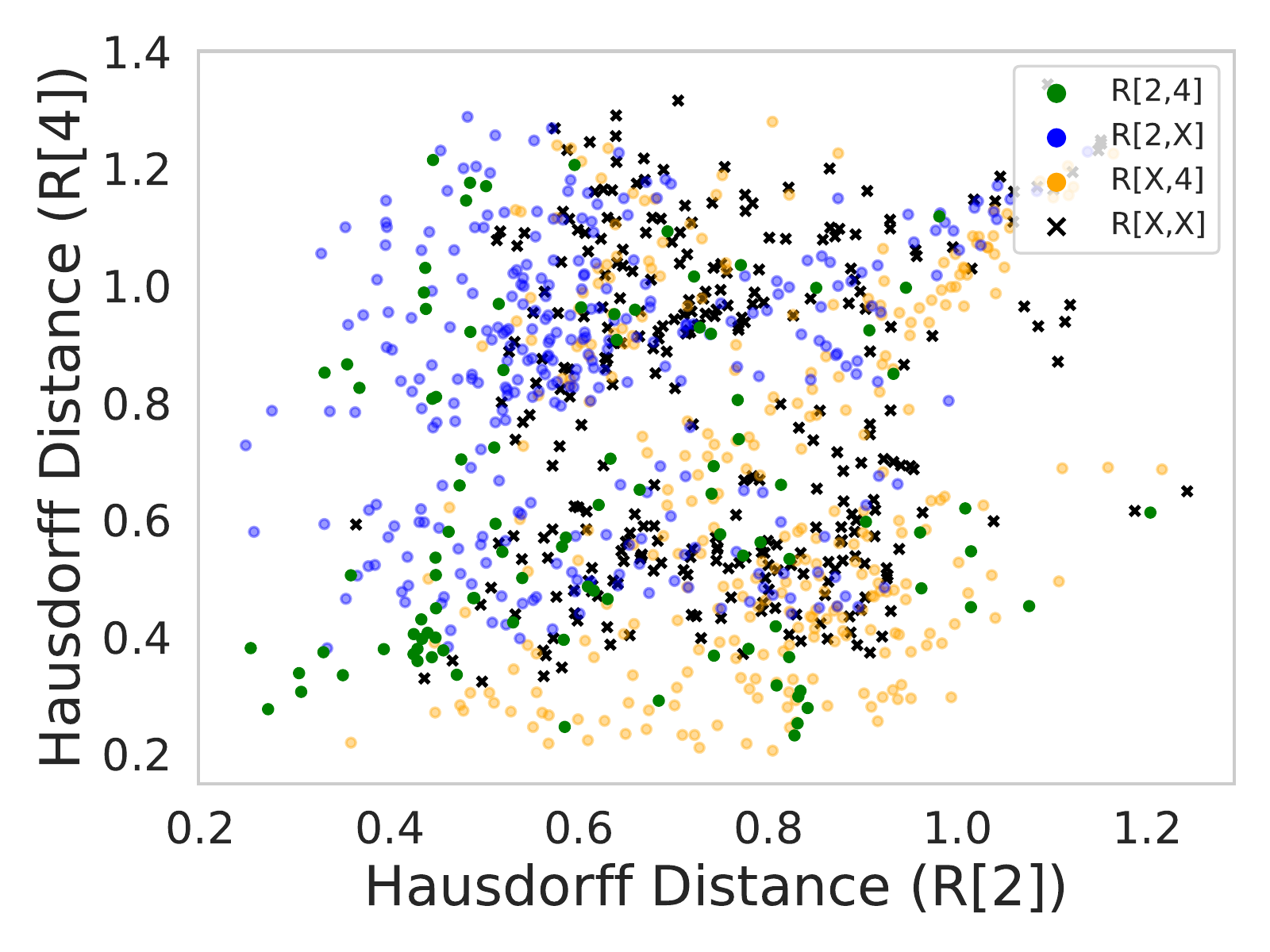} &  \includegraphics[width=0.3\textwidth]{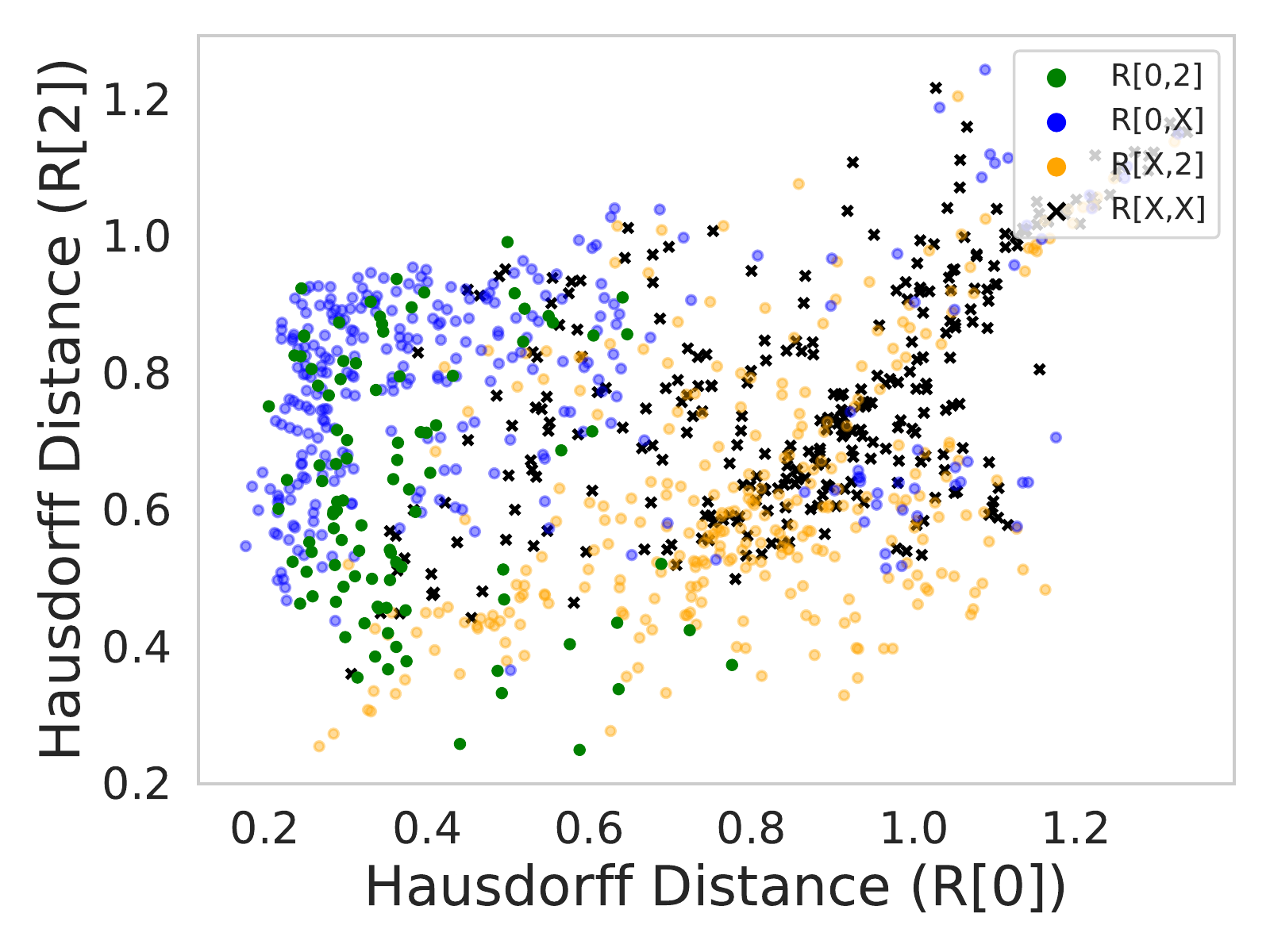}    \\
     $\rho=0.137$ & $\rho=0.156$ & $\rho=0.18$                                                                              \\
     \includegraphics[width=0.3\textwidth]{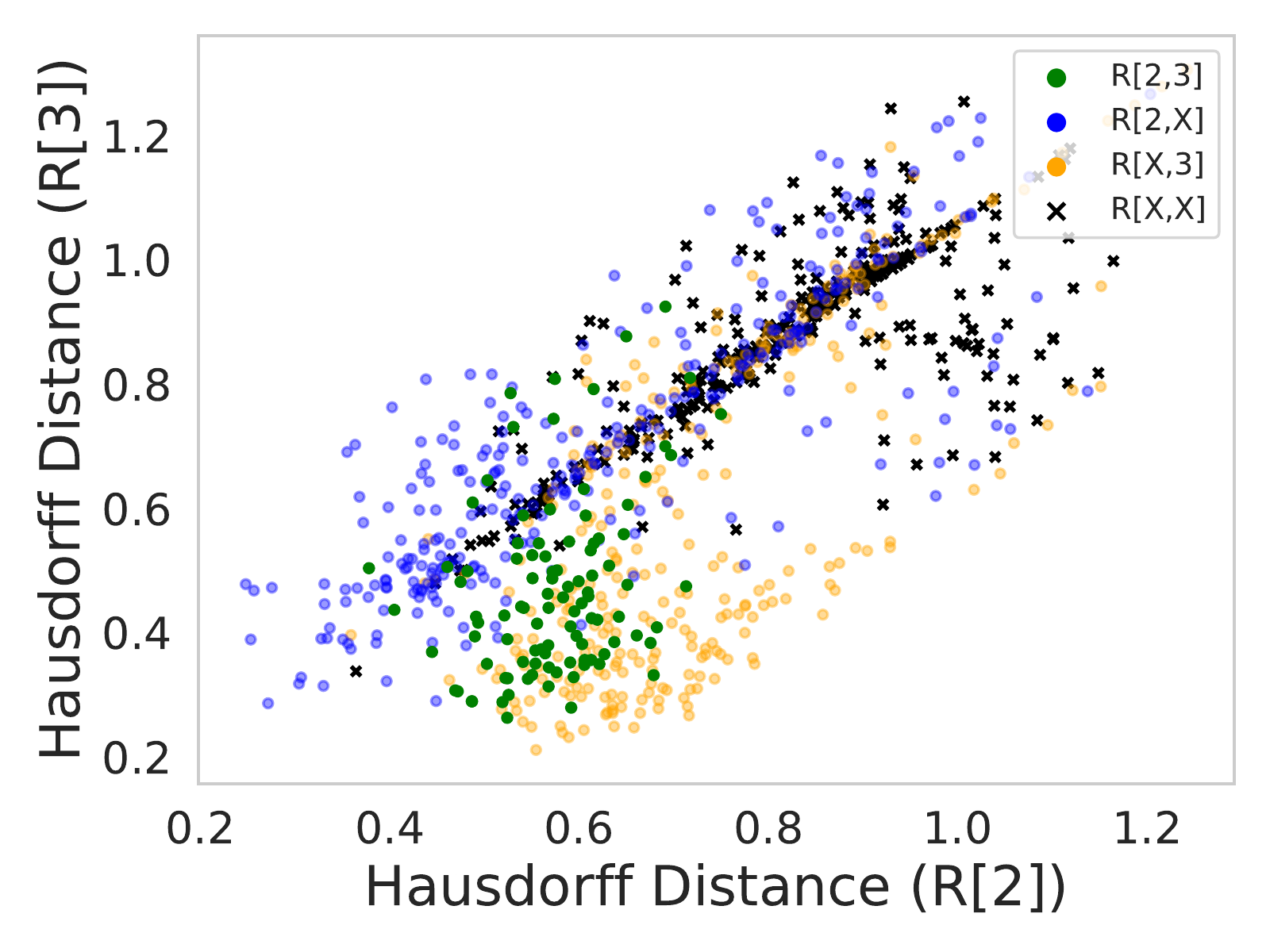} & &    \\
     $\rho=0.203$ &  &                                                                                   \\
    
 \end{tabular}
 
      \caption{Topographic maps and their associated topographic scores for each combination of features with unshared-visual referents}
\label{fig:sup_topo_unshared}
\end{figure}

\newpage

\subsection{Composition Matrix examples (Visual - Unshared Perspectives)}
\label{sup:compo_matr}
\begin{figure}[h!]
\centering 

 \begin{tabular}{cc}

     \includegraphics[width=0.4\textwidth]{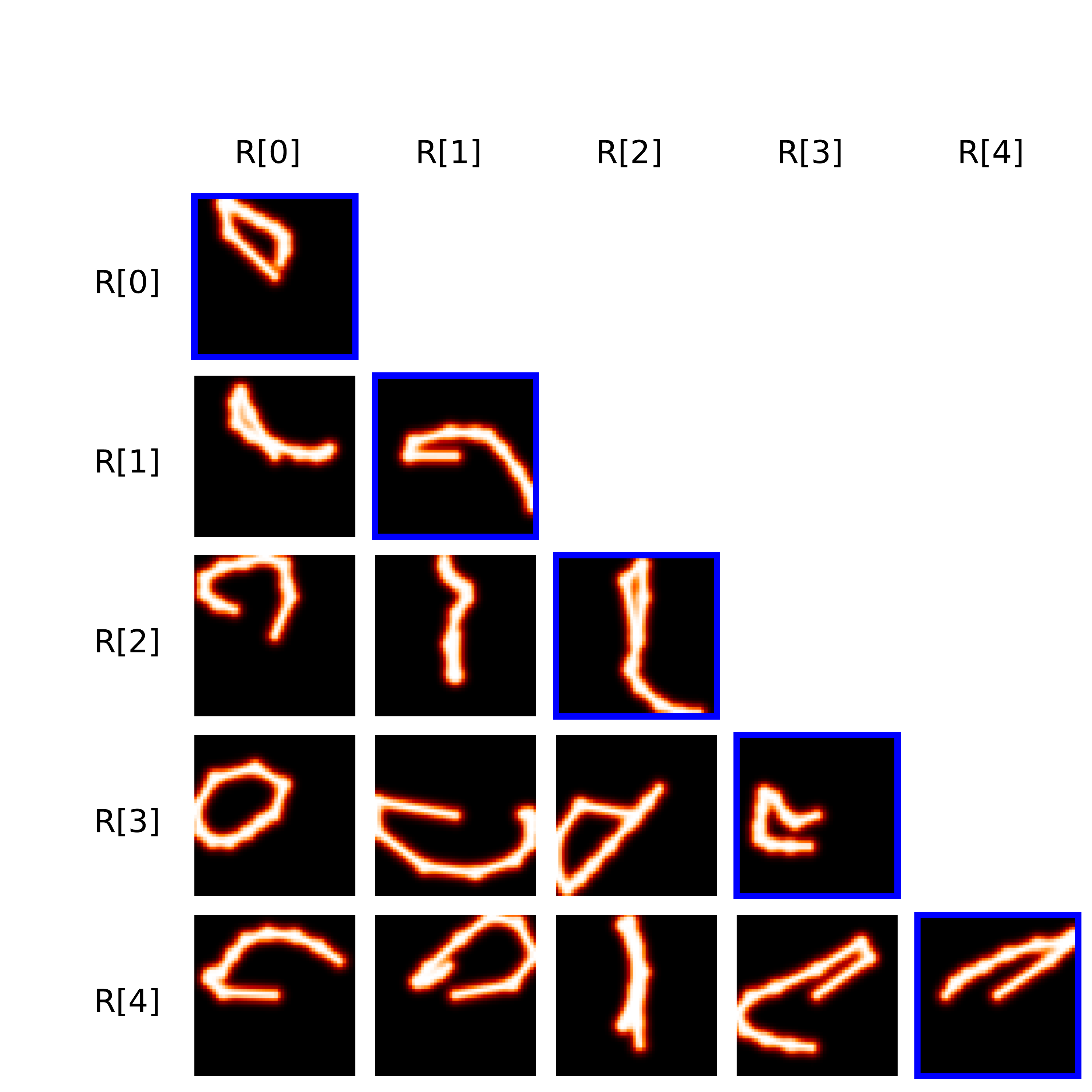} & 
     \includegraphics[width=0.4\textwidth]{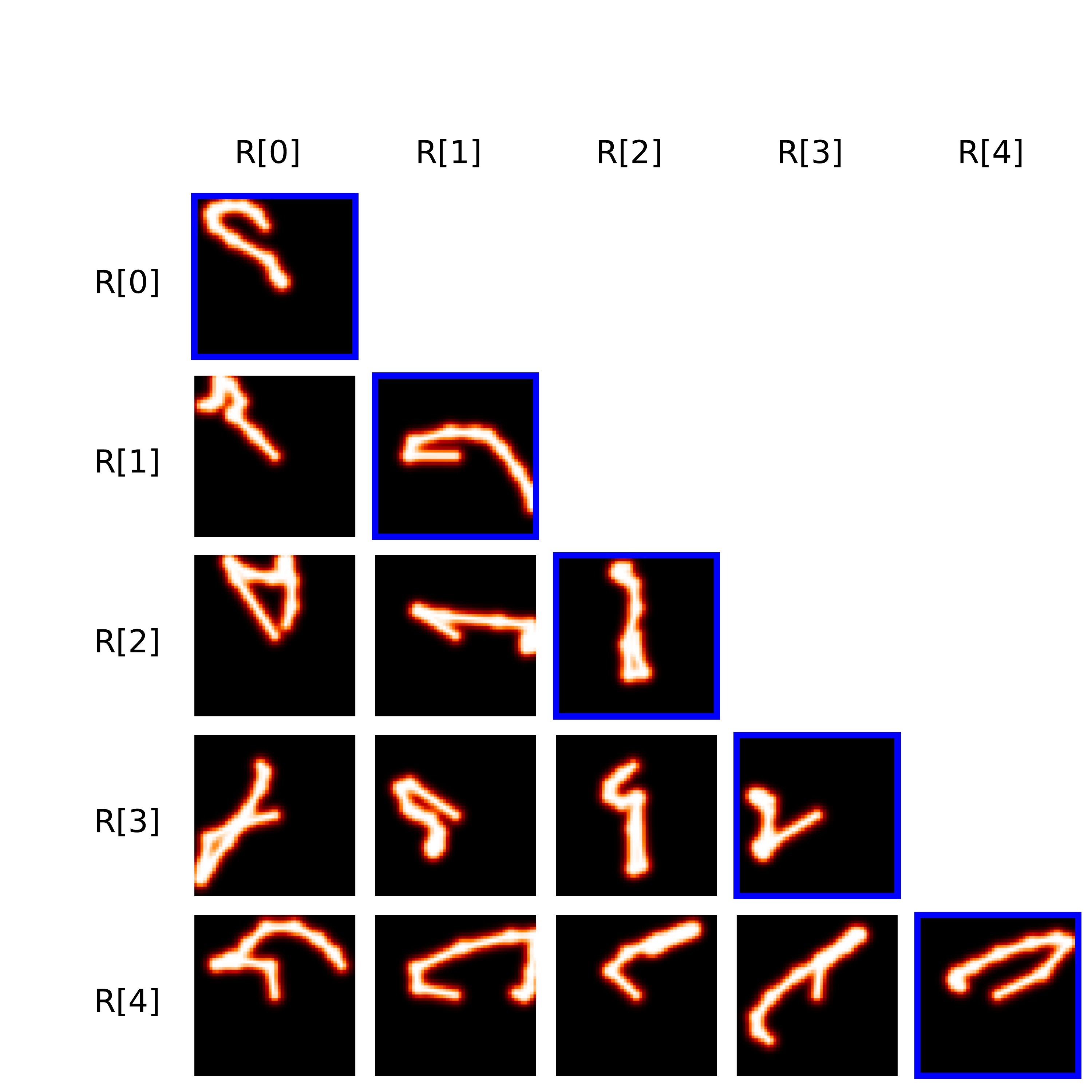} \\
     \includegraphics[width=0.4\textwidth]{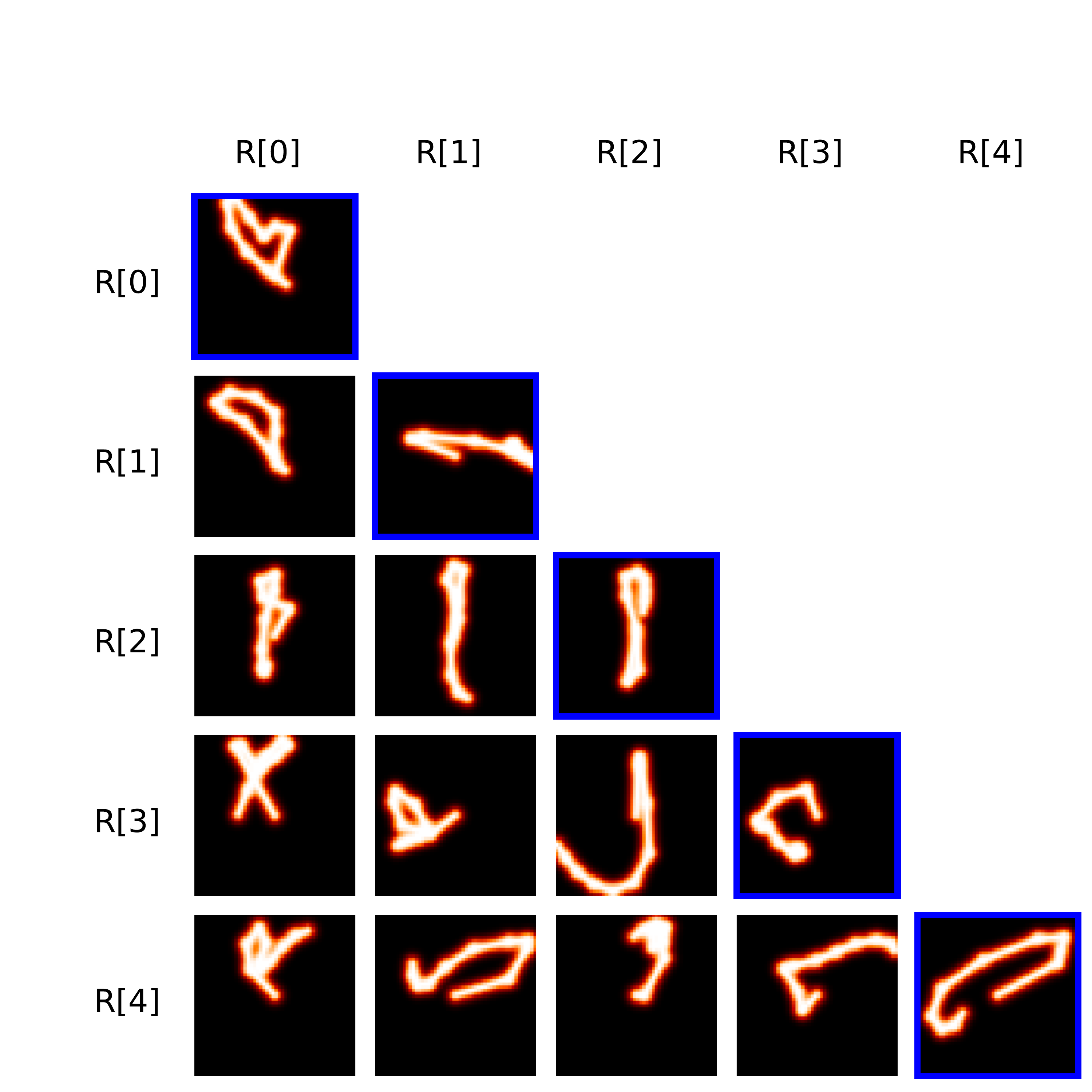}  &
     \includegraphics[width=0.4\textwidth]{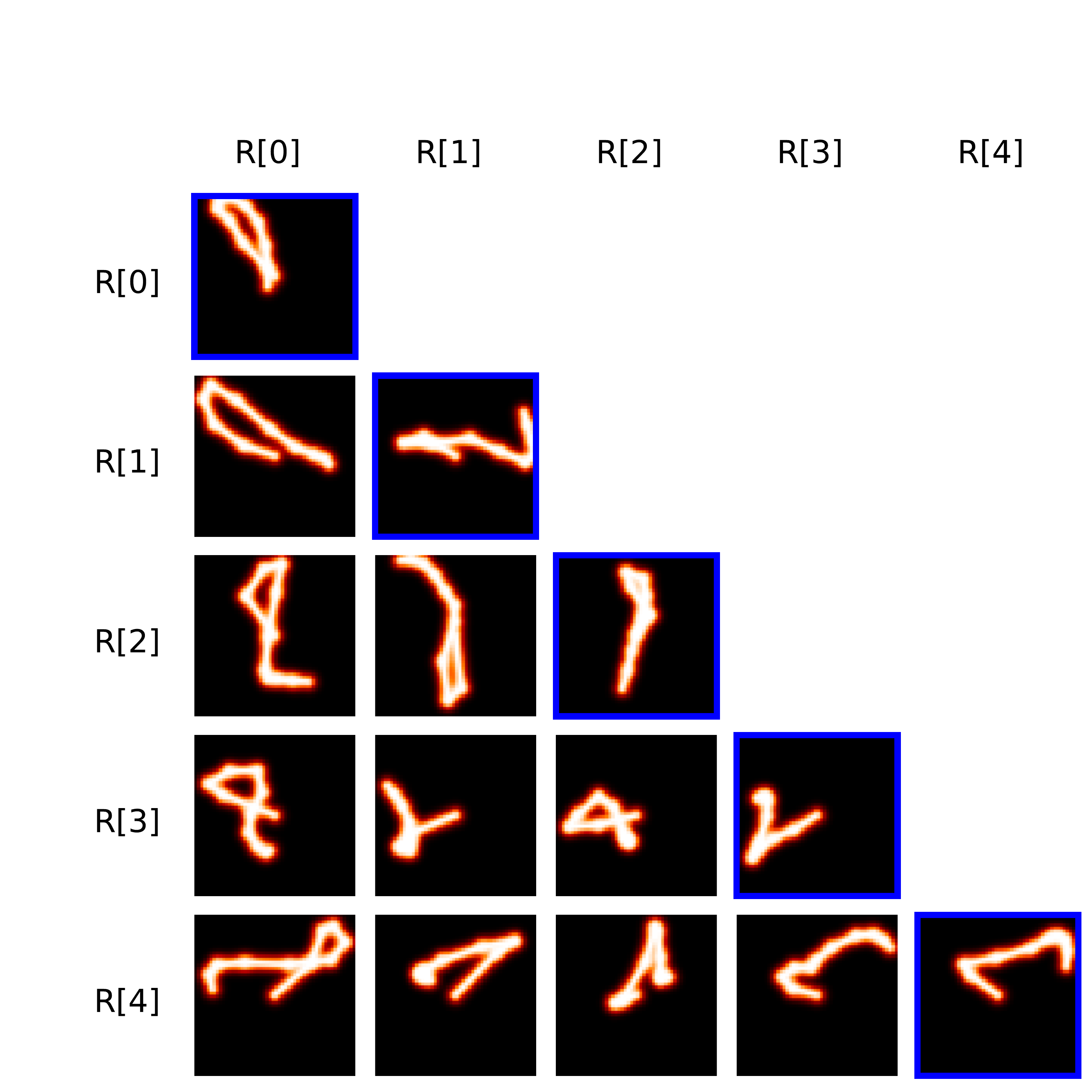}  \\
     \includegraphics[width=0.4\textwidth]{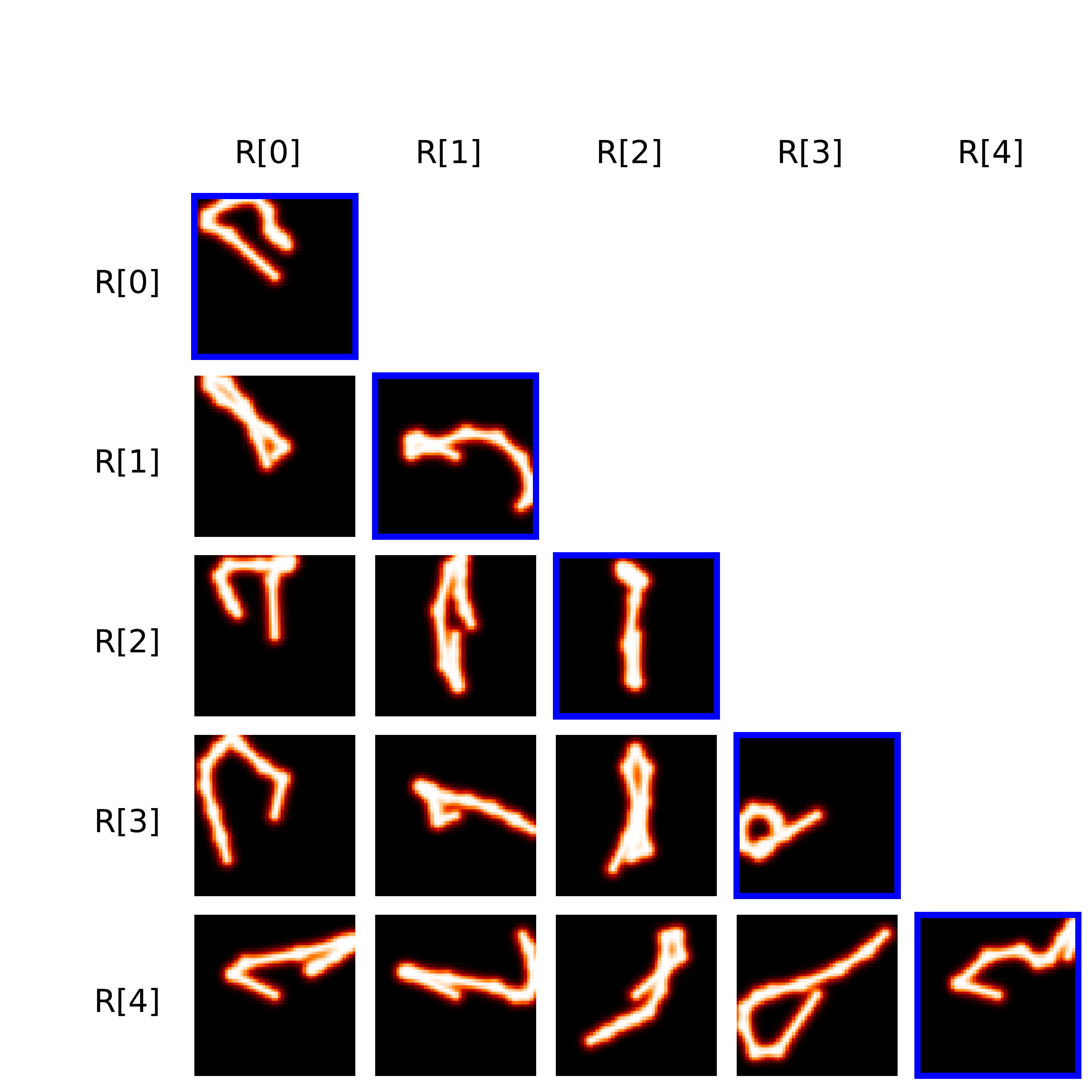} &  
     \includegraphics[width=0.4\textwidth]{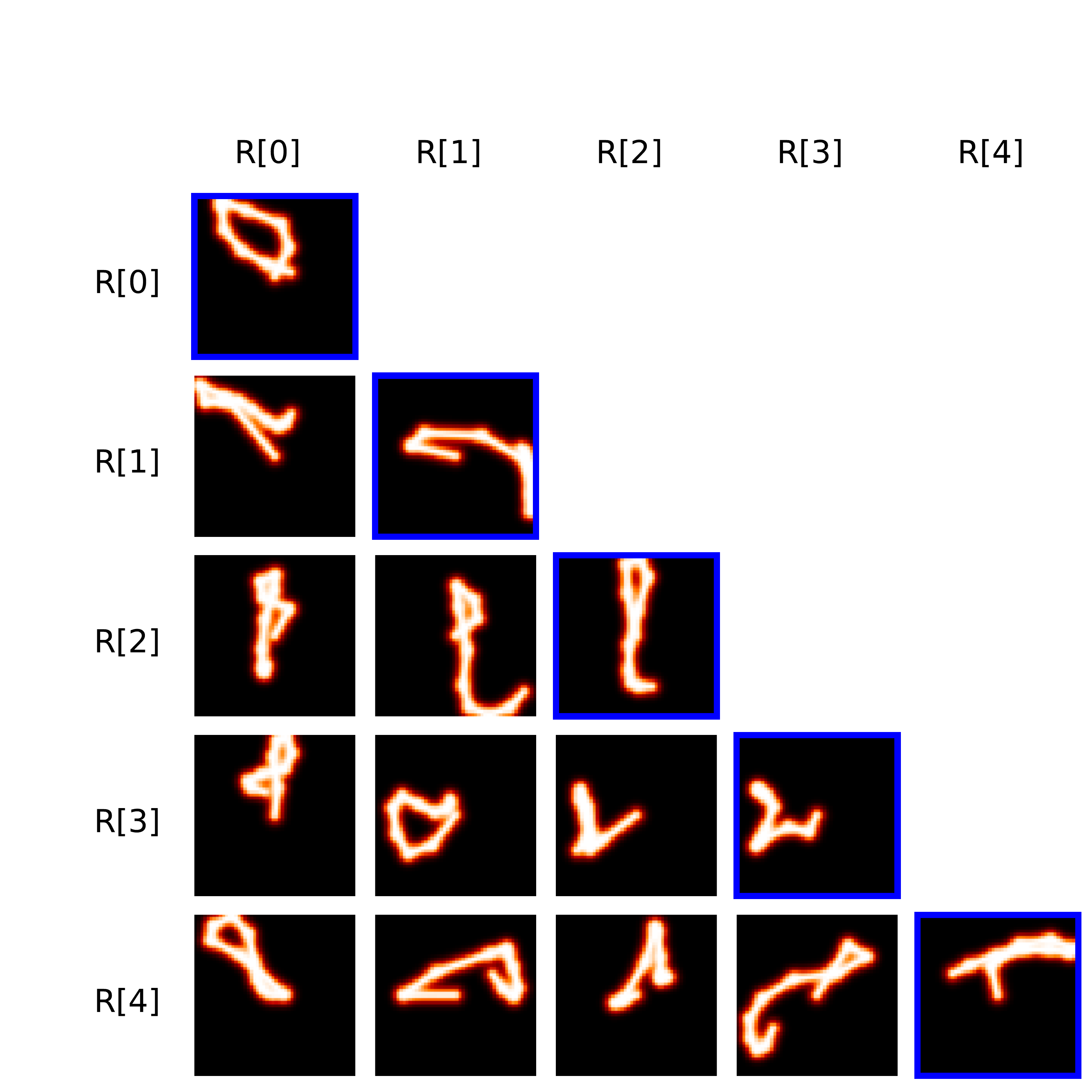} \\
    
 \end{tabular}

\caption{Instances of descriptive utterances for referents from $R_1$ (blue frames) and $R_2$.}
\end{figure}

\newpage
\subsection{T-SNEs of embeddings (Visual - Unshared Perspectives)}
\label{sup:tsnes}

\subsubsection{$R_2$ referents \& descriptive utterances}

\begin{figure}[h!]
\centering 

 \begin{tabular}{cc}

     \includegraphics[width=0.45\textwidth]{figures/tsnes/descriptive/9-TSNE-R2-R01.pdf} & 
     \includegraphics[width=0.45\textwidth]{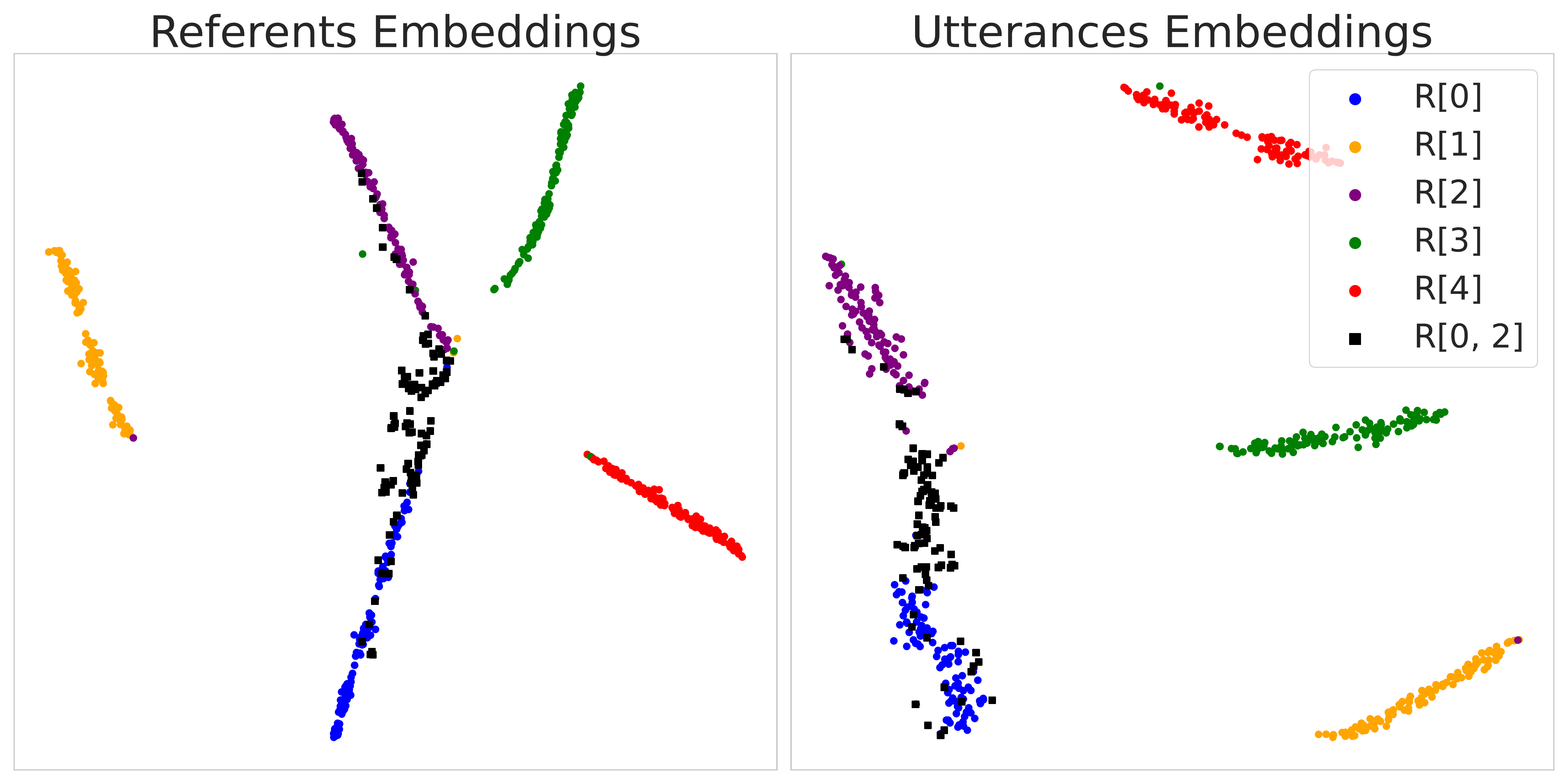} \\
     \includegraphics[width=0.45\textwidth]{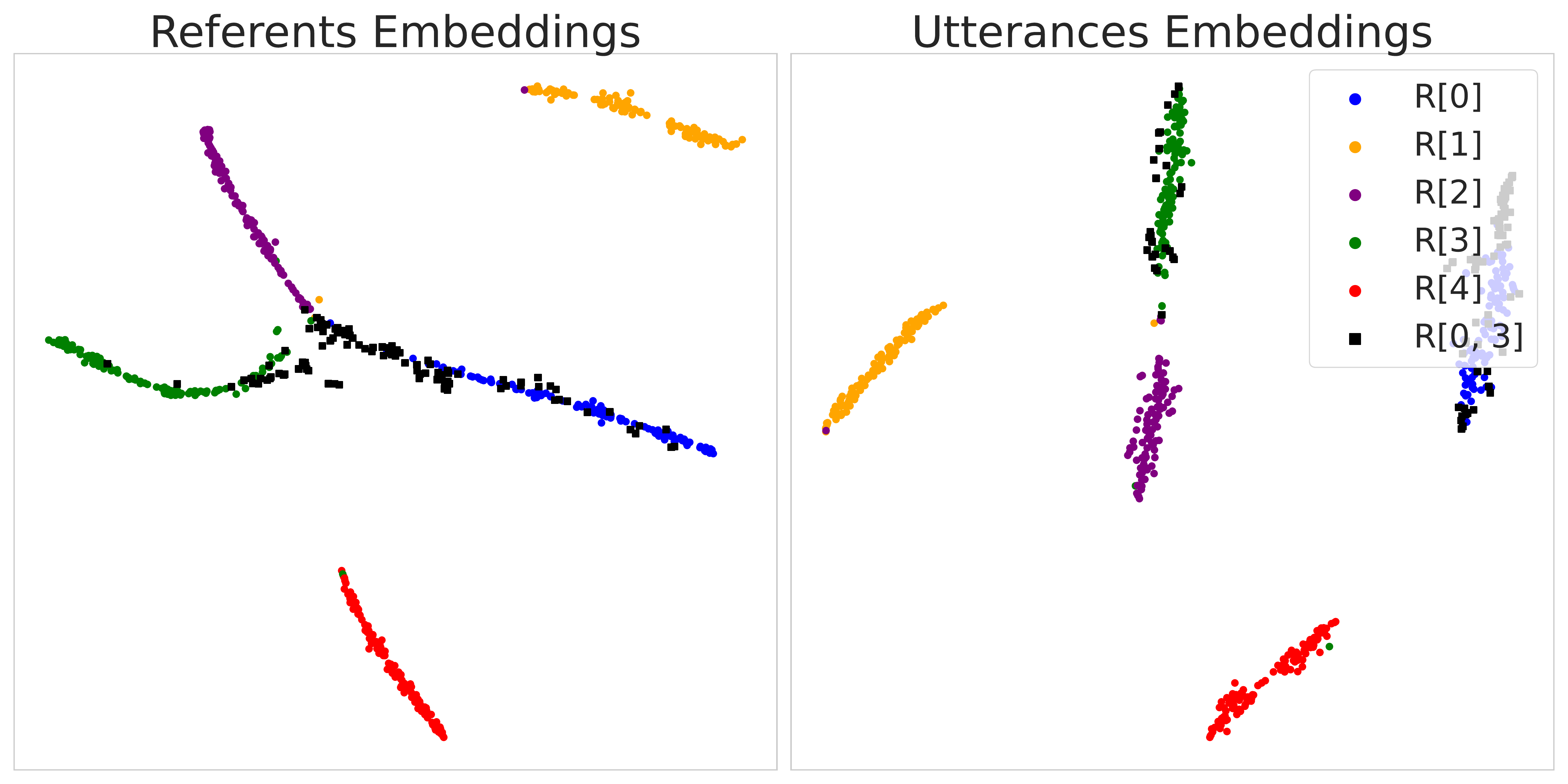}  &
     \includegraphics[width=0.45\textwidth]{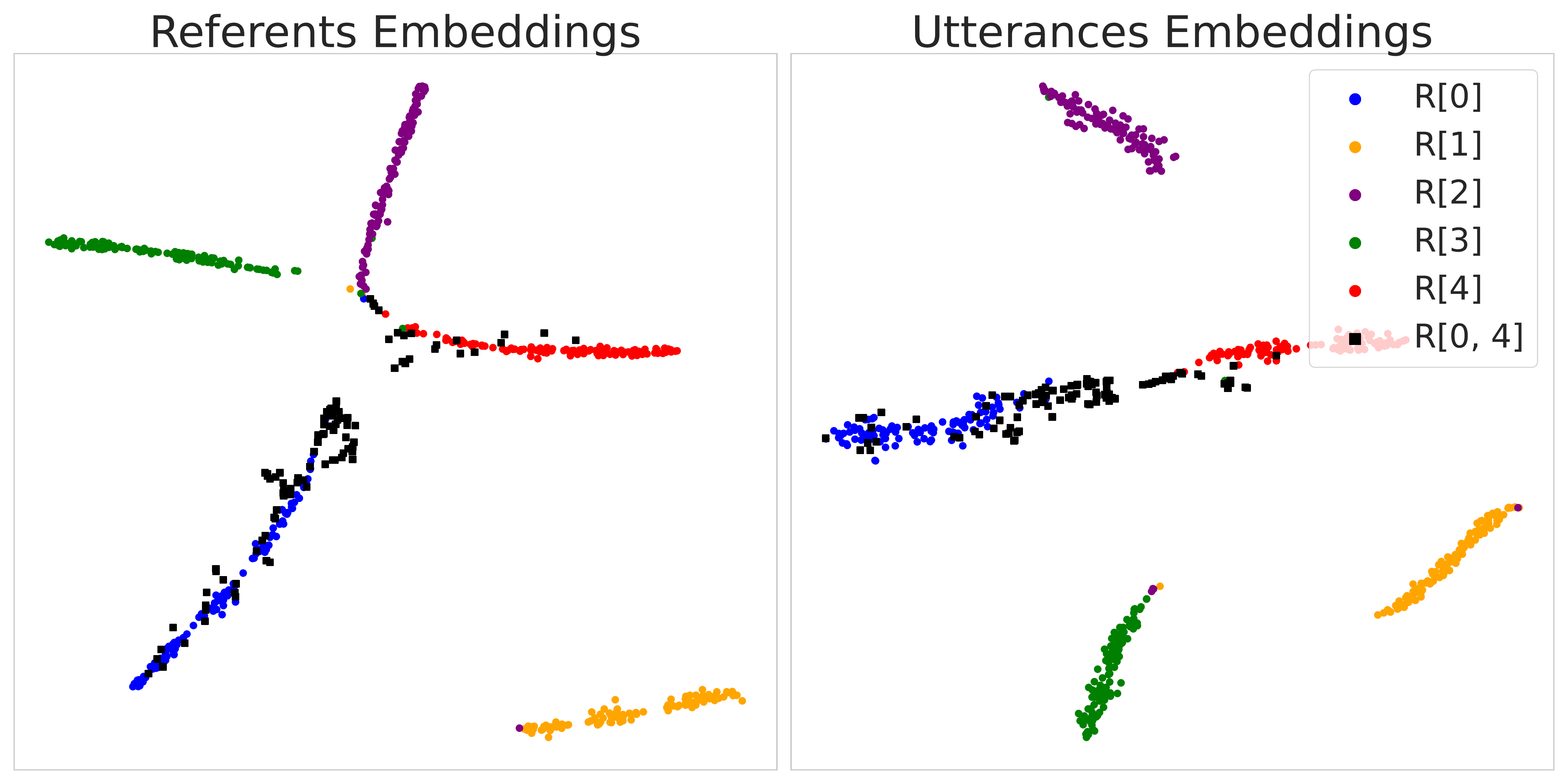}  \\
     \includegraphics[width=0.45\textwidth]{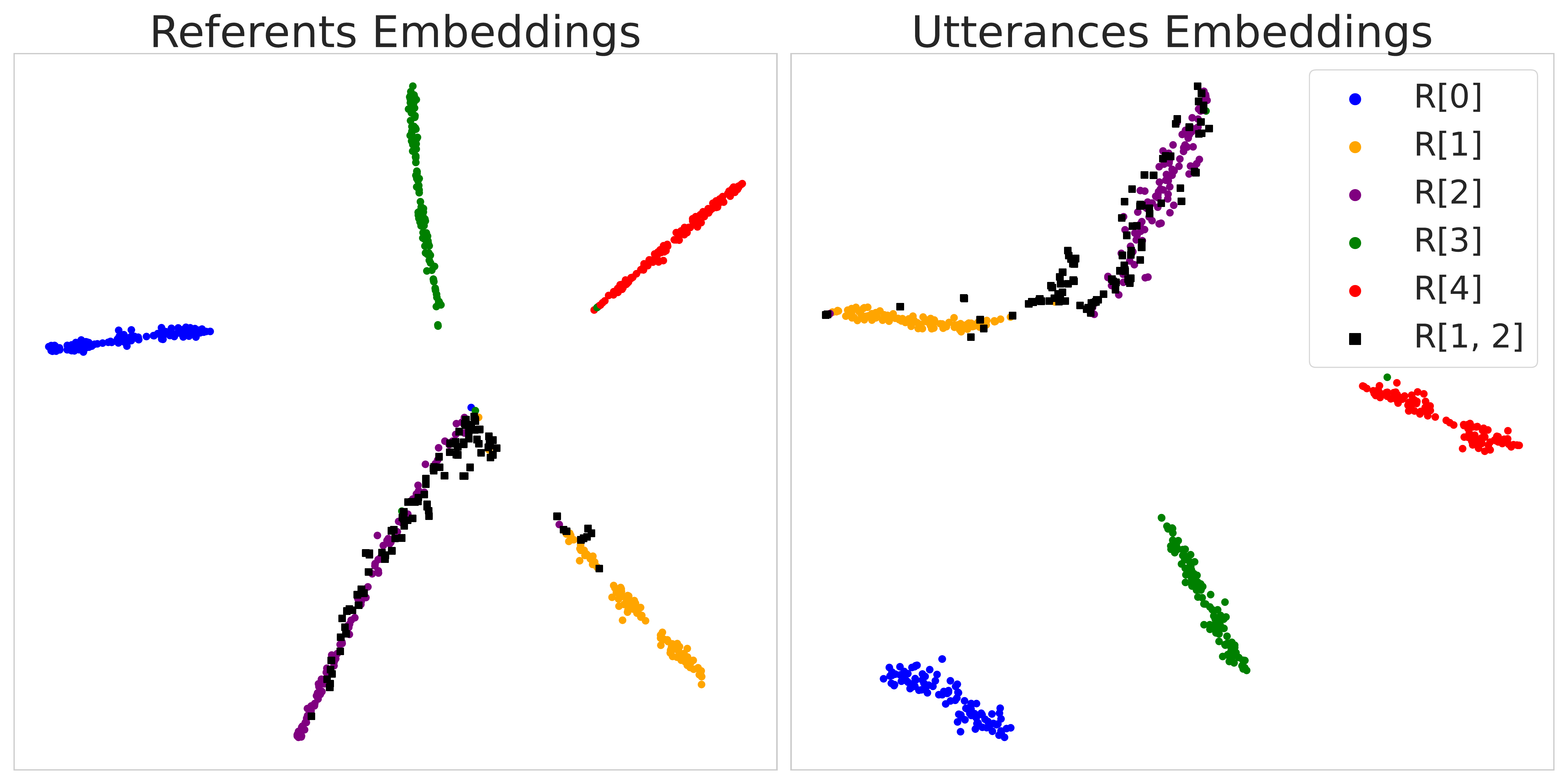} &  
     \includegraphics[width=0.45\textwidth]{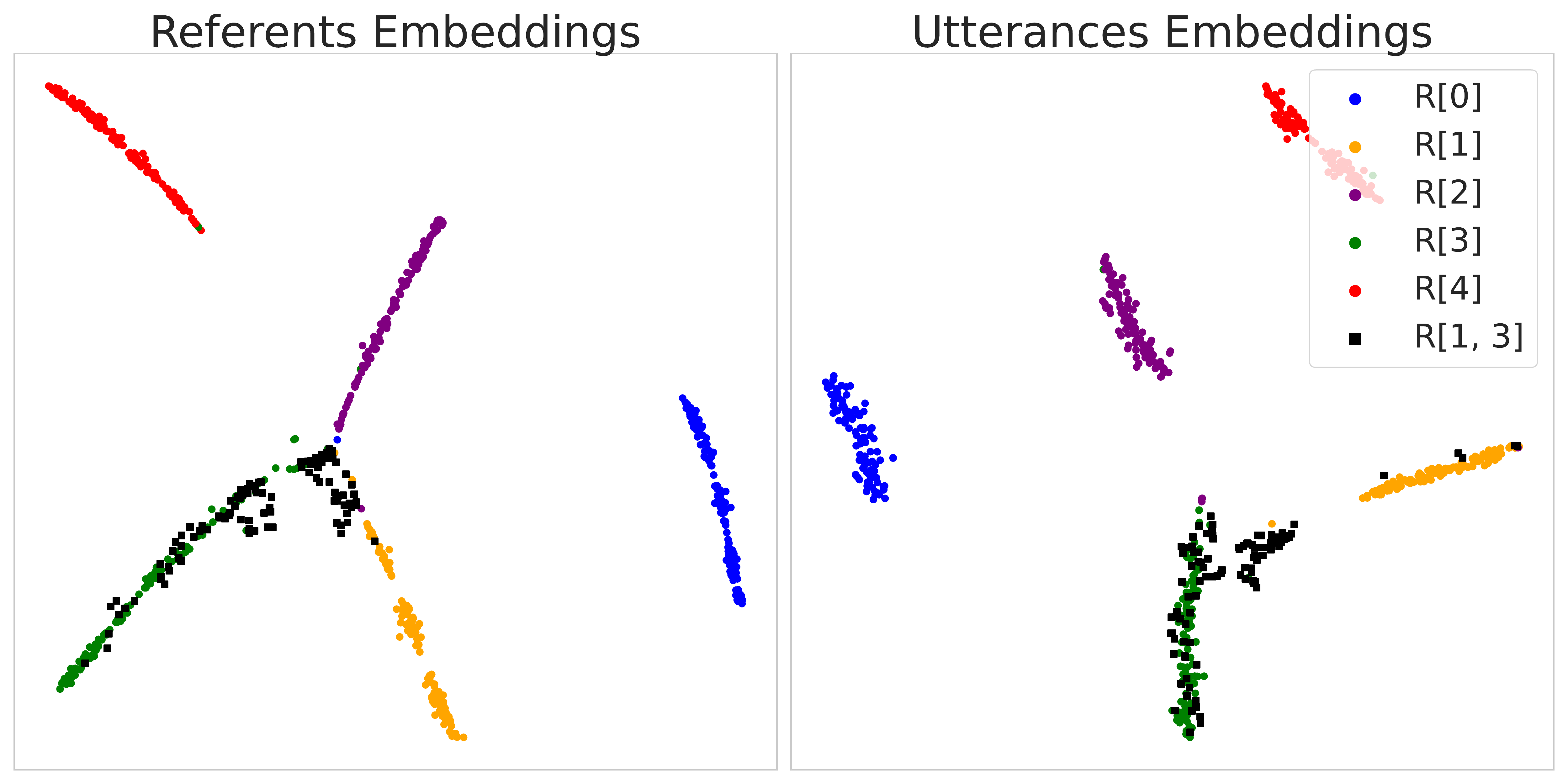}    \\
     \includegraphics[width=0.45\textwidth]{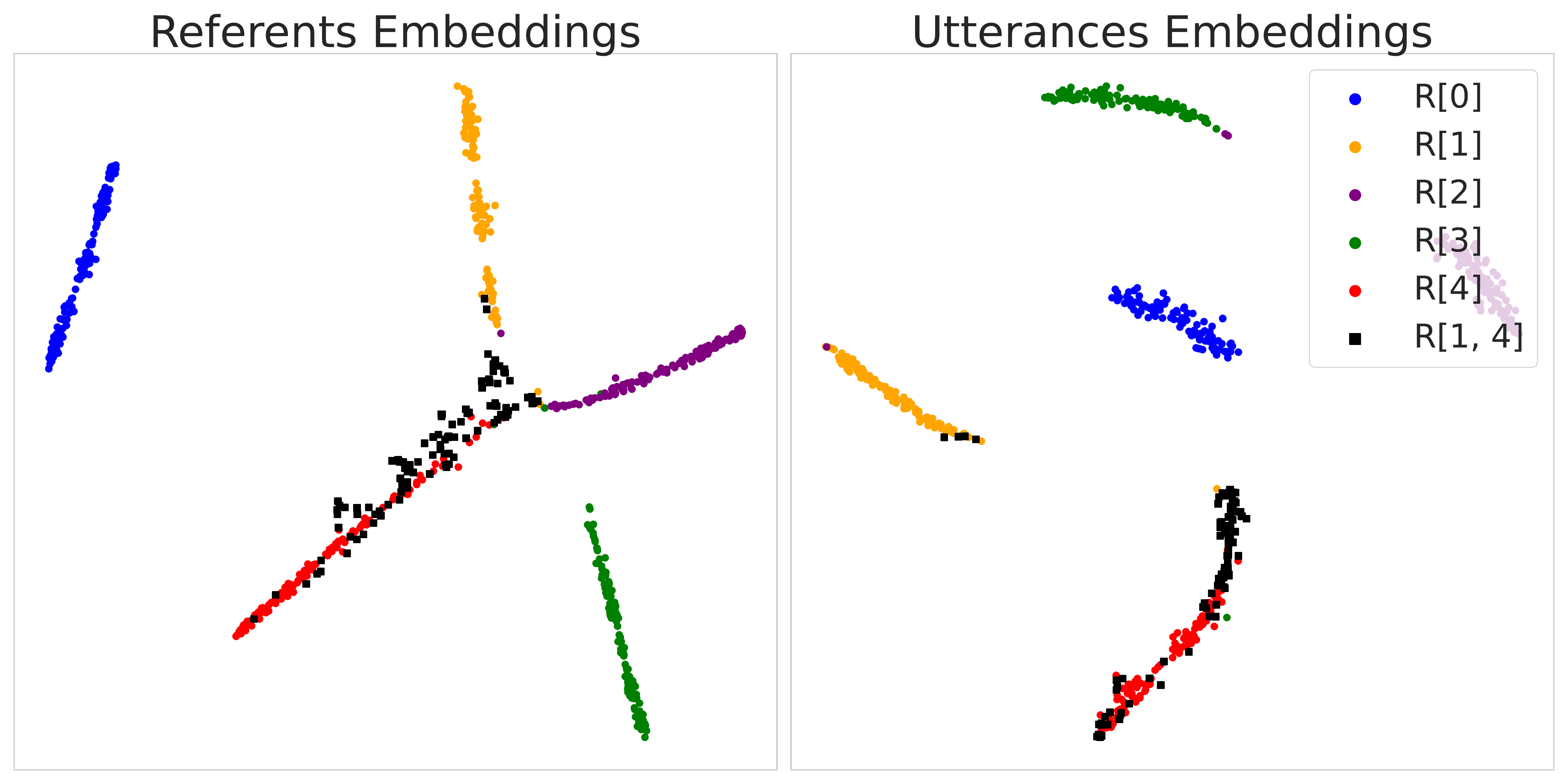} & 
     \includegraphics[width=0.45\textwidth]{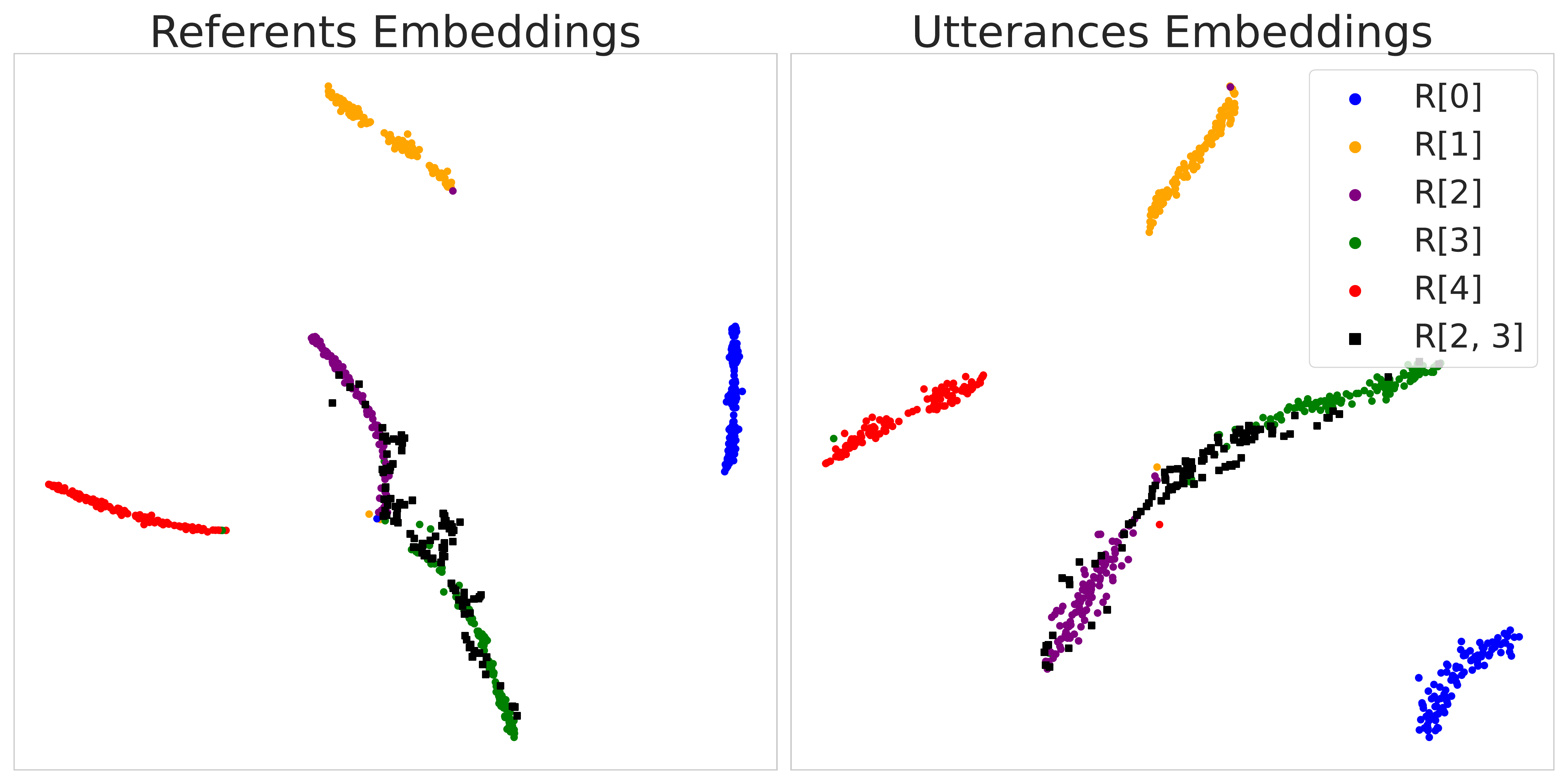} \\
     \includegraphics[width=0.45\textwidth]{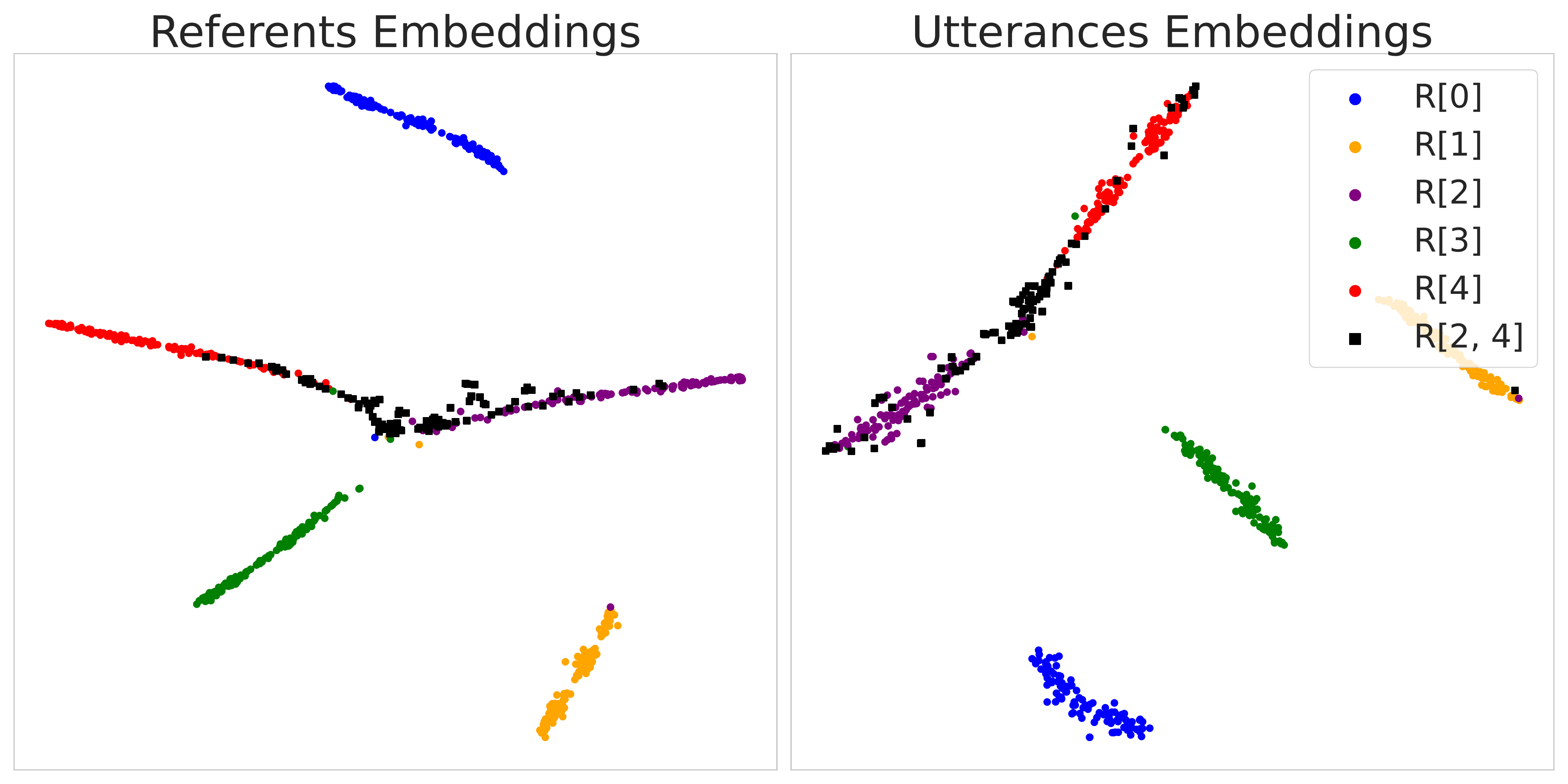}  &
     \includegraphics[width=0.45\textwidth]{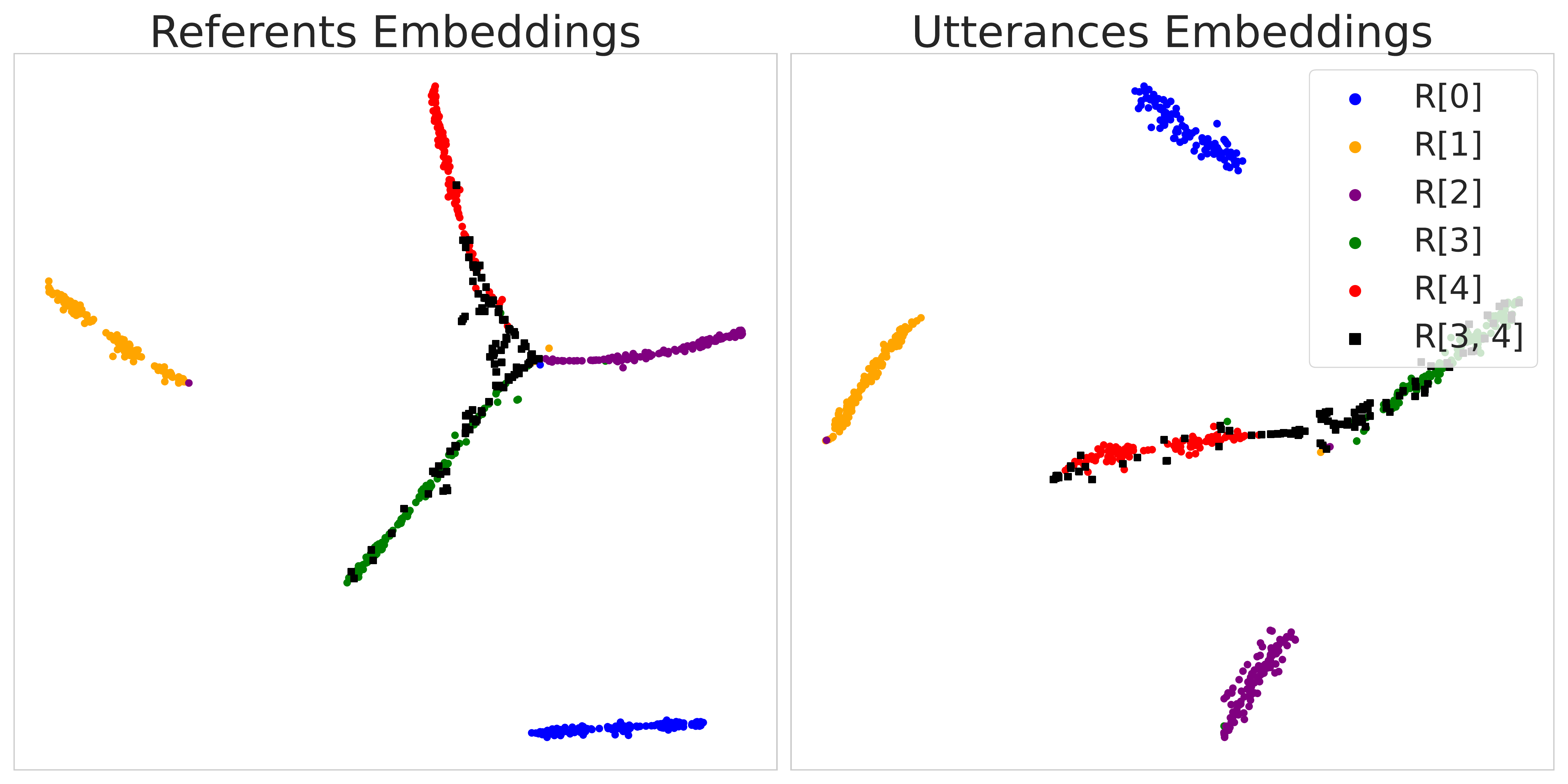}
    
 \end{tabular}

\caption{\textbf{T-sne of referent and descriptive utterance embeddings.} Embeddings are computed for 100 perspectives of referents from $R_2$. Training conditions are unshared visual referents.}
\end{figure}

\newpage

\subsubsection{$R_2$ referents \& discriminative utterances}

\begin{figure}[h!]
 \begin{tabular}{cc}

     \includegraphics[width=0.45\textwidth]{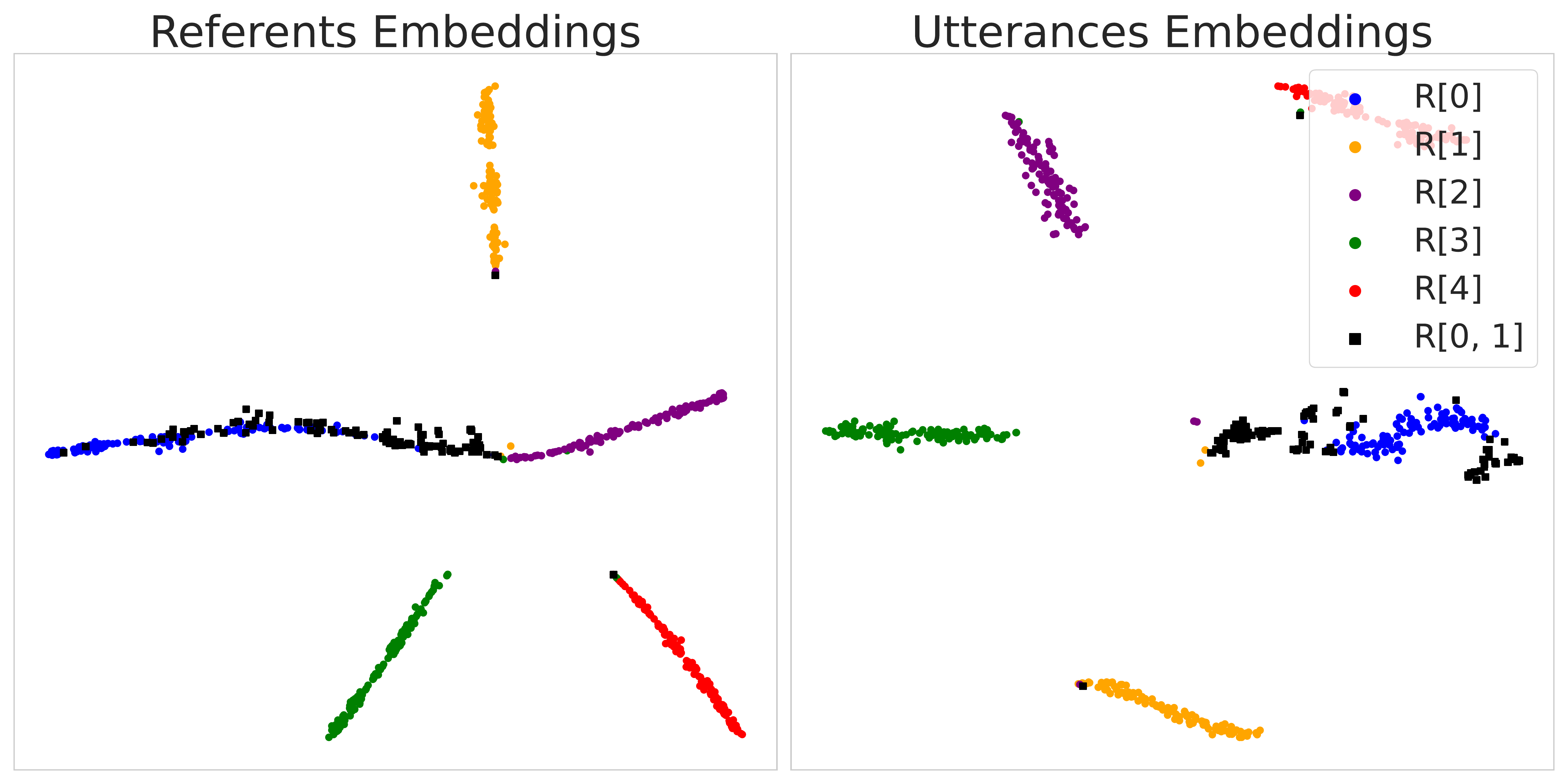} & 
     \includegraphics[width=0.45\textwidth]{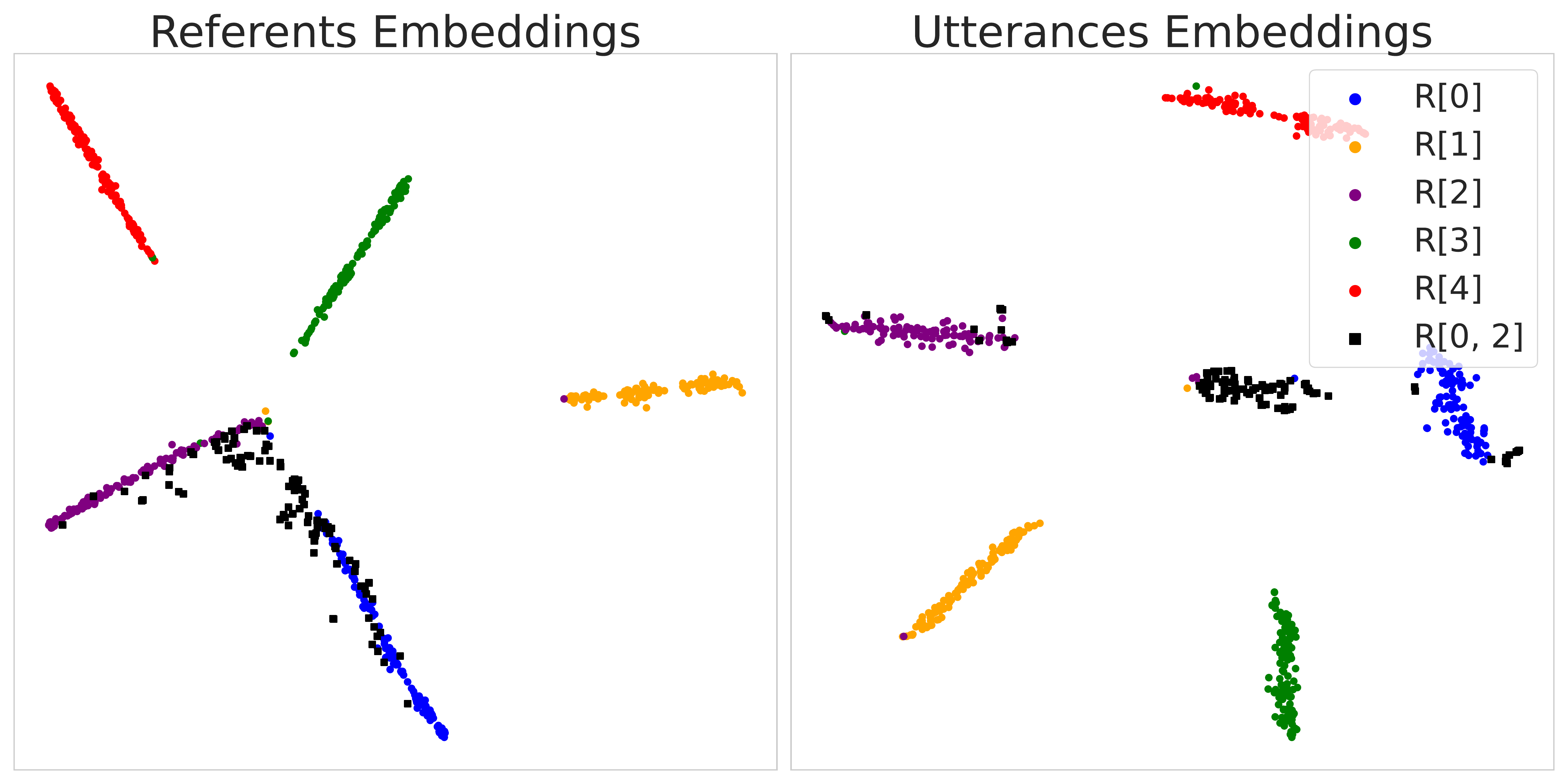} \\
     \includegraphics[width=0.45\textwidth]{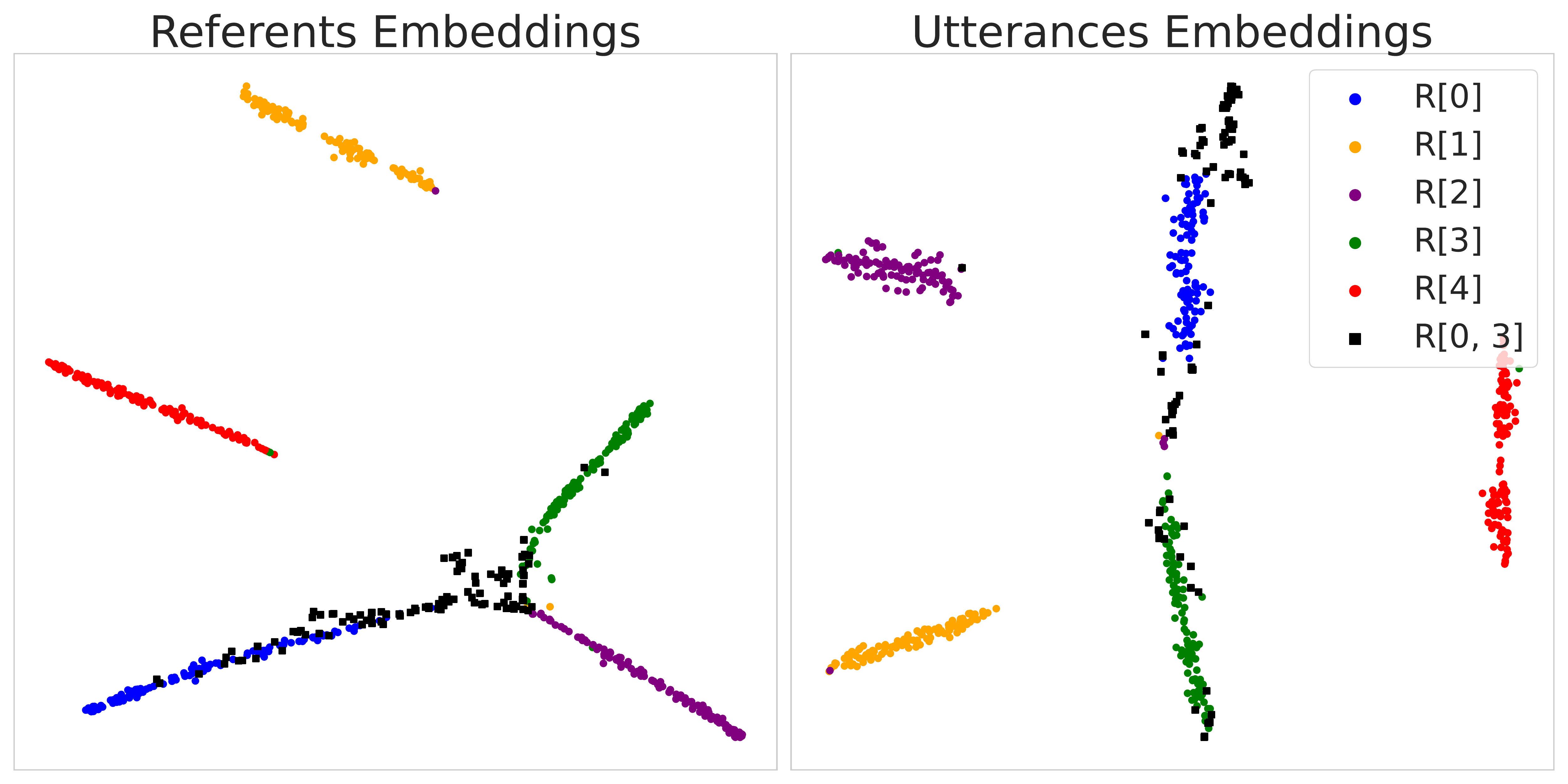}  &
     \includegraphics[width=0.45\textwidth]{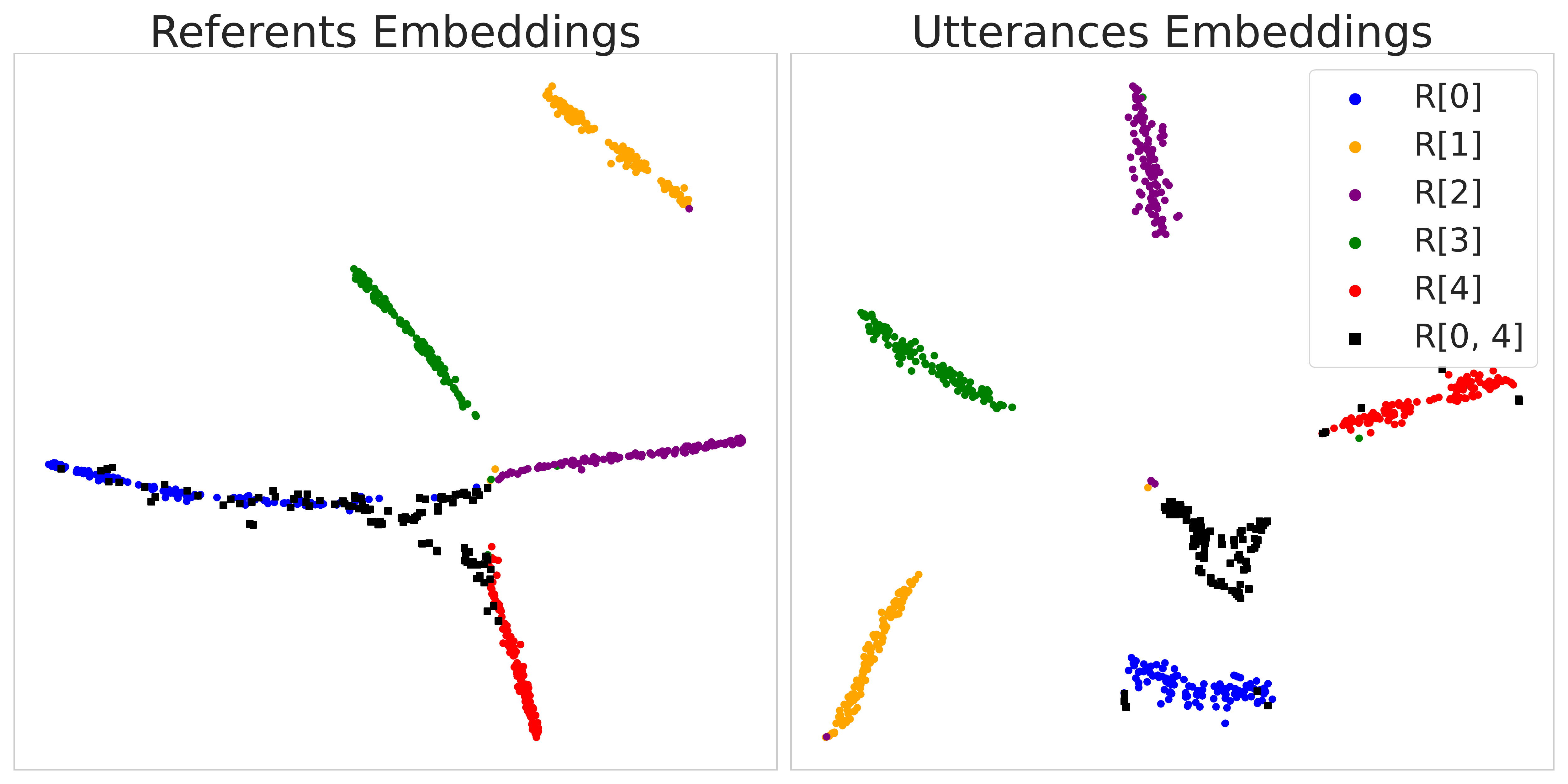}  \\
     \includegraphics[width=0.45\textwidth]{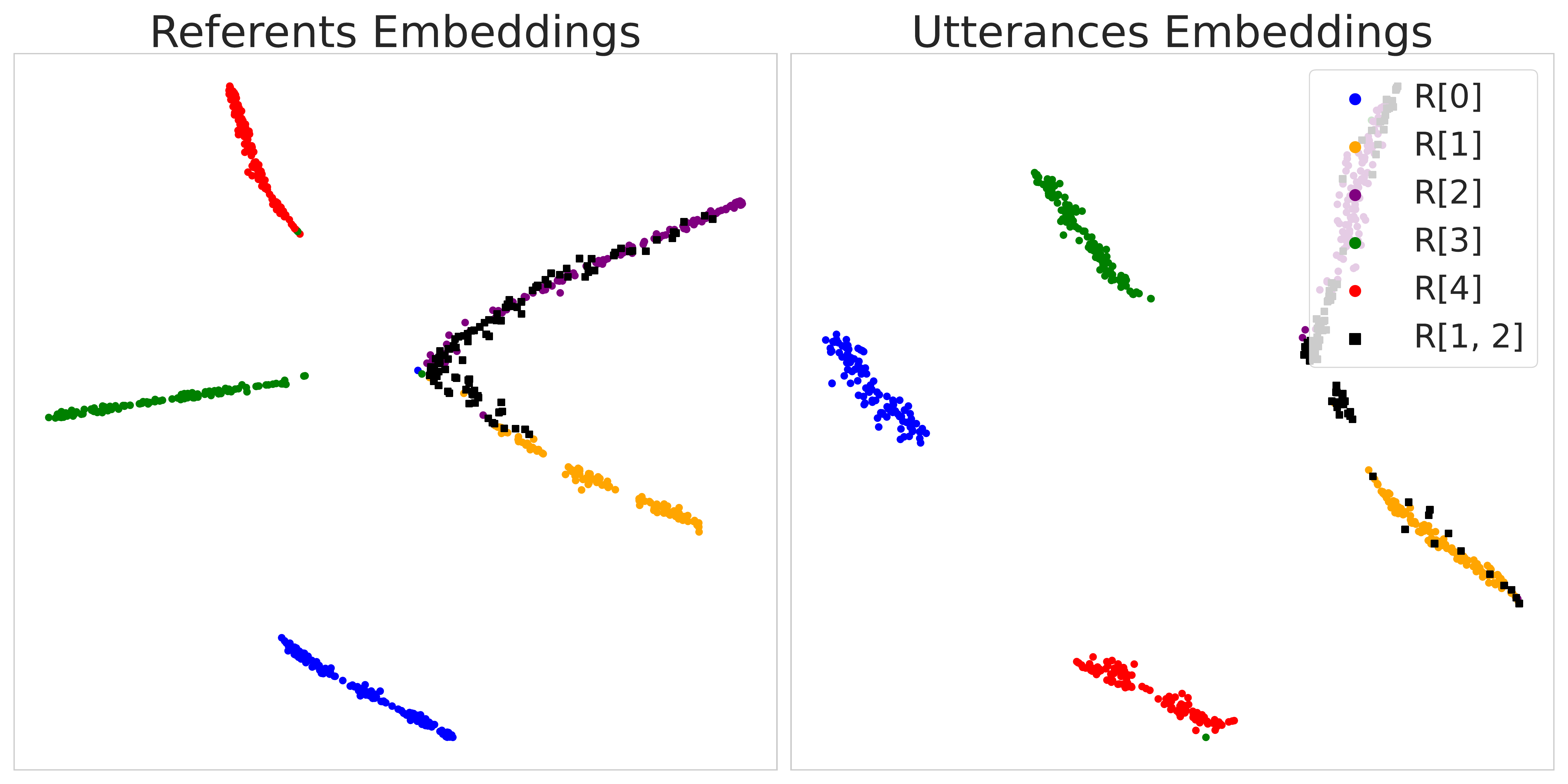} &  
     \includegraphics[width=0.45\textwidth]{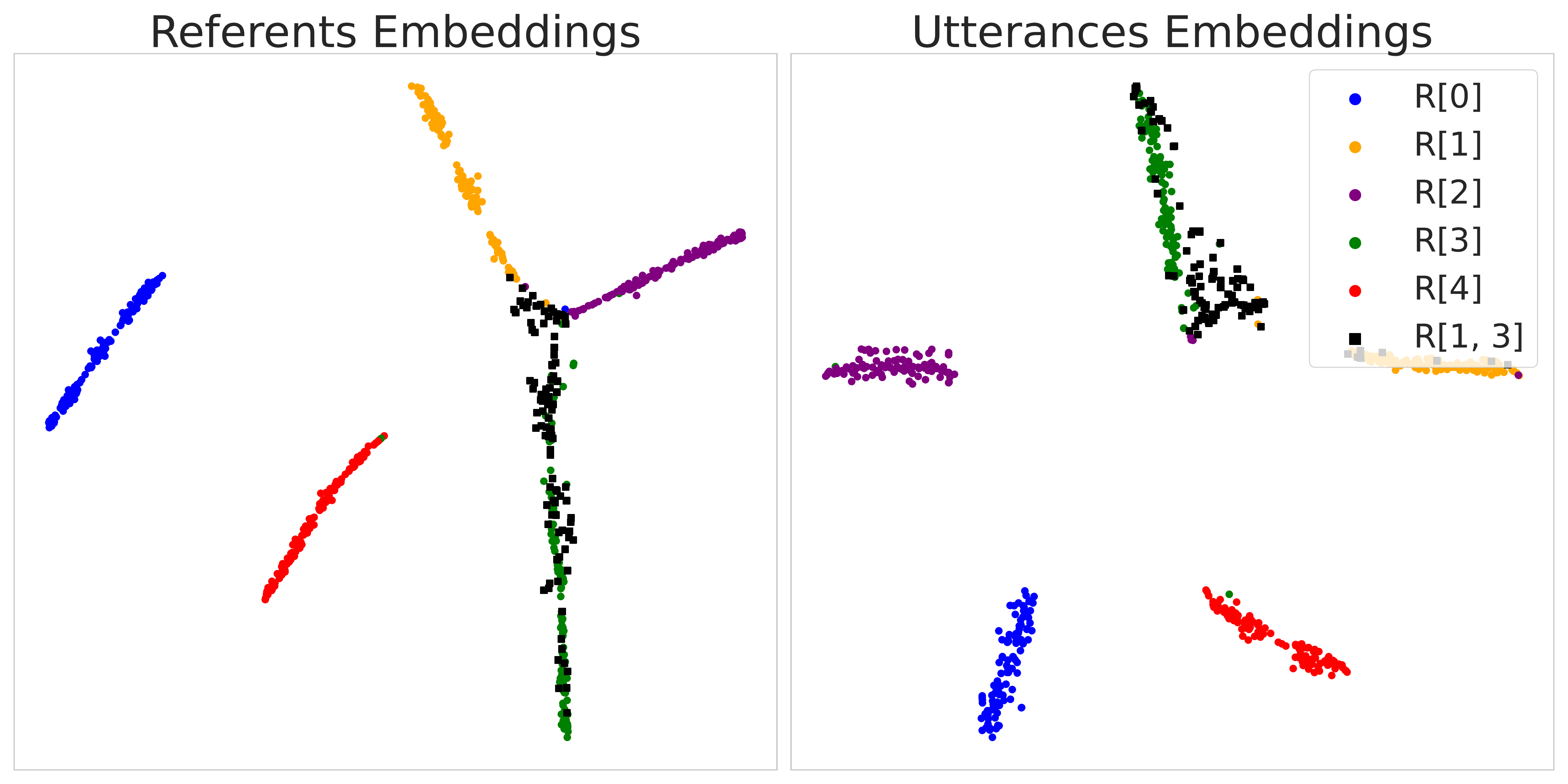}    \\
     \includegraphics[width=0.45\textwidth]{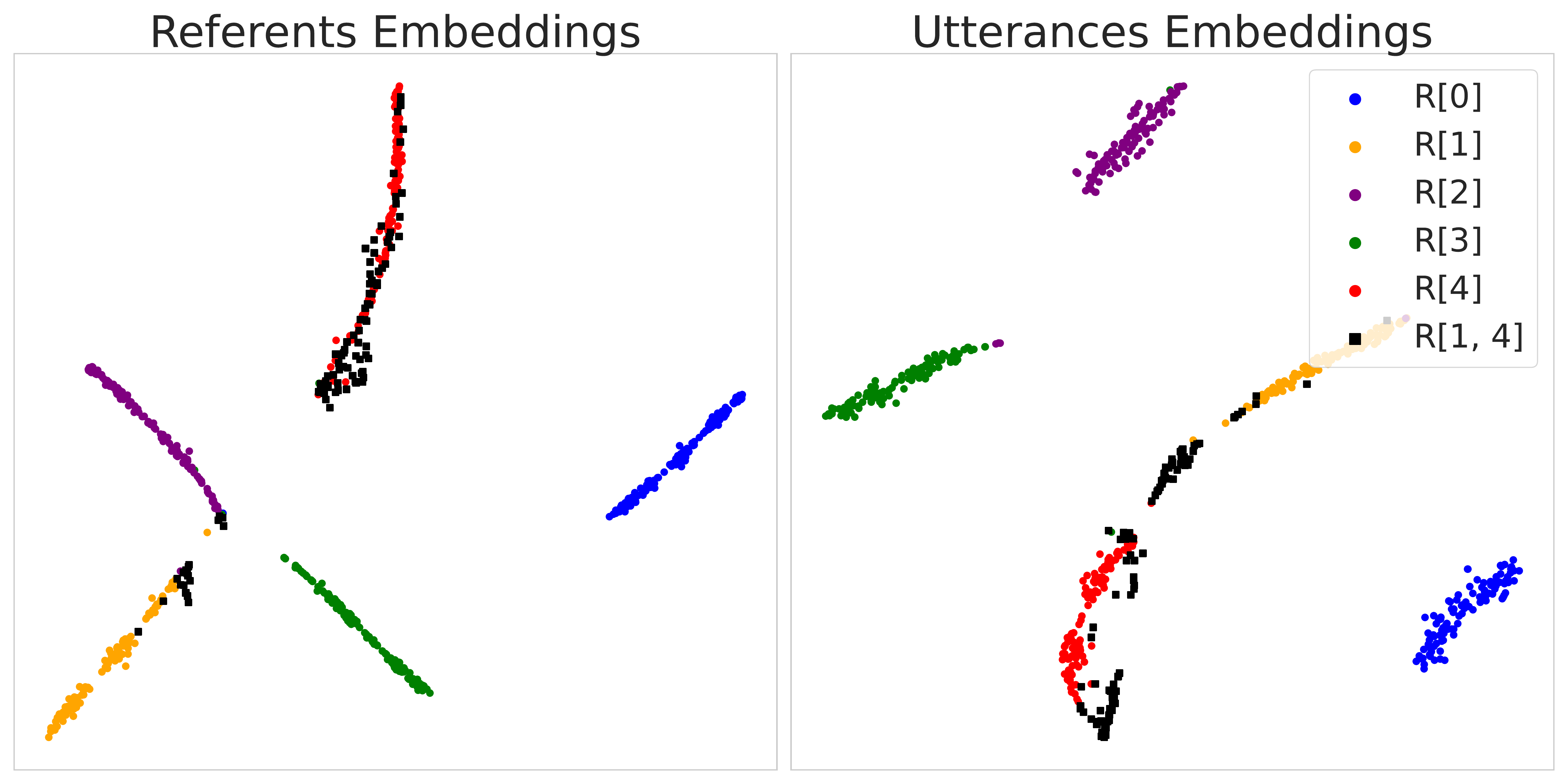} & 
     \includegraphics[width=0.45\textwidth]{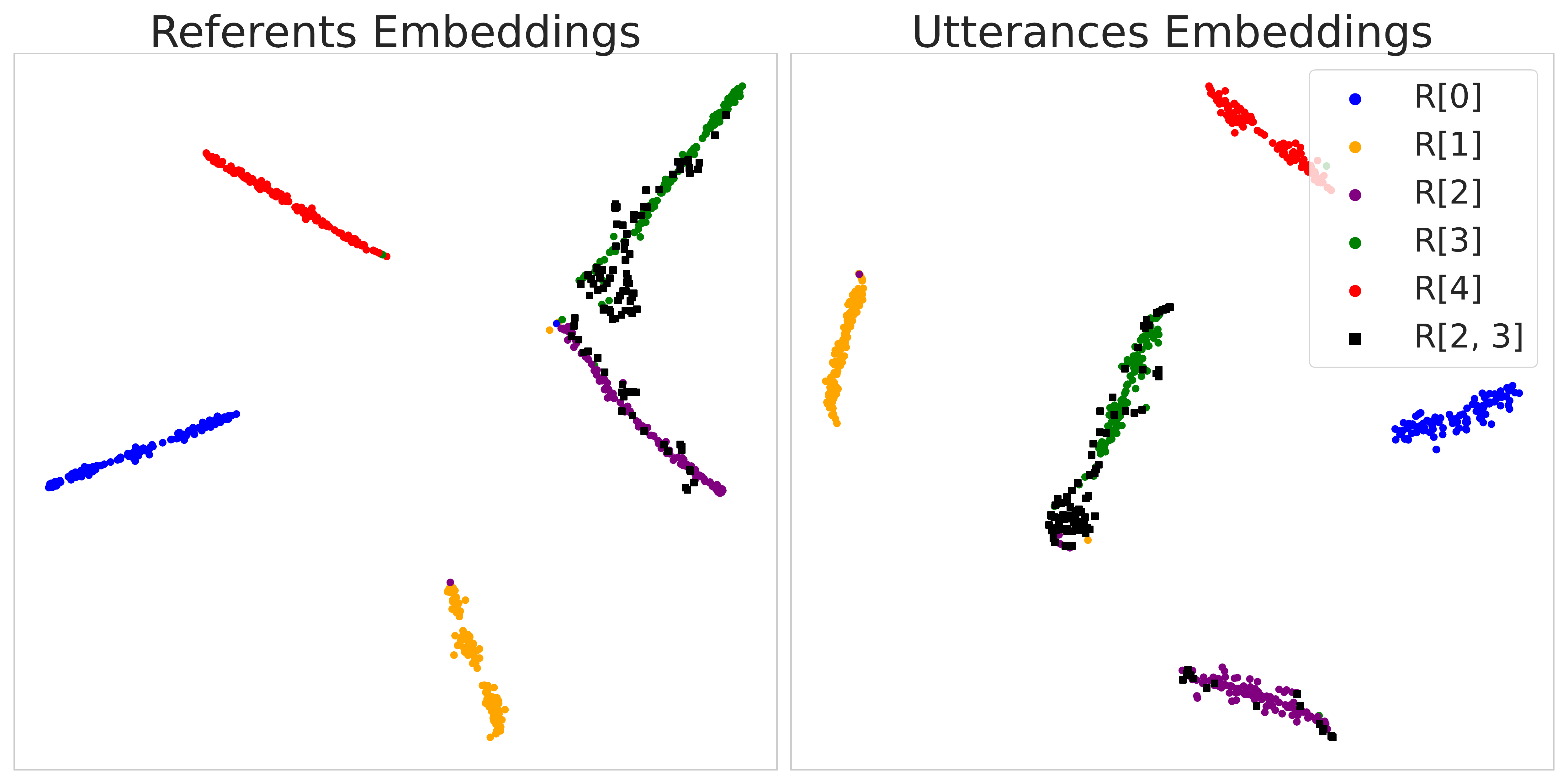} \\
     \includegraphics[width=0.45\textwidth]{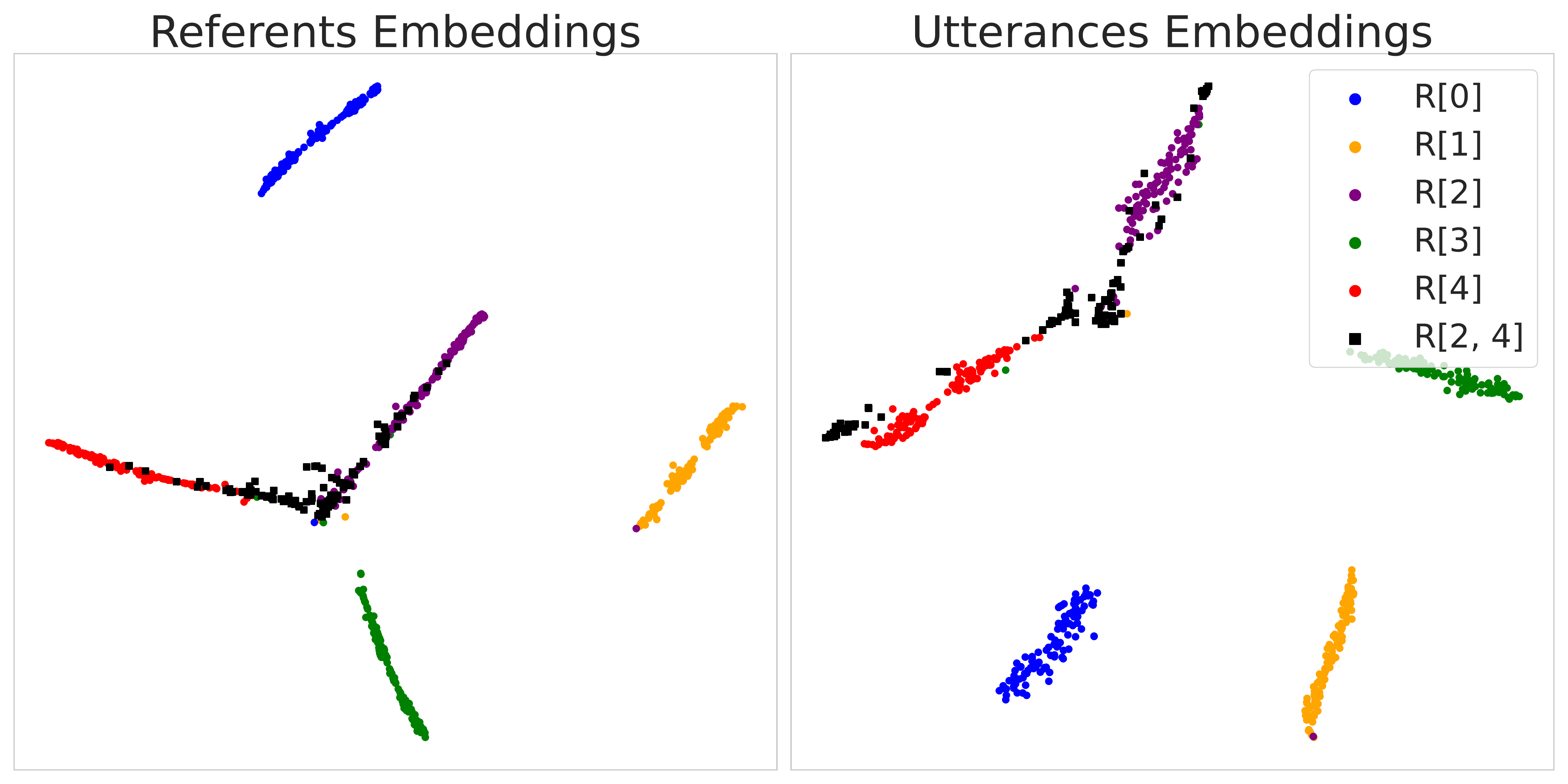}  &
     \includegraphics[width=0.45\textwidth]{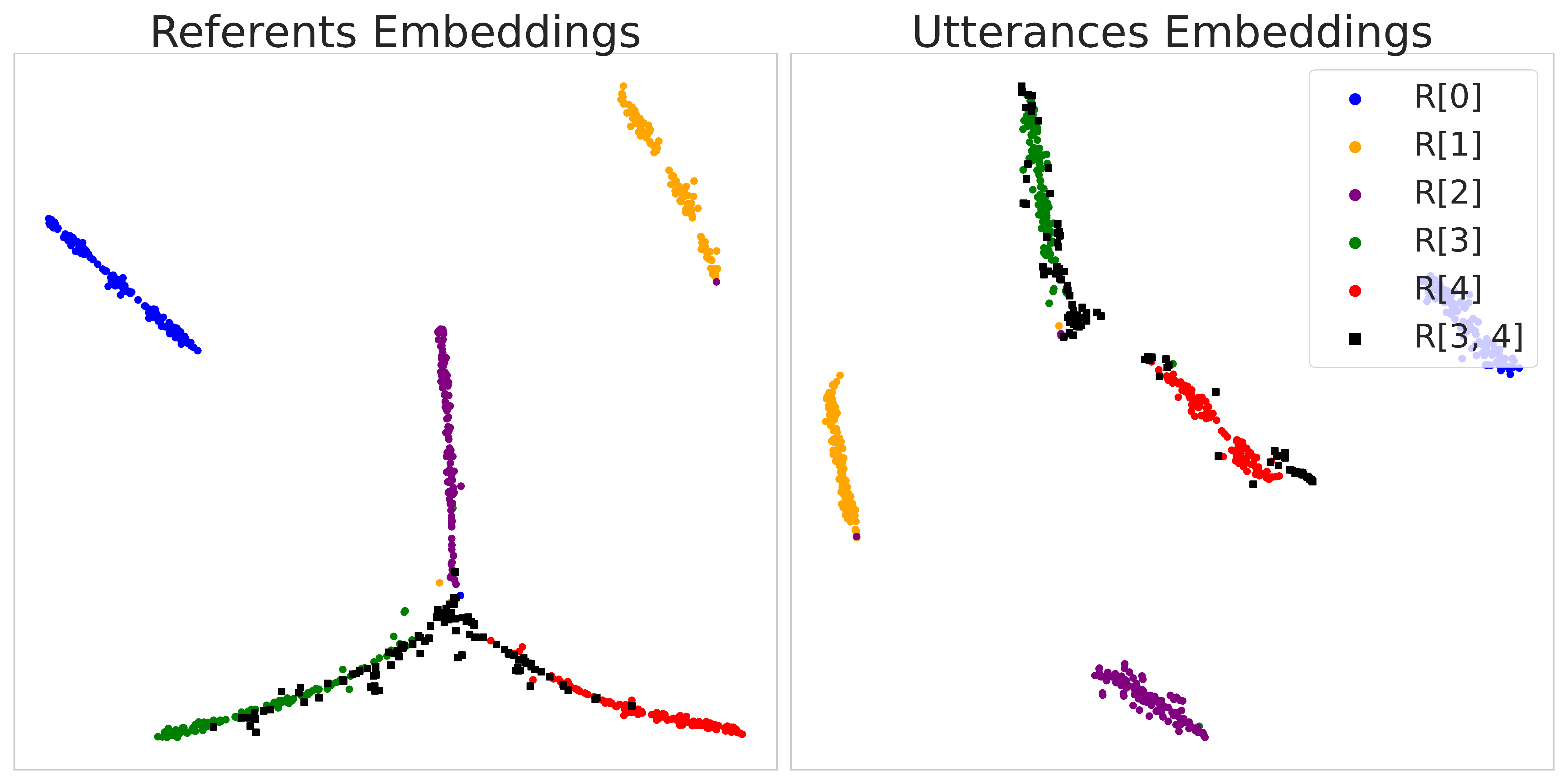}
 \end{tabular}
\caption{\textbf{T-sne of referent and discriminative utterance embeddings.}  Embeddings are computed for 100 perspectives of referents from $R_2$. Training conditions are unshared visual referents.}
\end{figure}


\end{document}